%% file: neurips_2025.tex
\documentclass{article}

\usepackage[preprint]{neurips_2025}

\usepackage[utf8]{inputenc} 
\usepackage[T1]{fontenc}    
\usepackage{hyperref}       
\usepackage{url}            
\usepackage{booktabs}       
\usepackage{amsfonts}       
\usepackage{nicefrac}       
\usepackage{microtype}      
\usepackage[dvipsnames]{xcolor}         

\usepackage{graphicx}
\usepackage{amsmath}
\usepackage{amsthm}
\usepackage{algorithm}
\usepackage{algorithmic}

\usepackage{wrapfig}
\usepackage{titletoc}
\usepackage[framemethod=tikz]{mdframed}

\usepackage{subcaption}
\usepackage{dsfont}
\usepackage{multirow}
\usepackage{enumitem}
\usepackage{stfloats}

\usepackage{xspace}
\def\eg{\emph{e.g}.\xspace}
\def\ie{\emph{i.e}.\xspace}

\input{math_commands.tex}

\title{New Paradigm of Adversarial Training: Releasing \\ Accuracy-Robustness Trade-Off via Dummy Class}

\author{%
  Yanyun Wang$^{1}$, Li Liu$^{1}$\thanks{Corresponding author. E-mail: \texttt{avrillliu@hkust-gz.edu.cn}.}, $\hspace{0.05em}$ Zi Liang$^{2}$, Yi R. (May) Fung$^{3}$, Qingqing Ye$^{2}$, Haibo Hu$^{2}$ \vspace{0.5em} \\
  $\hspace{0.2em}^{1}$The Hong Kong University of Science and Technology (Guangzhou)\\
  $\hspace{0.2em}^{2}$The Hong Kong Polytechnic University\\
  $\hspace{0.2em}^{3}$The Hong Kong University of Science and Technology\\
}

\begin{document}
\startcontents[main]

\maketitle

\begin{abstract}
Adversarial Training (AT) is one of the most effective methods to enhance the robustness of Deep Neural Networks (DNNs). However, existing AT methods suffer from an inherent accuracy-robustness trade-off. Previous works have studied this issue under the current AT paradigm, but still face over 10\% accuracy reduction without significant robustness improvement over simple baselines such as PGD-AT. This inherent trade-off raises a question: \textit{\textbf{Whether the current AT paradigm, which assumes to learn corresponding benign and adversarial samples as the same class, inappropriately mixes clean and robust objectives that may be essentially inconsistent}}. In fact, our empirical results show that up to 40\% of CIFAR-10 adversarial samples always fail to satisfy such an assumption across various AT methods and robust models, explicitly indicating the room for improvement of the current AT paradigm. To relax from this overstrict assumption and the tension between clean and robust learning, in this work, we propose a new AT paradigm by introducing an additional dummy class for each original class, aiming to accommodate hard adversarial samples with shifted distribution after perturbation. The robustness \textit{w.r.t.} these adversarial samples can be achieved by runtime recovery from the predicted dummy classes to the corresponding original ones, without conflicting with the clean objective on accuracy of benign samples. Finally, based on our new paradigm, we propose a novel \textbf{DU}mmy \textbf{C}lasses-based \textbf{A}dversarial \textbf{T}raining (\textbf{DUCAT}) method that concurrently improves accuracy and robustness in a plug-and-play manner only relevant to \textit{logits}, \textit{loss}, and a proposed two-hot soft label-based supervised signal. Our method outperforms state-of-the-art (SOTA) benchmarks, effectively releasing the current trade-off. The code is available at \url{https://github.com/FlaAI/DUCAT}. 
\end{abstract}

\section{Introduction}\label{sec:intro}

Deep Neural Networks (DNNs) demonstrate remarkable performance in various real-world applications, but remain vulnerable to adversarial attacks~\citep{biggio2013evasion,szegedy2014intriguing}. Specifically, malicious input perturbations that are imperceptible to humans can cause significant changes in the output of DNNs~\citep{goodfellow2015explaining}, raising serious security concerns within both the public and research communities~\citep{wang2020improving}. In response, several defense mechanisms have been proposed to enhance the adversarial robustness of DNNs, such as \textit{defense distillation}~\citep{papernot2017practical}, \textit{feature squeezing}~\citep{xu2017feature}, \textit{randomization}~\citep{xie2018mitigating}, and \textit{input denoising}~\citep{guo2018countering,liao2018defense}. However, most of them have proven subsequently to rely on \textit{obfuscated gradients}~\citep{athalye2018obfuscated} and be ineffective against advanced adaptive attacks~\citep{tramer2020adaptive}, such as Auto-Attack (AA)~\citep{croce2020reliable}. 

Currently, Adversarial Training (AT)~\citep{goodfellow2015explaining,madry2018towards} is demonstrated as one of the most effective approaches to train inherently robust DNNs~\citep{athalye2018obfuscated,dong2020benchmarking}. Different from clean training, \textbf{the current AT paradigm} directly learns adversarially augmented samples with the original labels to yield robust models. Research indicates that this paradigm is equivalent to optimizing an upper bound of natural risk on the original data, and can thereby serve as a principle against adversarial attacks~\citep{tao2021better}. To the best of our knowledge, despite considerable advancements in the specific mechanism of existing AT methods, most of them remain adhering to this basic paradigm. The solid AT benchmarks such as PGD-AT~\citep{madry2018towards}, TRADES~\citep{zhang2019theoretically} and MART~\citep{wang2020improving} are representative examples.

\begin{wrapfigure}[22]{r}{7cm}
\centering
\vspace{-0.8em}
\includegraphics[width=0.93\linewidth]{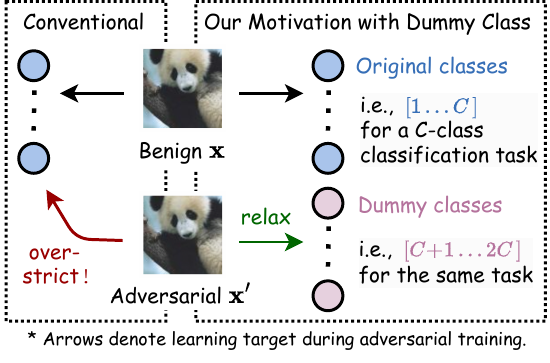}
\vspace{-0.2em}
\caption{Conceptual difference between conventional AT and ours. The current AT paradigm assumes that the adversarial sample $\mathbf{x}'$ should be assigned to the same class as the benign sample $\mathbf{x}$, which may be \textbf{overstrict}. In this work, we suggest introducing more \textbf{dummy classes} one-to-one corresponding to original ones, so that some hard $\mathbf{x}'$ with shifted distribution can be accommodated without significantly hurting clean learning on $\mathbf{x}$.
}\label{fig:intro}
\end{wrapfigure}

However, recent studies indicate that existing AT methods suffer from \textbf{an inherent trade-off} between adversarial robustness and clean accuracy~\citep{tsipras2019robustness,zhang2019theoretically,raghunathan2019adversarial,raghunathan2020understanding,wang2020once,bai2021recent}. This means improving robustness via these AT methods is at the cost of reducing model accuracy compared with the standard training, which can seriously hurt the experience of benign users and greatly reduce the use of AT among real-world DNN application providers. A widely recognized reason of this problem is that AT equally requires predictive consistency within the $\epsilon$-ball of each sample, which can complicate decision boundaries, particularly for those samples near class margins, ultimately degrading clean generalization~\citep{dong2022exploring,rade2022reducing,cheng2022cat,yang2022one}. To learn robustness \textit{w.r.t.} such samples more appropriately, FAT~\citep{zhang2020attacks} and HAT~\citep{rade2022reducing} utilized adaptive perturbations in AT to smooth decision boundaries; Cons-AT~\citep{tack2022consistency} encouraged the similarity of predictive distributions between adversarial samples derived from different augmentations of the same instance, thereby enhancing learnable patterns and reducing label noise; SOVR~\citep{kanai2023one} explicitly increased \textit{logits} margin of important samples by switching from \textit{cross-entropy} to a new one-vs-the-rest loss. 

Despite several previous efforts, there has been limited substantial progress in addressing this trade-off problem. Most impressive advancements in recent years come from those methods introducing extra data~\citep{alayrac2019labels,carmon2019unlabeled,najafi2019robustness,li2022semi} or utilizing generative models~\citep{wang2023better}. However, they not only demand more resources and computational costs but also violate the conventional fairness assumption of AT (\ie, no additional data should be incorporated~\citep{pang2021bag}). Other advanced AT techniques apart from these still typically experience a drop in clean accuracy exceeding 10\% to date, and the state-of-the-art (SOTA) robustness has just improved slowly (\ie, about 1\% per year on average since 2018)~\citep{wei2023cfa,dong2023enemy,li2023data,jia2024revisiting}. This inherent trade-off \textbf{raises a question about the current AT paradigm} on uniformly learning accuracy and robustness:

\textit{Is the current AT paradigm, which compels DNNs to learn to classify the corresponding benign and adversarial samples into the same class, really appropriate and necessary for achieving robustness?}

\begin{figure*}[htbp]
    \centering
    \includegraphics[width=0.967\linewidth]{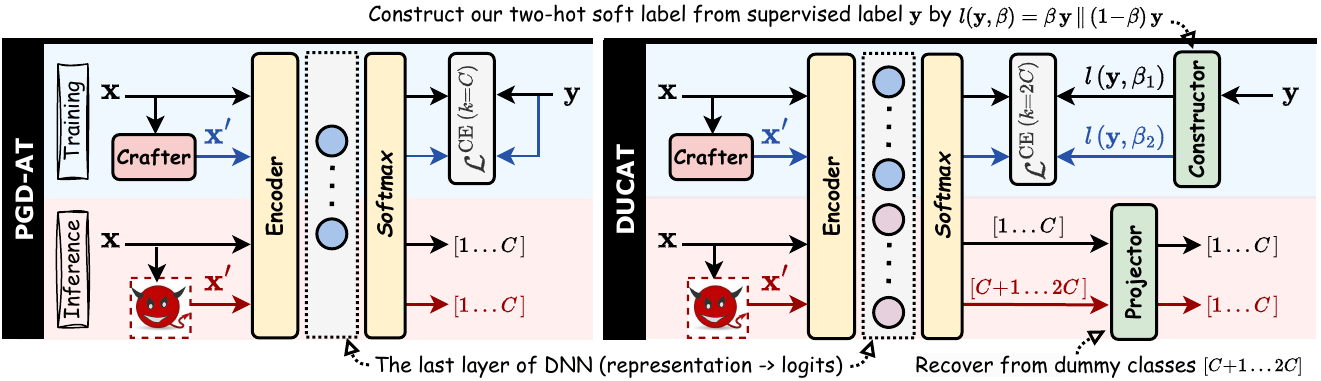}
    \caption{Comparison of the proposed DUCAT under \textbf{our new paradigm} of adversarial training and PGD-AT under the conventional paradigm. Currently, DNNs are adversarially trained to directly classify unseen inference-time $\mathbf{x}'$ to the same class as $\mathbf{x}$. In contrast, we suggest $C$ more \textbf{dummy classes} as in Figure~\ref{fig:intro}, along with a uniquely designed \textbf{two-hot soft label-based learning} to bridge them with original classes, so that DNNs can also assign $\mathbf{x}'$ to dummy classes and ensure robustness on them by inference-time recovery from predicted $[C\!+\!1\,...\,2C]$ to original $[1\,...\,C]$. This new paradigm relaxes the \textbf{overstrict} current assumption, releasing accuracy-robustness \textbf{trade-off} it causes.}
    \label{fig:roadmap}
\end{figure*}

In this work, as straightforward evidence supporting our deduction, we reveal that certain samples that \textbf{always fail} to meet the above objective of the current AT paradigm widely exist across various AT methods and different robust models. This observation implies that \textbf{the conventional assumption of categorizing adversarial samples into the same class as the corresponding benign ones may be overstrict}. As such, with the learning of benign and adversarial samples as two essentially inconsistent targets, the attempt to unify them in the current AT paradigm can be improper, which is likely to be blamed for the existing trade-off between clean accuracy and adversarial robustness. In response, we propose a new AT paradigm introducing additional dummy classes for certain adversarial samples that differ in distribution from the original ones to relax the current overstrict assumption. Then, we propose a novel AT method called \textbf{DU}mmy \textbf{C}lasses-based \textbf{A}dversarial \textbf{T}raining (\textbf{DUCAT}) accordingly, releasing the inherent trade-off between clean accuracy and robustness. 

Our core idea is to create dummy classes with the same number as the original ones and respectively attribute them as the primary targets for benign and adversarial samples during training. Importantly, we do not suggest a strict separation between corresponding benign and adversarial samples (\ie, to make the original classes completely benign and the dummy ones completely adversarial), because assuming corresponding benign and adversarial samples have completely different distributions is still excessive, which may cause memorization of hard-label training samples, resulting in overfitting to specific adversaries as shown by our toy case in Section~\ref{subsec:toy}. Instead, we construct unique two-hot soft labels to explicitly bridge corresponding original and dummy classes as the suboptimal alternative target of each other, so that the separation also becomes learnable, to utilize the potential of DNNs further. During inference time, such one-to-one correspondences enable the detection and recovery of adversarial samples, thereby achieving robustness without significantly compromising clean learning on benign samples. Specifically, if a test sample is classified into a dummy class, this serves as an indication of a potential adversarial attack, allowing us to recover its clean prediction through a projection back to the corresponding original class. It can then be viewed as the final prediction of the DUCAT-defended robust classifier. Separated from the computation graph of DNN and implemented in a run-time-only manner, such a projection further degrades the ability of real-world attackers.

\textbf{Contributions.} 
\textbf{1)} For the first time, we empirically reveal that \textbf{always-failed} samples widely exist in conventional AT, explicitly suggesting the assumption of the current AT paradigm to learn benign and adversarial samples with the same labels is \textbf{overstrict}; 
\textbf{2)} We propose a \textbf{new AT paradigm} introducing dummy classes to relax the current assumption, \textbf{releasing} the inherent accuracy-robustness trade-off caused by it, and correspondingly suggest a \textbf{new two-hot soft label-based learning strategy} for it;
\textbf{3)} A novel \textbf{plug-and-play DUCAT} method is proposed, concurrently improving the accuracy and robustness of four common AT benchmarks in large-scale experiments on CIFAR-10, CIFAR-100 and Tiny-ImageNet, significantly outperforming 18 existing SOTAs.

\section{Involving Dummy Classes in Adversarial Training}\label{sec:main}

\subsection{Motivation: Always-Failed Cases Widely Exist in Adversarial Training}\label{subsec:motivation}

The undesirable progress regarding the trade-off between clean accuracy and adversarial robustness in the past years makes us suspect that the assumption of the current AT paradigm to assign corresponding benign and adversarial samples to the same class may be inappropriate and unnecessary. As intuitive evidence of this deduction, by two proof-of-concept experiments, we demonstrate that up to 40\% adversarial samples that \textbf{always fail} to meet such an assumption generally exist in conventional AT crossing various AT methods and different robust models. The experiments are conducted on CIFAR-10~\citep{krizhevsky2009learning} and ResNet-18~\citep{he2016deep}, with, as mentioned in Section~\ref{sec:intro}, the most popular AT benchmarks, PGD-AT, TRADES, MART, and a representative SOTA, Consitency-AT. The training details are the same as our main experiments in Appendix~\ref{subapp:training}. For adversary, we adopt a typical and effective adversarial attack, PGD-10~\citep{madry2018towards}.

\begin{wrapfigure}[15]{r}{5.30cm}
\centering
\vspace{-1.3em}
\includegraphics[width=0.9\linewidth]{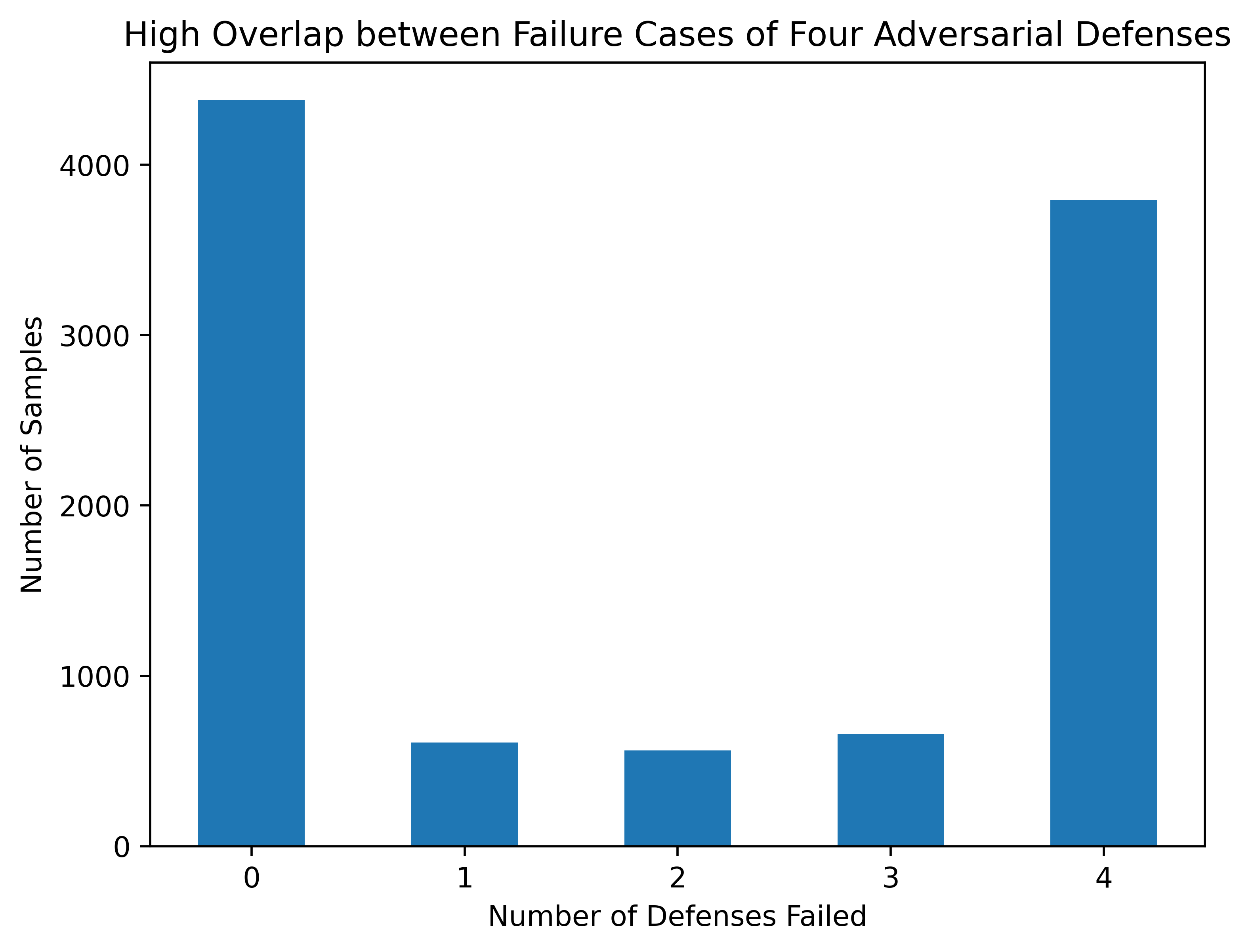}
\vspace{-0.2em}
\caption{High overlap between adversarial samples evading four AT benchmarks, implying such failures are more likely from an inappropriate learning objective under the current paradigm, rather than any specific AT methods.
}\label{fig:cross_methods_1}
\end{wrapfigure}

\paragraph{Always-Failed Cases Crossing AT Methods.}\label{para:always}
We first demonstrate that there is a high overlap in both successful and unsuccessful cases of the four AT methods. Specifically, for every benign test sample, with a high probability (\ie, $>$ 80\%), its corresponding adversarial sample can either be uniformly defended by the four methods or escape from all of them. As shown in Figure~\ref{fig:cross_methods_1}, we report the numbers of samples that can respectively beat \textbf{0}, \textbf{1}, \textbf{2}, \textbf{3}, \textbf{4} (all) experimental defenses. The samples on which all the defenses are uniformly worked or failed make up the overwhelming majority. Besides, the confusion matrixes of the models protected by these four AT methods under the adversary are shown in Figure~\ref{fig:cross_methods_2} in Appendix~\ref{app:comp}, where similar failure patterns can be noted crossing all of them. Based on these observations, we state that \textit{always-failed adversarial samples widely exist no matter the specific choice of AT methods}. 

\begin{figure}[htbp]
    \centering
    \includegraphics[height=3.5cm]{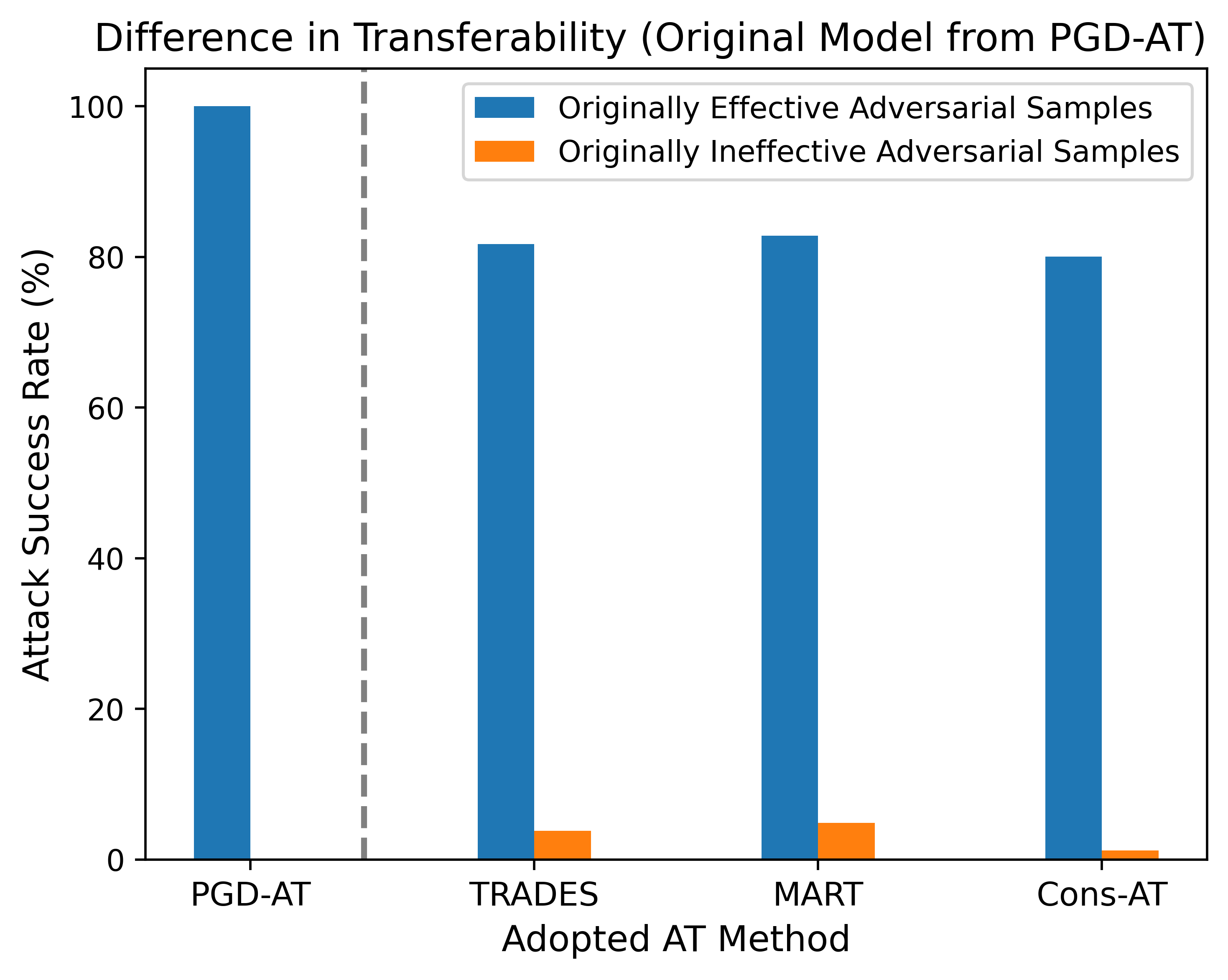}
    \includegraphics[height=3.5cm]{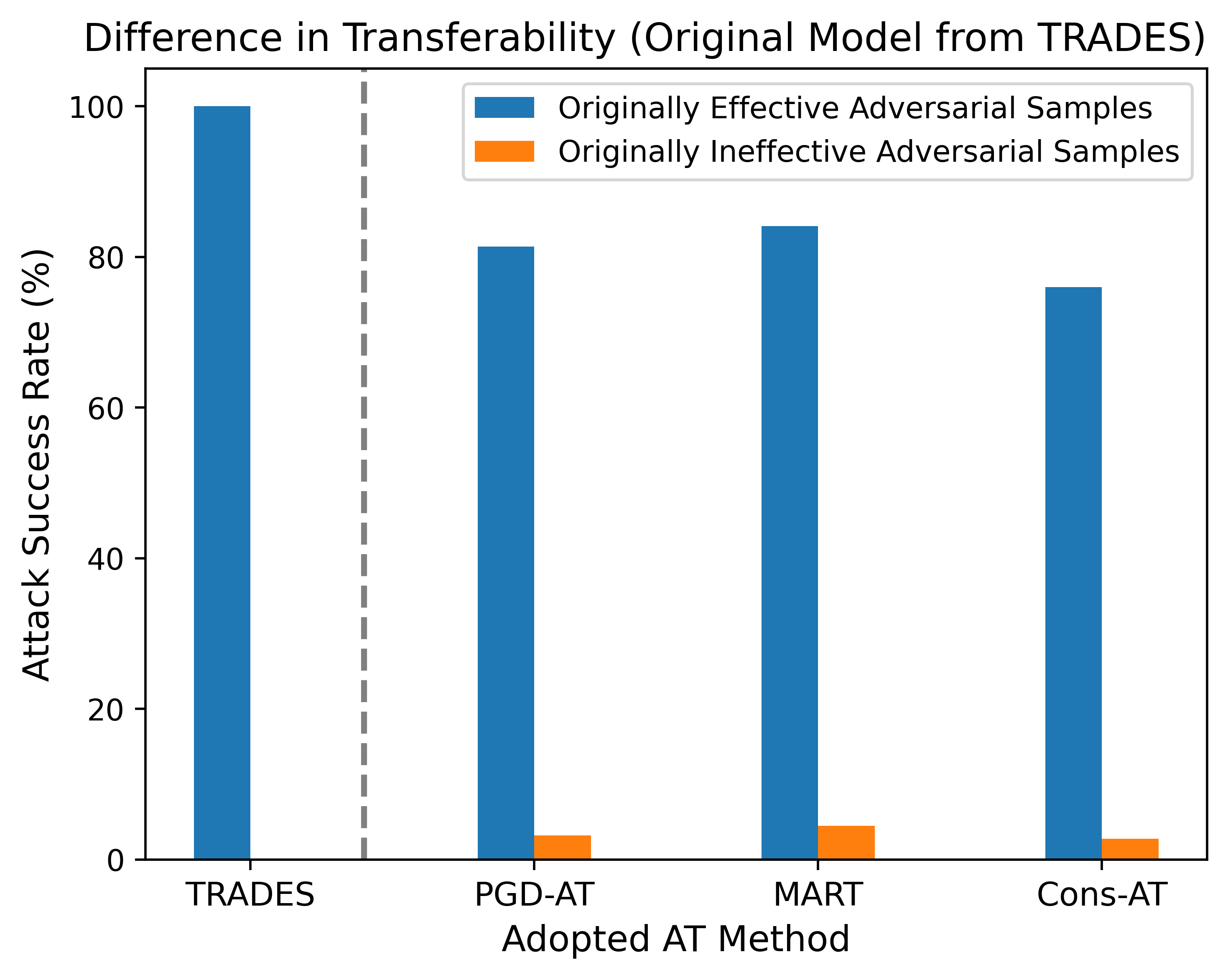}\\
    \includegraphics[height=3.5cm]{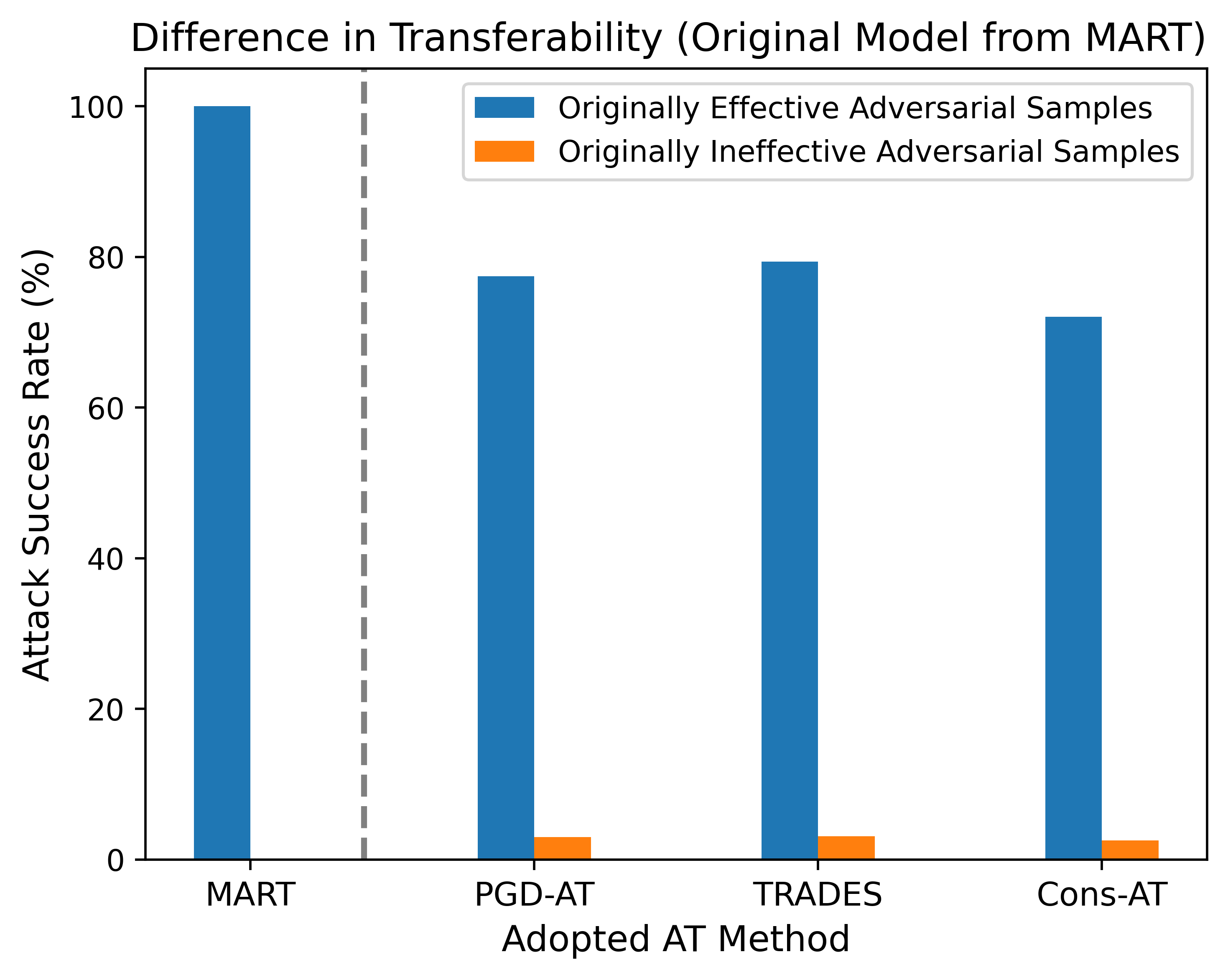}
    \includegraphics[height=3.5cm]{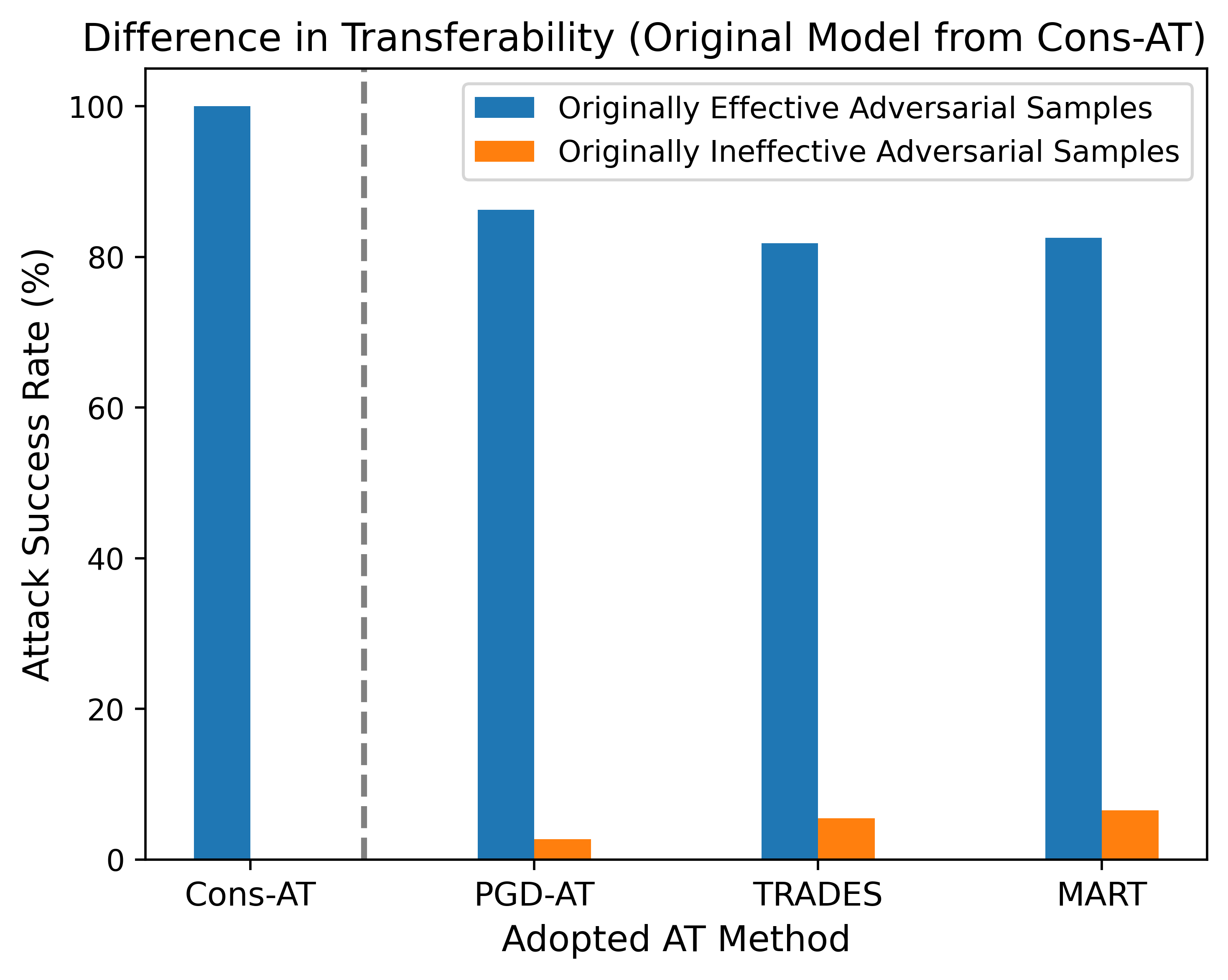}
    \caption{Robust models already enhanced by different AT methods are still highly likely to be uniformly beaten by the successful adversarial samples generated based on any one of these models. 
    Such a deconstruction of adversarial transferability between robust models reveals the model-independent vulnerability of certain samples, which further supports our deduction for the current AT paradigm.}
\label{fig:cross_models}
\end{figure}

\paragraph{Always-Failed Cases Crossing Robust Models.}
By respectively adopting the four AT benchmarks to train a model from scratch and then test it by the same white-box adversary, above we have revealed that the always-failed cases are independent of specific AT methods. In this part, through black-box transfer attacks, we further demonstrate that such cases should also not be attributed to the weakness of any specific robust models. To be specific, we first train four different robust models respectively by the benchmarks and select any one of them as the surrogate model, based on which we generate adversarial samples for the transfer attack. Then we divide the generated samples into two subsets based on whether successfully attack the surrogate model, and respectively use them to attack the other three models. Figure~\ref{fig:cross_models} shows the results with the four robust models serving as the surrogate one in order, where significant differences in transferability between the originally successful and unsuccessful subsets can be observed. This means for every single adversarial sample generated based on any robust model, if it is originally effective/ineffective in attacking this surrogate model, then with a high probability, it would also work/fail on other robust models. This implies that \textit{always-failed cases generally exist regardless of the specific robust model as the target}.

\begin{wrapfigure}[14]{r}{6.6cm}
\centering
\vspace{0.5em}
\includegraphics[width=\linewidth]{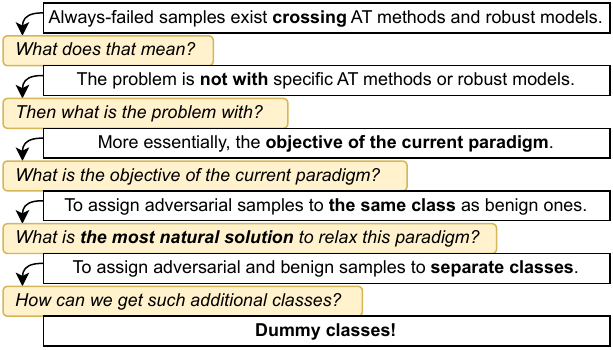}
\caption{The logic flow of our motivation to introduce dummy classes in the new AT paradigm.}\label{fig:logic}
\end{wrapfigure}

Now that the widespread existence of always-failed samples is confirmed crossing different AT methods and robust models, compared with simply attributing this to specific technology or implementation details, it seems more reasonable to rethink whether it is the paradigm to be blamed in essence. Based on the analysis in this section, it is not difficult to notice that \textit{always learning to classify the adversarial samples to the same class as the corresponding benign ones is likely to be a too-hard assumption}. In that case, trying to compulsively unify the learning of benign and adversarial samples, which may be two essentially inconsistent optimization objectives, becomes a logical and reasonable source of the existing trade-off between clean accuracy and adversarial robustness. The logic of our motivation is summed up with a flow chart as Figure~\ref{fig:logic}, which inspired us to propose a new AT paradigm in which the overstrict current assumption can be relaxed via dummy classes.

\subsection{New Paradigm of Adversarial Training with Dummy Classes}\label{sec:paradigm}

A natural idea to relax the current assumption is to introduce additional dummy classes to accommodate the adversarial samples that are hard to classify into the original class, based on which we propose our new AT paradigm. In this section, we first provide preliminaries \textit{w.r.t.} relevant concepts, followed by a formal formulation of our paradigm. Then, through a toy case deliberately causing overfitting to the training adversary, we demonstrate the necessity of designing two-hot soft labels rather than adopting conventional hard ones.

\subsubsection{Preliminaries}

In clean training, for a classification task with $C \geq 2$ as the number of classes, given a dataset $\mathcal{D} = \{(\mathbf{x}_i, y_i)\}_{i = 1, ..., n}$ with $\mathbf{x}_i \in \mathbb{R}^{d}$ and $y_i \in \{1, ..., C\}$ respectively denoting a natural sample and its supervised label, as well as $\mathbf{y}_i \in \mathbb{N}^{C}$ being the \textit{one-hot} format of $y_i$, ERM aims to learn a classifier $h_{\bm{\theta}} : \mathbf{x} \rightarrow \mathbb{R}^{C}$, such that $h_{\bm{\theta}}(\mathbf{x}_i) = \mathbf{q} (\mathbf{x}_i, \bm{\theta})$ represents the output \textit{logits} with each element $\mathbf{q}^{(k)} (\mathbf{x}_i, \bm{\theta})$ corresponding to class $k$, to minimize the empirical risk $\mathbb{E}_{\mathbf{x},\mathbf{y} \sim \mathcal{D}} \, [\mathcal{L} (h_{\bm{\theta}}(\mathbf{x}), \mathbf{y})]$ through a classification loss $\mathcal{L}$ like \textit{cross-entropy} (\textit{CE}). Such a learning objective is formulated as:
\begin{equation}
\begin{aligned}
\bm{\theta}^{*} = \argmin_{\bm{\theta}} \frac{1}{n} \sum_{i=1}^{n} \mathcal{L}(h_{\bm{\theta}}(\mathbf{x}_i), \mathbf{y}_i).
\end{aligned}
\end{equation}

AT augments ERM to $\mathbb{E}_{\mathbf{x}',\mathbf{y} \sim \mathcal{D}} \, [\max_{\, \mathbf{x}' \in \mathcal{P}(\mathbf{x})} \mathcal{L} (h_{\bm{\theta}}(\mathbf{x}'), \mathbf{y})]$, where $\mathbf{x}'$ denotes the adversarial sample \textit{w.r.t.} $\mathbf{x}$ and $\mathcal{P}(\mathbf{x})$ is a pre-defined perturbation set, to approximate the minimal empirical loss even under the strongest attack, directly learning the concept of robustness~\citep{madry2018towards}. Thus, the current AT objective can be reformulated as a \textit{min-max} problem:
\begin{equation}\label{eq:at}
\begin{aligned}
\bm{\theta}^{*} = \argmin_{\bm{\theta}} \frac{1}{n} \sum_{i=1}^{n} \max_{\, \mathbf{x}_i' \in \mathcal{P}(\mathbf{x}_i)} \mathcal{L}(h_{\bm{\theta}}(\mathbf{x}_i'), \mathbf{y}_i).
\end{aligned}
\end{equation}

On the other hand, to the best of our knowledge, the only previous work that involves the concept of dummy class is DuRM~\citep{wang2023frustratingly}, which aims to develop a general, simple and effective improvement to ERM for better generalization in various tasks. Specifically, in response to an assumption that ERM generalizes inadequately when the existence of outliers increases uncertainty and varies training and test landscapes~\citep{cha2021swad}, DuRM enlarges the dimension of output \textit{logits}, providing implicit supervision for existing classes and increasing the degree of freedom. Formally, given $(\cdot \| \cdot)$ denoting \textit{concat}, DuRM proposed to add $C_d$ dummy classes and transfer the original $C$-class classification task to a $(C + C_d)$-class classification problem:
\begin{equation}
\begin{aligned}
\bm{\theta}^{*} = \argmin_{\bm{\theta}} \frac{1}{n} \sum_{i=1}^{n} \, \mathcal{L} \, \big( \, h_{\bm{\theta}}(\mathbf{x}_{i}) \, \| \, h_{\bm{\theta}}^{\text{DuRM}}(\mathbf{x}_{i}), \, \mathbf{y}_{i} \, \| \, \mathbf{0} \, \big),
\end{aligned}
\end{equation}
where $h_{\bm{\theta}}^{\text{DuRM}}(\mathbf{x}_{i}) \in \mathbb{R}^{C_d}$ denotes the output \textit{logits} of DuRM and $\mathbf{0}$ is a zero vector in the same dimension, identifying no supervision information for these additional dummy classes. As discussed in Appendix~\ref{subapp:albeit}, although inheriting the term ``dummy'', \textbf{our work is different in essence from DuRM} regarding motivation, significance and achievement. Especially, DuRM expects \textbf{no samples} classified to dummy classes, while ours is just \textbf{the opposite} as to be seen in Section~\ref{subsubsec:formulation}. Thus, \textbf{the specific significances behind ``dummy'' are completely different between DuRM and ours}.

\subsubsection{Paradigm Formulation}\label{subsubsec:formulation}

Provided the statement in Section~\ref{subsec:motivation} that the current assumption in conventional AT paradigm to learn adversarially perturbed samples with the same supervised label as the corresponding benign ones is likely to be a too-hard optimization objective, we define a novel AT paradigm to relax this assumption. Formally, for a $C$-class classification task, we propose to append another $C$ dummy classes to build \textbf{one-to-one correspondence} between original classes and the dummy ones. Then to optimize this new $2 \cdot C$-class classification problem, we introduce dummy label $\dot{\mathbf{y}}_{i} \, \| \, \ddot{\mathbf{y}}_{i}$ such that:
\begin{equation}
    \begin{aligned}
    \dot{\mathbf{y}}_{i}^{(k)} + \ddot{\mathbf{y}}_{i}^{(k)} = 
    \begin{cases}
        1, & k = y_i\\
        0, & k \neq y_i\\
    \end{cases},
    \end{aligned}
\end{equation}
where $\dot{\mathbf{y}}_{i} + \ddot{\mathbf{y}}_{i}$ equals to the original \textit{one-hot} label vector $\mathbf{y}_{i}$. With $\mathbf{x}'$ being adversarial sample from $\mathbf{x}$ and $h_{\bm{\theta}}^{\text{Dummy}}(\mathbf{x}_{i}) \in \mathbb{R}^{C}$ denoting the output \textit{logits} from the $C$ appended dummy classes, our new AT paradigm can be formulated as:
\begin{equation}\label{eq:para}
{\small
\begin{aligned}
    \bm{\theta}^{*} = \argmin_{\bm{\theta}} \frac{1}{n} \sum\limits_{i=1}\limits^{n} \Big( \mathcal{L} \big( h_{\bm{\theta}}(\mathbf{x}_{i}) \, \| \, h_{\bm{\theta}}^{\text{Dummy}}(\mathbf{x}_{i}), \, \dot{\mathbf{y}}_{i} \, \| \, \ddot{\mathbf{y}}_{i} \big) + \! \max\limits_{\mathbf{x}_i' \in \mathcal{P}(\mathbf{x}_i)} \! \mathcal{L} \big( h_{\bm{\theta}}(\mathbf{x}_{i}') \, \| \, h_{\bm{\theta}}^{\text{Dummy}}(\mathbf{x}_{i}'), \, \dot{\mathbf{y}}_{i}' \, \| \, \ddot{\mathbf{y}}_{i}' \big) \Big),
\end{aligned}
}
\end{equation}
where $\dot{\mathbf{y}}_{i}'$ and $\ddot{\mathbf{y}}_{i}'$ can be differently weighted with $\dot{\mathbf{y}}_{i}$ and $\ddot{\mathbf{y}}_{i}$. For example, it would be valid to assign $\dot{\mathbf{y}}_{i}^{(k)}$ and $\ddot{\mathbf{y}}_{i}^{(k)}$ as 0.5, while assign $\dot{\mathbf{y}}_{i}'^{(k)}$ and $\ddot{\mathbf{y}}_{i}'^{(k)}$ respectively as 1 and 0, both of which add up to 1.

At inference time, we acquire the final output of the robust DNN classifier by projecting each dummy class to the corresponding original ones. As such a projection is separated from the computation graph of DNN and can be merely implemented in the run time, the robust mechanism can be simply explained as detecting (possible) adversarial attacks when certain samples are classified to the dummy classes and recovering such samples back to the corresponding original classes. We defer detailed discussion on the threat model including adversary capacity to Appendix~\ref{subapp:threat}. With $\hat{y}$ denoting the final predicted class label, the projection is formulated as:
\begin{equation}\label{eq:proj}
    \begin{aligned}
    \hat{y} = 
    \begin{cases}
        g_{\bm{\theta}}(\mathbf{x}_{i}), \!& g_{\bm{\theta}}(\mathbf{x}_{i}) \!\leq C\\
        g_{\bm{\theta}}(\mathbf{x}_{i}) \!- C, \!& g_{\bm{\theta}}(\mathbf{x}_{i}) \!> C\\
    \end{cases},
    \hspace{0.5em} \text{where} \hspace{0.5em}
    g_{\bm{\theta}}(\mathbf{x}_{i}) = \argmax_{k = 1, ..., 2 \cdot C} \big( h_{\bm{\theta}}(\mathbf{x}_{i}) \, \| \, h_{\bm{\theta}}^{\text{Dummy}}(\mathbf{x}_{i}) \big)^{(k)}.
    \end{aligned}
\end{equation}

Our new paradigm explicitly distinguishes the learning of benign and adversarial samples, relaxing the overstrict assumption in the current AT paradigm that always pursues consistent class distribution for them. This is expected to release the unnecessary tension between the standard and robust optimization objectives, and as a result, release the trade-off between clean accuracy and robustness observed in existing AT methods. Notably, we do \textbf{not} simply separate benign and adversarial samples into completely different classes (\ie, respectively adopting $\mathbf{y}_{i} \, \| \, \mathbf{0}$, and $\mathbf{0} \, \| \, \mathbf{y}_{i}$ as their new supervised labels). Instead, we propose to construct novel two-hot soft labels, so that the separation also becomes a certain learnable pattern within the optimization process, to further utilize the potential of DNNs. We demonstrate the significance of this novel idea by a case study in the following Section~\ref{subsec:toy}.

\subsubsection{Two-Hot Soft Label Matters}\label{subsec:toy}

Within the new AT paradigm, apart from the dummy class, \textbf{another novel concept proposed is the two-hot soft label}. In this section, we demonstrate that abusing the new AT paradigm with hard labels rather than two-hot soft labels would lead to a generalization problem from training to test adversaries, which confirms the necessity and contribution of this design. Specifically, here we directly assign that $\dot{\mathbf{y}}_{i} = \ddot{\mathbf{y}}_{i}' = \mathbf{y}$ and $\ddot{\mathbf{y}}_{i} = \dot{\mathbf{y}}_{i}' = \mathbf{0}$, which means there is no supervision information to explicitly bridge corresponding benign and adversarial samples, and they are viewed to belong to absolutely different classes. We implement such a toy case with the same settings as the proof-of-concept experiments in Section~\ref{subsec:motivation}, except additionally introducing Auto-Attack (AA)~\citep{croce2020reliable} to evaluate

\begin{wrapfigure}[17]{r}{4.7cm}
\centering
\includegraphics[width=0.88\linewidth]{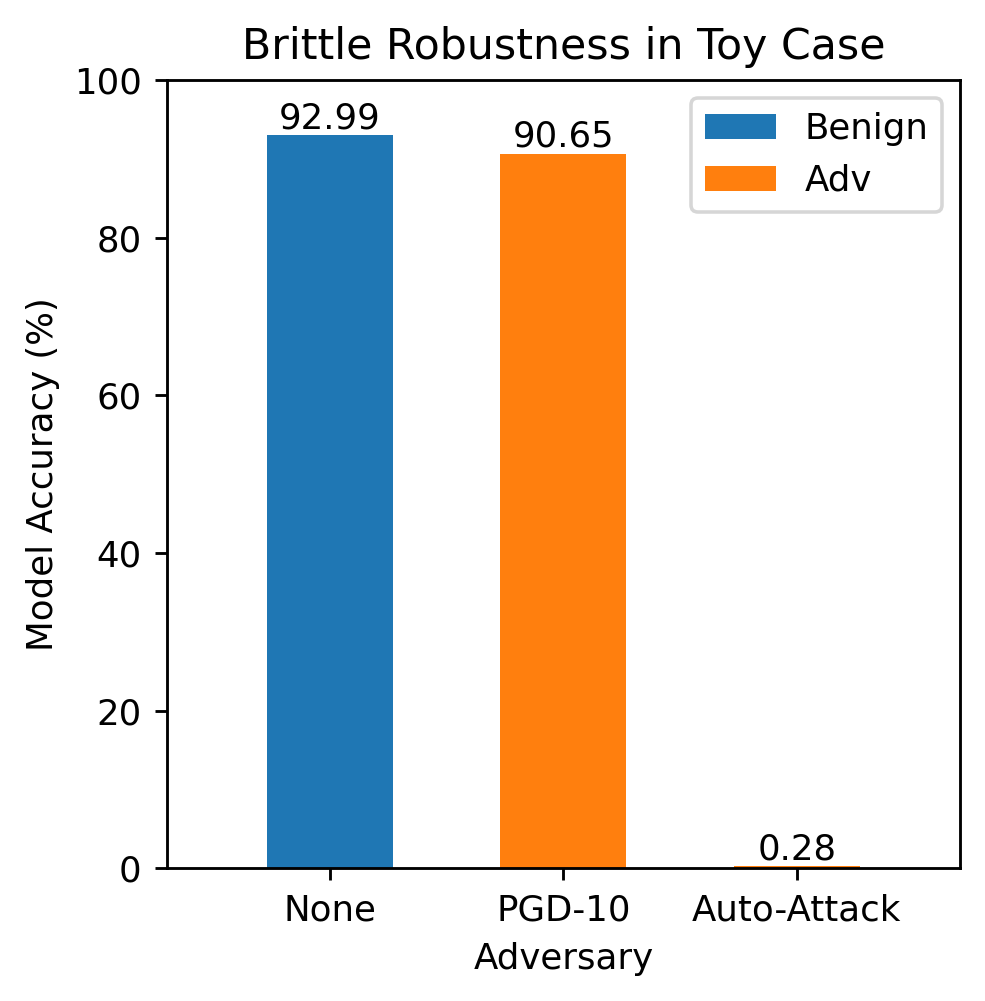}
\vspace{-0.5em}
\caption{A toy case using hard labels achieves brittle robustness. The result is surprisingly high under training adversary PGD-10 but collapses under unseen AA.}\label{fig:toy}
\end{wrapfigure}

the robust generalization. As shown in Figure~\ref{fig:toy}, our result demonstrates that although this toy case can achieve surprisingly high accuracy both on clean data and under PGD-10 adversary, it immediately collapses once facing AA, which reveals that it fails to learn the real robustness that can generalize to unseen adversaries. On the contrary, it simply overfits the PGD-10 training samples. This toy case suggests that the most straightforward idea to completely separate the learning of benign and adversarial samples may be undesirable, implying the importance of building generalized correspondence between them. This not only supports the design of two-hot soft labels in our paradigm but also provides empirical guidance for our specific AT method proposed in the following Section~\ref{subsec:DUCAT}. Besides, please kindly note that our two-hot soft labels are \textbf{significantly different from the conventional soft labels}, such as the ones in \citet{muller2019does}, \citet{shafahi2019label}, and \citet{wu2024annealing}. We defer the detailed discussion to Appendix~\ref{subapp:two-hot}.

\subsection{Proposed Method: Dummy Classes-based Adversarial Training}\label{subsec:DUCAT}

Following our new AT paradigm, we specifically propose a novel AT method named \textbf{DU}mmy \textbf{C}lasses-based \textbf{A}dversarial \textbf{T}raining (\textbf{DUCAT}). Formally, for the target model $\bm{\theta}$ and certain benign sample $\mathbf{x}_{i}$, given $\mathbf{\hat{x}}_i'$ representing the adversarial sample generated through $\mathbf{\hat{x}}_i' =$ $\argmax_{\, \mathbf{x}_i' \in \mathcal{P}(\mathbf{x}_i)}$ $\mathds{1}(\argmax_{k = 1, ..., C} h_{\bm{\theta}}(\mathbf{x}_i')^{(k)} \neq y_i)$, then based on Equations~(\ref{eq:para}) and (\ref{eq:proj}), our new empirical risk to be minimized can be formulated with \textit{0-1 loss}~\citep{zhang2019theoretically} and \textit{indicator function} $\mathds{1}(\cdot)$ as:
\begin{equation}\label{eq:risk}
\begin{aligned}
    \mathcal{R}^{\text{DUCAT}} (\bm{\theta}, \mathbf{x}_{i}) := \,\, & \alpha \cdot \big( \beta_1 \cdot \mathds{1} (g_{\bm{\theta}}(\mathbf{x}_{i}) \neq y_i) + (1 - \beta_1) \cdot \mathds{1} (g_{\bm{\theta}}(\mathbf{x}_{i}) \neq (y_i + C) ) \big) \\
    & + (1 - \alpha) \cdot \big( \beta_2 \cdot \mathds{1} ( g_{\bm{\theta}}(\mathbf{x}_{i}') \neq (y_i + C) ) + (1 - \beta_2) \cdot \mathds{1} ( g_{\bm{\theta}}(\mathbf{x}_{i}') \neq y_i) \big),
\end{aligned}
\end{equation}
where $\alpha$, $\beta_1$, $\beta_2$ are hyper-parameters, with discussion on their specific values deferred to Appendix~\ref{app:ablation}.

Such a deconstruction of our target risk is expected to facilitate the understanding of its principle and novelty. Specifically, existing AT methods either only focus on the adversarial risk $\mathds{1} ( \, h_{\bm{\theta}}(\mathbf{x}_{i}') \neq y_i)$ or jointly considering the benign ones like $\mathds{1} ( \, h_{\bm{\theta}}(\mathbf{x}_{i}) \neq y_i) + \mathds{1} ( \, h_{\bm{\theta}}(\mathbf{x}_{i}') \neq y_i)$~\citep{bai2021recent}. Our toy case in Appendix~\ref{subsec:toy} attributes benign and adversarial samples to completely irrelevant risks as $\mathds{1} ( \, h_{\bm{\theta}}(\mathbf{x}_{i}) \neq y_i) + \mathds{1} ( \, h_{\bm{\theta}}(\mathbf{x}_{i}') \neq (y_i \!+\! C))$, which is also confirmed undesirable. In contrast, our proposed risk expects the benign and adversarial samples with the same original label to be classified as an original class and a dummy one. The important difference is that \textbf{neither} the original class is necessary to be completely benign \textbf{nor} the dummy one is expected totally adversarial. 

The simple principle behind this idea is that assuming corresponding benign and adversarial samples to have completely different distributions is as unadvisable as assuming the same distributions between them. Instead, we utilize the weights $\beta_1$ and $\beta_2$ to inject our preference in a softer manner, suggesting that when learning the primary target label of a benign sample $\mathbf{x}_{i}$ (or adversarial sample $\mathbf{x}_{i}'$) is found too hard for DNNs, automatically shunting them to the corresponding dummy (or benign) class is a more acceptable alternative option compared with other completely irrelevant classes. This approach is expected to not only reduce the overfitting to specific adversaries caused by the memorization of hard-label training adversarial samples~\citep{stutz2020confidence,dong2022exploring,cheng2022cat}, but also establish explicit connections between corresponding benign and adversarial samples, laying a solid foundation for the inference time projection from dummy classes to the corresponding original ones as provided in Equation~(\ref{eq:proj}).

Finally, while \textit{0-1 loss} benefits conceptual analysis from the perspective of the risk, the optimization directly over it can be computationally intractable. So to end up with a real-world practical AT method, we introduce \textit{CE}, the most commonly used for both supervised classification and conventional AT, as the surrogate loss to optimize the proposed risk $\mathcal{R}^{\text{DUCAT}}$ in Equation~(\ref{eq:risk}). It is also feasible to adopt more advanced loss functions such as \textit{boosted CE}~\citep{wang2020improving}, which creates conditions for integrating the proposed DUCAT with conventional AT methods. Formally, provided that $l \, (\mathbf{y}_i, \beta)$ denotes the two-hot soft label constructed as the supervision signal, such that:
\begin{equation}
\begin{aligned}
    l \, (\mathbf{y}_i, \beta) = (\beta \cdot \mathbf{y}_i) \, \| \, ((1 - \beta) \cdot \mathbf{y}_i).
\end{aligned}
\end{equation}
The \textbf{overall objective} of DUCAT can be formulated as:
\begin{equation}
{\small
\begin{aligned}
    \mathcal{L}^{\text{DUCAT}}(\bm{\theta}, \mathcal{D}, \alpha, \beta_1, \beta_2) := \frac{1}{n} \sum_{i=1}^{n} \Big( \alpha \cdot \mathcal{L}^{\text{surro}} \big(\mathbf{x}_i, l(\mathbf{y}_i, \beta_1)\big) + (1 - \alpha) \cdot \mathcal{L}^{\text{surro}} \big(\mathbf{\hat{x}}_i', l(\mathbf{y}_i, (1 - \beta_2))\big) \Big).
\end{aligned}
}
\end{equation}
Then with $\mathcal{L}^{\text{CE}}$ serving as the surrogate loss $\mathcal{L}^{\text{surro}}$, we have:
\begin{equation}
{\small
\begin{aligned}
    \mathcal{L}^{\text{DUCAT}}(\bm{\theta}, \mathcal{D}, \alpha, \beta_1, \beta_2) = \,\, & \frac{1}{n} \sum_{i=1}^{n} \Big( - \alpha \cdot \big( \beta_1 \cdot \log (\mathbf{p}^{(y_i)} (\mathbf{x}_i, \bm{\theta})) + (1 - \beta_1) \cdot \log (\mathbf{p}^{(y_i + C)} (\mathbf{x}_i, \bm{\theta})) \big) \\
    & \hspace{0.1em} - (1 - \alpha) \cdot \big( \beta_2 \cdot \log (\mathbf{p}^{(y_i + C)} (\mathbf{\hat{x}}_i', \bm{\theta})) + (1 - \beta_2) \cdot \log (\mathbf{p}^{(y_i)} (\mathbf{\hat{x}}_i', \bm{\theta})) \big) \Big),
\end{aligned}
}
\end{equation}
in which with $\sigma : \mathbb{R}^{2C} \!\rightarrow\! (0, 1)^{2C}$ representing the \textit{softmax}, $\mathbf{p} (\mathbf{x}_i, \bm{\theta})$ denotes the output probabilistic prediction from DUCAT (\ie, with the length of $2C$ and the components adding up to one), such that: 
\begin{equation}
\begin{aligned}
\mathbf{p} (\mathbf{x}_i, \bm{\theta}) = \sigma \big( h_{\bm{\theta}}(\mathbf{x}_{i}) \,\|\, h_{\bm{\theta}}^{\text{Dummy}}(\mathbf{x}_{i}) \big) = \sigma \big( \mathbf{q} (\mathbf{x}_i, \bm{\theta}) \,\|\, \mathbf{q}^{\text{Dummy}}(\mathbf{x}_i, \bm{\theta}) \big).
\end{aligned}
\end{equation}

\begin{table*}[htb]
  \caption{The results of integrating the proposed DUCAT to four AT benchmarks on CIFAR-10, CIFAR-100 and Tiny-ImageNet with ResNet-18 under $\ell_{\infty}$ adversaries, clearly releasing the inherent trade-off between clean accuracy and robustness. All results are acquired from three runs.}
  \label{tab:resnet18}
  \centering
  \setlength{\tabcolsep}{3pt}
  \renewcommand\arraystretch{1.4}
  \resizebox{13.9cm}{!}{
  \begin{tabular}{clllllll}
    \toprule
    \multicolumn{1}{c}{\textbf{Dataset}} & \multicolumn{1}{c}{\textbf{Method}} & \multicolumn{1}{c}{\hspace{-1em} \textbf{Clean}} & \multicolumn{1}{c}{\hspace{-1em} \textbf{PGD-10}} & \multicolumn{1}{c}{\hspace{-1em} \textbf{PGD-100}} & \multicolumn{1}{c}{\hspace{-1em} \textbf{AA}} & \multicolumn{1}{c}{\hspace{-1em} \textbf{Mean}} & \multicolumn{1}{c}{\hspace{-1em} \textbf{NRR}} \\
    \midrule
    & \hspace{0.3em} PGD-AT
      & 82.92
      & 51.81
      & 50.34
      & 46.74
      & 64.830
      & 59.782 \\
    & \hspace{1.3em} \textbf{+ DUCAT} \hspace{0.4em}
      & 88.81 $^{\color{ForestGreen}{(+7.1\%)}}$
      & 65.10 $^{\color{ForestGreen}{(+25.7\%)}}$
      & 62.71 $^{\color{ForestGreen}{(+24.6\%)}}$
      & 58.61 $^{\color{ForestGreen}{(+25.4\%)}}$
      & 73.710 $^{\color{ForestGreen}{(+13.7\%)}}$
      & 70.617 $^{\color{ForestGreen}{(+18.1\%)}}$ \\
    \cmidrule(r){2-8}
    & \hspace{0.3em} TRADES
      & 79.67 
      & 52.14 
      & 51.88 
      & 47.62  
      & 63.645 
      & 59.610 \\
    & \hspace{1.3em} \textbf{+ DUCAT} \hspace{0.4em}
      & 77.74 $^{\color{Salmon}{(-2.4\%)}}$
      & 52.66 $^{\color{ForestGreen}{(+1.0\%)}}$
      & 51.98 $^{\color{ForestGreen}{(+0.2\%)}}$
      & 48.17 $^{\color{ForestGreen}{(+1.2\%)}}$
      & 62.955 $^{\color{Salmon}{(-1.1\%)}}$
      & 59.483 $^{\color{Salmon}{(-0.2\%)}}$ \\
    \cmidrule(r){2-8}
    \multirow{2}{*}{CIFAR-10}
    & \hspace{0.3em} MART
      & 77.93  
      & 53.61  
      & 52.83  
      & 46.70  
      & 62.315 
      & 58.402 \\
    & \hspace{1.3em} \textbf{+ DUCAT} \hspace{0.4em}
      & 80.65 $^{\color{ForestGreen}{(+3.5\%)}}$
      & 58.42 $^{\color{ForestGreen}{(+9.0\%)}}$
      & 57.81 $^{\color{ForestGreen}{(+9.4\%)}}$
      & 50.18 $^{\color{ForestGreen}{(+7.5\%)}}$
      & 65.415 $^{\color{ForestGreen}{(+5.0\%)}}$
      & 61.867 $^{\color{ForestGreen}{(+5.9\%)}}$ \\
    \cmidrule(r){2-8}
    & \hspace{0.3em} Cons-AT
      & 83.42  
      & 53.20  
      & 51.68  
      & 47.72  
      & 65.570 
      & 60.711  \\
    & \hspace{1.3em} \textbf{+ DUCAT} \hspace{0.4em}
      & 89.51 $^{\color{ForestGreen}{(+7.3\%)}}$
      & 66.83 $^{\color{ForestGreen}{(+25.6\%)}}$
      & 63.80 $^{\color{ForestGreen}{(+23.5\%)}}$
      & 57.18 $^{\color{ForestGreen}{(+19.8\%)}}$
      & 73.345 $^{\color{ForestGreen}{(+11.8\%)}}$
      & 69.782 $^{\color{ForestGreen}{(+14.9\%)}}$ \\
    \cmidrule(r){2-8}
    & \hspace{0.3em} Average
      & 80.99  
      & 52.69  
      & 51.68  
      & 47.20  
      & 64.090 
      & 59.626 \\
    & \hspace{1.3em} \textbf{+ DUCAT} \hspace{0.4em}
      & 84.18 $^{\color{ForestGreen}{(+3.9\%)}}$
      & 60.75 $^{\color{ForestGreen}{(+15.3\%)}}$
      & 59.08 $^{\color{ForestGreen}{(+14.3\%)}}$
      & 53.54 $^{\color{ForestGreen}{(+13.4\%)}}$
      & 68.856 $^{\color{ForestGreen}{(+7.4\%)}}$
      & 65.437 $^{\color{ForestGreen}{(+9.7\%)}}$ \\
    \midrule
    & \hspace{0.3em} PGD-AT
      & 56.56  
      & 29.27  
      & 28.71  
      & 25.02  
      & 40.790 
      & 34.693  \\
    & \hspace{1.3em} \textbf{+ DUCAT} \hspace{0.4em}
      & 70.71 $^{\color{ForestGreen}{(+25.0\%)}}$
      & 33.17 $^{\color{ForestGreen}{(+13.3\%)}}$
      & 29.56 $^{\color{ForestGreen}{(+3.0\%)}}$
      & 25.20 $^{\color{ForestGreen}{(+0.7\%)}}$
      & 47.955 $^{\color{ForestGreen}{(+17.6\%)}}$
      & 37.158 $^{\color{ForestGreen}{(+7.1\%)}}$ \\
    \cmidrule(r){2-8}
    & \hspace{0.3em} TRADES
      & 55.39  
      & 29.61  
      & 29.28  
      & 24.51  
      & 39.950 
      & 33.983  \\
    & \hspace{1.3em} \textbf{+ DUCAT} \hspace{0.4em}
      & 55.41 $^{\color{Gray}{(+0.0\%)}}$
      & 30.69 $^{\color{ForestGreen}{(+3.6\%)}}$
      & 30.38 $^{\color{ForestGreen}{(+3.8\%)}}$
      & 25.23 $^{\color{ForestGreen}{(+2.9\%)}}$
      & 40.320 $^{\color{ForestGreen}{(+0.9\%)}}$
      & 34.672 $^{\color{ForestGreen}{(+2.0\%)}}$ \\
    \cmidrule(r){2-8}
    \multirow{2}{*}{CIFAR-100}
    & \hspace{0.3em} MART
      & 49.83  
      & 30.60  
      & 30.31  
      & 25.00  
      & 37.415 
      & 33.295  \\
    & \hspace{1.3em} \textbf{+ DUCAT} \hspace{0.4em}
      & 56.73 $^{\color{ForestGreen}{(+13.8\%)}}$
      & 41.78 $^{\color{ForestGreen}{(+36.5\%)}}$
      & 34.32 $^{\color{ForestGreen}{(+13.2\%)}}$
      & 27.44 $^{\color{ForestGreen}{(+9.8\%)}}$
      & 42.085 $^{\color{ForestGreen}{(+12.5\%)}}$
      & 36.989 $^{\color{ForestGreen}{(+11.1\%)}}$ \\
    \cmidrule(r){2-8}
    & \hspace{0.3em} Cons-AT
      & 58.53  
      & 29.99  
      & 29.13  
      & 25.39  
      & 41.960 
      & 35.417  \\
    & \hspace{1.3em} \textbf{+ DUCAT} \hspace{0.4em}
      & 72.29 $^{\color{ForestGreen}{(+23.5\%)}}$
      & 33.98 $^{\color{ForestGreen}{(+13.3\%)}}$
      & 30.73 $^{\color{ForestGreen}{(+5.5\%)}}$
      & 25.66 $^{\color{ForestGreen}{(+1.1\%)}}$
      & 48.975 $^{\color{ForestGreen}{(+16.7\%)}}$
      & 37.876 $^{\color{ForestGreen}{(+6.9\%)}}$ \\
    \cmidrule(r){2-8}
    & \hspace{0.3em} Average
      & 55.08  
      & 29.87  
      & 29.36  
      & 24.98  
      & 40.029 
      & 34.347  \\
    & \hspace{1.3em} \textbf{+ DUCAT} \hspace{0.4em}
      & 63.79 $^{\color{ForestGreen}{(+15.8\%)}}$
      & 34.91 $^{\color{ForestGreen}{(+16.9\%)}}$
      & 31.25 $^{\color{ForestGreen}{(+6.4\%)}}$
      & 25.88 $^{\color{ForestGreen}{(+3.6\%)}}$
      & 44.834 $^{\color{ForestGreen}{(+12.0\%)}}$
      & 36.674 $^{\color{ForestGreen}{(+6.8\%)}}$ \\
    \midrule
    & \hspace{0.3em} PGD-AT
      & 46.32  
      & 21.75  
      & 21.52  
      & 17.07  
      & 31.695 
      & 24.947  \\
    & \hspace{1.3em} \textbf{+ DUCAT} \hspace{0.4em}
      & 56.18 $^{\color{ForestGreen}{(+21.3\%)}}$
      & 24.23 $^{\color{ForestGreen}{(+11.4\%)}}$
      & 22.54 $^{\color{ForestGreen}{(+4.7\%)}}$
      & 18.68 $^{\color{ForestGreen}{(+9.4\%)}}$
      & 37.430 $^{\color{ForestGreen}{(+18.1\%)}}$
      & 28.037 $^{\color{ForestGreen}{(+12.4\%)}}$ \\
    \cmidrule(r){2-8}
    & \hspace{0.3em} TRADES
      & 46.75  
      & 21.62  
      & 21.52  
      & 16.60  
      & 31.675 
      & 24.500  \\
    & \hspace{1.3em} \textbf{+ DUCAT} \hspace{0.4em}
      & 46.90 $^{\color{ForestGreen}{(+0.3\%)}}$
      & 22.38 $^{\color{ForestGreen}{(+3.5\%)}}$
      & 22.08 $^{\color{ForestGreen}{(+2.6\%)}}$
      & 17.27 $^{\color{ForestGreen}{(+4.0\%)}}$
      & 32.085 $^{\color{ForestGreen}{(+1.3\%)}}$
      & 25.244 $^{\color{ForestGreen}{(+3.0\%)}}$ \\
    \cmidrule(r){2-8}
    \hspace{0.1em} \multirow{2}{*}{Tiny-ImageNet} \hspace{0.1em}
    & \hspace{0.3em} MART
      & 39.70  
      & 22.98  
      & 22.79  
      & 17.18  
      & 28.440 
      & 23.982  \\
    & \hspace{1.3em} \textbf{+ DUCAT} \hspace{0.4em}
      & 43.37 $^{\color{ForestGreen}{(+9.2\%)}}$
      & 25.23 $^{\color{ForestGreen}{(+9.8\%)}}$
      & 25.68 $^{\color{ForestGreen}{(+12.7\%)}}$
      & 18.41 $^{\color{ForestGreen}{(+7.2\%)}}$
      & 30.890 $^{\color{ForestGreen}{(+8.6\%)}}$
      & 25.848 $^{\color{ForestGreen}{(+7.8\%)}}$ \\
    \cmidrule(r){2-8}
    & \hspace{0.3em} Cons-AT
      & 46.54  
      & 22.58  
      & 21.70  
      & 17.60  
      & 32.070 
      & 25.541 \\
    & \hspace{1.3em} \textbf{+ DUCAT} \hspace{0.4em}
      & 59.64 $^{\color{ForestGreen}{(+28.1\%)}}$
      & 25.40 $^{\color{ForestGreen}{(+12.5\%)}}$
      & 23.22 $^{\color{ForestGreen}{(+7.0\%)}}$
      & 18.92 $^{\color{ForestGreen}{(+7.5\%)}}$
      & 39.280 $^{\color{ForestGreen}{(+22.5\%)}}$
      & 28.727 $^{\color{ForestGreen}{(+12.5\%)}}$ \\
    \cmidrule(r){2-8}
    & \hspace{0.3em} Average
      & 44.83  
      & 22.23  
      & 21.88  
      & 17.11  
      & 30.970 
      & 24.743 \\
    & \hspace{1.3em} \textbf{+ DUCAT} \hspace{0.4em}
      & 51.52 $^{\color{ForestGreen}{(+14.9\%)}}$
      & 24.31 $^{\color{ForestGreen}{(+9.3\%)}}$
      & 23.38 $^{\color{ForestGreen}{(+6.8\%)}}$
      & 18.32 $^{\color{ForestGreen}{(+7.1\%)}}$
      & 34.921 $^{\color{ForestGreen}{(+12.8\%)}}$
      & 26.964 $^{\color{ForestGreen}{(+9.0\%)}}$ \\
    \bottomrule
  \end{tabular}
  }
\end{table*}

\section{Experiments}\label{sec:exp}

Following previous AT works, the evaluation of the proposed DUCAT is conducted on CIFAR-10, CIFAR-100~\citep{krizhevsky2009learning} and Tiny-ImageNet~\citep{li2015tiny} datasets with ResNet-18~\citep{he2016deep} and WideResNet-28-10~\citep{zagoruyko2016wide} architectures. The three most commonly used AT baselines, namely PGD-AT, TRADES, MART, and an advanced method, Cons-AT, are adopted as experimental benchmarks. Additionally, we involve 18 related SOTAs specifically for the trade-off problem in our comparison by simply adopting their advances reported in the original papers. These AT methods are respectively detailed in Appendix~\ref{app:details_at1} and \ref{app:details_at2}. 

There are more details supplemented in our Appendix. An example algorithm for PGD-AT + DUCAT is provided in Appendix~\ref{app:algo}. The training details and threat models are deferred to Appendix~\ref{subapp:training} and ~\ref{subapp:threat}. Beyond the main experiments in this section, we further evaluate the generalization to three other categories of threat models in Appendix~\ref{app:other}, respectively with different perturbation budget $\epsilon$, $\ell_{2}$ norm, and targeted attacks. Also, two types of secondary related works, respectively introducing synthetic data and inference-time defense for a better trade-off, are additionally compared in Appendix~\ref{app:comp}. Then in Appendix~\ref{app:ablation}, we conduct comprehensive ablation studies on four hyper-parameters, $t$, $\alpha$, $\beta_1$, and $\beta_2$, followed by further discussions in Appendix~\ref{app:discussion} regarding the difference with DuRM~\citep{wang2023frustratingly} and conventional soft-label strategies, as well as the outliers observed in TRADES + DUCAT.

\paragraph{Effectiveness in Improving Benchmarks.}\label{para:effec}
As a plug-and-play method, DUCAT can be easily integrated into the four experimental benchmarks. Here we demonstrate the improvement brought by such integration to confirm the wide effectiveness of our new paradigm and method. Specifically, we report clean accuracy on benign test samples, adversarial robustness under three adversaries, namely PGD-10, PGD-100~\citep{madry2018towards} and AA~\citep{croce2020reliable} with default settings and random start, and two measures for the accuracy-robustness trade-off, namely Mean and NRR, as detailed in Appendix~\ref{subapp:exp_measure}. All the reported results are averages of three runs, with the specific performance of each run acquired on the best checkpoint achieving the highest PGD-10 accuracy. The results on ResNet-18 in Table~\ref{tab:resnet18} show that DUCAT significantly improves all benchmarks in both accuracy and robustness across all datasets, demonstrating its general effectiveness and confirming the success of our new paradigm in releasing the current trade-off between clean and adversarial targets. Additional results on WideResNet-28-10 are provided by Table~\ref{tab:wrn2810} in Appendix~\ref{app:setup}, and a discussion about the relatively less improvement of TRADES + DUCAT is in Appendix~\ref{subapp:trades+ducat}.

\begin{wrapfigure}[22]{r}{6.1cm}
\centering
\vspace{-1.1em}
\includegraphics[width=\linewidth]{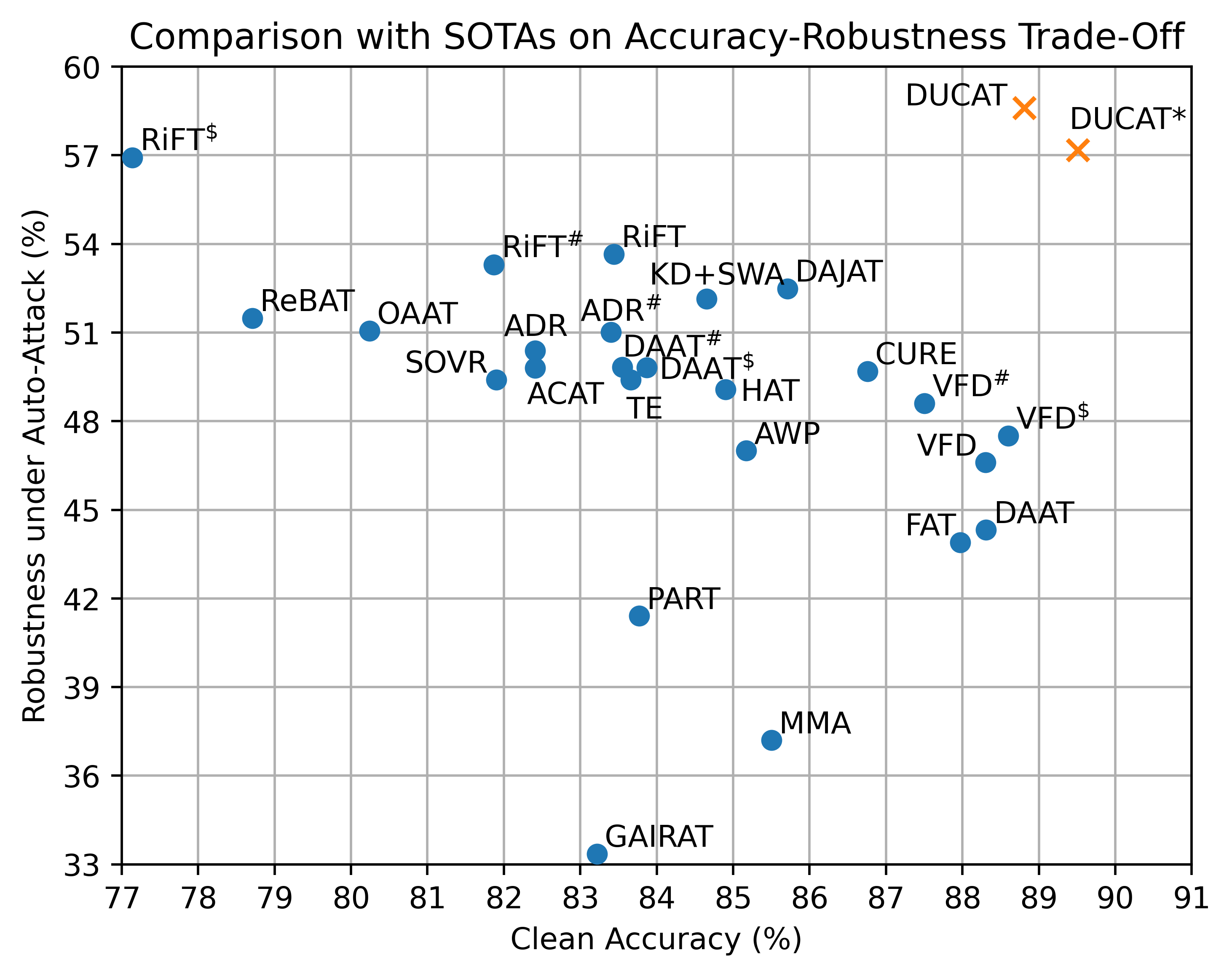}
\vspace{-1.3em}
\caption{Comparison between the proposed DUCAT and 18 current SOTAs in the trade-off problem on CIFAR-10 and ResNet-18. The superscripts $\#$, $\$$, and $*$ in the figure refer to the integration with TRADES, MART and Cons-AT, respectively. The result demonstrates the significant advancement contributed by our work.}\label{fig:sotas}
\end{wrapfigure}

\paragraph{Advancement beyond 18 SOTAs.}\label{para:sotas}
To better demonstrate the advancement contributed by our work, we compare it with the SOTA related works directly aiming at the same accuracy-robustness trade-off problem. Specifically, for CIFAR-10 and ResNet-18 without extra training data, we investigate the advances achieved in the trade-off problem over the past five years, including \citet{ding2020mma}, \citet{wu2020adversarial}, \citet{zhang2020attacks}, \citet{zhang2021geometry}, \citet{chen2021robust}, \citet{addepalli2022efficient}, \citet{addepalli2022scaling}, \citet{dong2022exploring}, \citet{rade2022reducing}, \citet{yang2022one}, \citet{kanai2023one}, \citet{zhu2023improving}, \citet{cao2024vanilla}, ~\citet{gowda2024conserve}, \citet{wu2024annealing}, \citet{zhang2024improving} and \citet{wang2024balance}, and make a comparison with the results reported in their original papers. Considering the fairness, for works without original results under such settings, we turn to reliable external sources including the \textit{RobustBench} leaderboard~\citep{croce2021robustbench} and other published papers such as \citet{liu2021probabilistic}. The comparison result is shown in Figure~\ref{fig:sotas}, where we highlight DUCAT with PGD-AT and Cons-AT, because they are the ones that respectively achieve the best adversarial robustness and clean accuracy among our solutions under the comparison settings. More details are deferred to Table~\ref{tab:sotas} in Appendix~\ref{app:details_at2}.

\begin{wrapfigure}[13]{r}{5cm}
\centering
\vspace{-0.1em}
\includegraphics[width=0.9\linewidth]{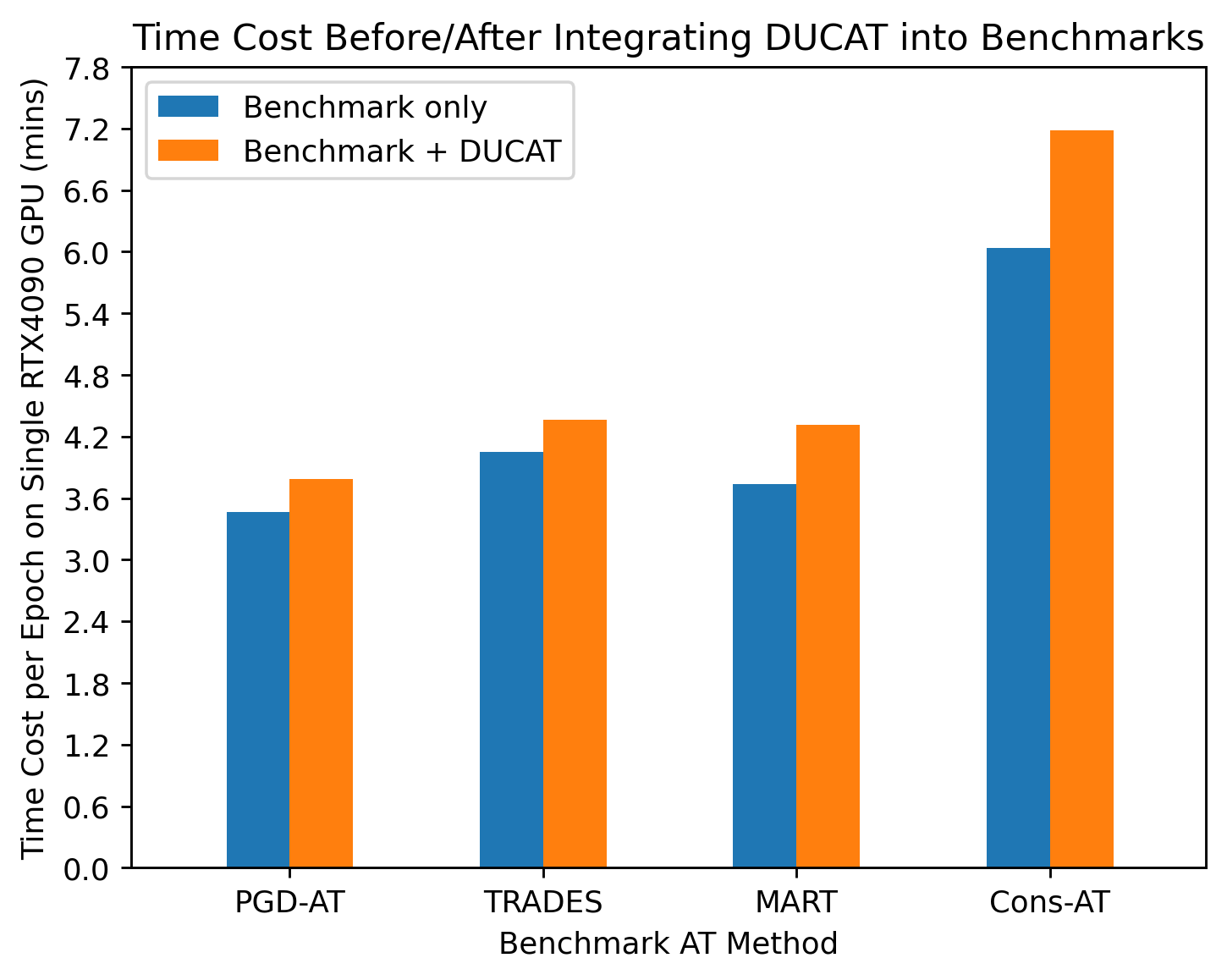}
\vspace{-0.2em}
\caption{DUCAT only introduces slightly more time costs (\ie, 0.59 minutes per epoch, +12.9\% on average across the four benchmarks).}\label{fig:time_cost}
\end{wrapfigure}

\paragraph{Low Additional Cost in Efficiency.}
As DUCAT doubles the last fully connected layer, the model complexity can accordingly increase, which indeed brings an additional price in the training efficiency. However, compared with the parameters of the whole model, the new parameters introduced are expected to be minor, and as a result, would not significantly degrade the efficiency. Figure~\ref{fig:time_cost} shows the time cost before and after integrating DUCAT into the four experimental benchmarks, where the additional time costs on RTX4090 are only 0.32, 0.32, 0.58 and 1.15 minutes per epoch, respectively. Considering the corresponding improvements in Table~\ref{tab:resnet18}, we believe such costs are acceptable. This result confirms the additional implementation of DUCAT can be lightweight, which also fits our vision of a more practical AT for real-world applications.

\section{Conclusion}\label{sec:conclu}

In this work, we reveal that always-failed samples widely exist in conventional adversarial training (AT), based on which we explicitly attribute the inherent trade-off between clean accuracy and robustness to the overstrict assumption of the current AT paradigm. In response, we suggest a new AT paradigm with dummy classes to relax this assumption, and propose a plug-and-play DUCAT method accordingly, which mitigates the trade-off and outperforms 18 SOTAs by integrating into four benchmarks. For future works, developing more advanced methods under the new AT paradigm might be a promising direction (\eg, with synthetic data like in Appendix~\ref{subapp:synthetic} or additional regularization technologies such as weight averaging). We hope this work could attract more attention and inspire further studies on different paths from the current one, that achieve both accurate and robust DNNs.

\bibliographystyle{ref}
\bibliography{ref}

\clearpage
\appendix
\section*{\Large Appendix \hspace{1em}}
\stopcontents[main]
\startcontents[app] 
\begin{mdframed}[
    topline=true,bottomline=true,
    leftline=false,rightline=false,
    linewidth=1pt,
    innertopmargin=.6\baselineskip,
    innerbottommargin=1.6\baselineskip,
    skipabove=\baselineskip,skipbelow=\baselineskip]
  \printcontents[app]{l}{1}{}
\end{mdframed}

\section*{Question Index}

As the appendix contains comprehensive details that may not be of interest to every reader, to make better use of it, we provide the following question index to help readers quickly locate the points regarding their main concerns.

\begin{itemize}[leftmargin=2em] 
    \item 
    \textit{May the adversary bypass dummy classes by their correspondence to benign ones?} . . . \ref{subapp:threat}\&\ref{subapp:target} 
    \item 
    \textit{Why AT with external/synthetic data is not involved in the main experiments?} . . . . . . . . . . \ref{subapp:synthetic}
    \item 
    \textit{Is the design of our dummy class similar to DuRM, thus limiting our novelty?} . . . . . . . . . \ref{subapp:albeit}
    \item 
    \textit{Is the design of our two-hot soft label similar to existing ones, thus limiting our novelty?} . . . \ref{subapp:two-hot}
    \item 
    \textit{Why is the improvement of integrating DUCAT to TRADES relatively less than others?} . . . \ref{subapp:trades+ducat}
\end{itemize}

\section{Involved Attacks and AT Methods}\label{app:details}

In this section, we supplement the introduction of the adversarial attack technologies serving as adversaries in this work, including PGD and AA. Also, we introduce more details about the AT methods involved, including four benchmarks namely PGD-AT~\citep{madry2018towards}, TRADES~\citep{zhang2019theoretically}, MART~\citep{wang2020improving} and Cons-AT~\citep{tack2022consistency}, as well as 18 SOTAs on the trade-off problem, respectively AWP~\citep{wu2020adversarial}, FAT~\citep{zhang2020attacks}, MMA~\citep{ding2020mma}, GAIRAT~\citep{zhang2021geometry}, KD + SWA~\citep{chen2021robust}, ACAT~\citep{addepalli2022efficient}, DAAT~\citep{yang2022one}, DAJAT~\citep{addepalli2022efficient}, HAT~\citep{rade2022reducing}, OAAT~\citep{addepalli2022scaling}, TE~\citep{dong2022exploring}, RiFT~\citep{zhu2023improving}, SOVR~\citep{kanai2023one}, ADR~\citep{wu2024annealing}, CURE~\citep{gowda2024conserve}, PART~\citep{zhang2024improving}, VFD~\citep{cao2024vanilla} and ReBAT~\citep{wang2024balance}.

\subsection{Adversarial Robustness and Attack Methods}\label{app:details_attack}

With the remarkable development of DNNs in various real-world applications~\citep{zhang2023comprehensive,qin2024tool,zhou2025multi}, certain security concerns have received increasing attention in recent years~\citep{liang2024alignment,lin2025backdoordm,liang2025does}.
Among them, adversarial robustness refers to the performance of DNN classifiers under malicious perturbations by any possible adversaries, which is commonly measured with the test accuracy under adversarial attacks~\citep{bai2021recent}. The concept of adversarial attack is proposed by \citet{biggio2013evasion} and \citet{szegedy2014intriguing}, followed by a basic write-box method, \textit{fast gradient sign method} (FGSM)~\citep{goodfellow2015explaining}, that generates the perturbation based on the gradient of the loss function \textit{w.r.t.} the input sample. The \textit{projected gradient descent attack} (PGD)~\citep{madry2018towards} can be viewed as a variant of FGSM, which first produces the perturbation by iteratively running FGSM, and then projects it back into the $\epsilon$-ball of the input sample. Nowadays, as Auto-Attack (AA)~\citep{croce2020reliable} forms a parameter-free and user-independent ensemble of attacks, which can identify frequent pitfalls in AT practice including over-adjustment of hyper-parameters and gradient obfuscation or masking, it has been widely recognized as one of the most reliable adversaries to exam robustness.  

\subsection{Experimental AT Benchmarks}\label{app:details_at1}

\textbf{PGD-AT}~\citep{madry2018towards}\textbf{.} This is the first method demonstrated to be effective for solving the \textit{min-max} problem of the current AT paradigm in Equation~(\ref{eq:at}) and training moderately robust DNNs~\citep{athalye2018obfuscated,wang2020improving}. To be specific, based on \textit{Danskin’s Theorem}~\citep{danskin2012theory}, PGD-AT proposed to find a constrained maximizer of the inner maximization by PGD, which is believed sufficiently close to the optimal attack, and then use the maximizer as an actual data point for the outer minimization through gradient descent.

\textbf{TRADES}~\citep{zhang2019theoretically}\textbf{.} As another typical AT method, TRADES proposed to decompose the robust error into natural error and boundary error, thus directly balancing the trade-off between natural accuracy and robustness. Specifically, boundary error occurs when the specific data point is sufficiently close to the decision boundary that can easily cross it under slight perturbation, which is believed as one reason for the existence of adversarial samples~\citep{bai2021recent,wang2024tsfool}. 

\textbf{MART}~\citep{wang2020improving}\textbf{.} A problem of TRADES is that the boundary error is designed to push each pair of benign and adversarial samples together, no matter whether the benign data are classified correctly or not~\citep{bai2021recent}. As a follow-up work, MART further investigates the influence of correctly classified and misclassified samples for adversarial robustness separately, and then suggests adopting additional boundary error\textit{ w.r.t.} misclassified samples. 

\textbf{Cons-AT}~\citep{tack2022consistency}\textbf{.} It is found that \textit{consistency regularization} forces the model to give the same output distribution when the input or weights are slightly perturbed, which fits the goal of AT when the perturbation is generated adversarially~\citep{zhang2022alleviating}. Cons-AT proposed a new target that the predictive distributions after attacking from two different augmentations of the same instance should be similar to each other. The underlying principle is that adversarial robustness essentially refers to model stability around naturally occurring inputs, learning to satisfy such a constraint should not inherently require labels~\citep{carmon2019unlabeled}, which also relaxes the current assumption from a different perspective.

\subsection{SOTA AT Methods on Trade-Off between Accuracy and Robustness}\label{app:details_at2}

The trade-off between clean accuracy and adversarial robustness has been explored from various perspectives under the current AT paradigm. Below we list 18 representative SOTAs on this problem in the past five years, which are adopted to demonstrate the advancement achieved by our new AT paradigm and DUCAT method in Section~\ref{para:sotas}. In this section, we show the sources and detailed records for these SOTAs in Table~\ref{tab:sotas}, which are the corresponding raw data we rely on to draw the aforementioned Figure~\ref{fig:sotas}. Besides, we also briefly introduce the principle for each of these SOTAs, except for FAT~\citep{zhang2020attacks}, HAT~\citep{rade2022reducing} and SOVR~\citep{kanai2023one} that have been introduced in Section~\ref{sec:intro}.

\begin{table*}[htb]
  \caption{Corresponding to Figure~\ref{fig:sotas} in Section~\ref{para:sotas}, this table illustrates the raw data of the 18 SOTAs on the trade-off problem, demonstrating the advancement of our work. Specifically, the best results for each measure are marked in bold, while the sub-optimal results are marked with underlines. Note that ReBAT is the only one here with PreActResNet-18 due to the lack of results on ResNet-18 in its original paper.}
  \label{tab:sotas}
  \centering
  \renewcommand\arraystretch{1.2}
  \resizebox{13.9cm}{!}{
  \begin{tabular}{cl|cccc|c}
  \toprule
  \textbf{Year} & \textbf{Method} & \textbf{Clean} & \textbf{AA} & \textbf{Mean} & \textbf{NRR} & \textbf{Record Source} \\
  \midrule
   & AWP~\citep{wu2020adversarial} & 85.17 & 47.00 & 66.085 & 60.573 & \citet{liu2021probabilistic} \\
   2020 & FAT~\citep{zhang2020attacks} & 87.97 & 43.90 & 65.935 & 58.571 & \citet{liu2021probabilistic} \\
   & MMA~\citep{ding2020mma} & 85.50 & 37.20 & 61.350 & 51.844 & Original paper \\
  \midrule
   \multirow{2}{*}{2021} & GAIRAT~\citep{zhang2021geometry} & 83.22 & 33.35 & 58.285 & 47.618 & \citet{liu2021probabilistic} \\
   & KD + SWA~\citep{chen2021robust} & 84.65 & 52.14 & 68.395 & 64.532 & Original paper \\
  \midrule
   & ACAT~\citep{addepalli2022efficient} & 82.41 & 49.80 & 66.105 & 62.083 & Original paper \\
   & DAAT~\citep{yang2022one} & 88.31 & 44.32 & 66.315 & 59.020 & Original paper \\
  \multirow{2}{*}{2022} & DAJAT~\citep{addepalli2022efficient} & 85.71 & 52.48 & 69.095 & 65.100 & \textit{RobustBench} \\
   & HAT~\citep{rade2022reducing} & 84.90 & 49.08 & 66.990 & 62.202 & Original paper \\
   & OAAT~\citep{addepalli2022scaling} & 80.24 & 51.06 & 65.650 & 62.408 & \textit{RobustBench} \\
   & TE~\citep{dong2022exploring} & 83.66 & 49.40 & 66.530 & 62.119 & Original paper \\
  \midrule
  \multirow{2}{*}{2023} & RiFT~\citep{zhu2023improving} & 83.44 & 53.65 & 68.545 & 65.308 & Original paper \\
   & SOVR~\citep{kanai2023one} & 81.90 & 49.40 & 65.650 & 61.628 & Original paper \\
  \midrule
   & ADR~\citep{wu2024annealing} & 82.41 & 50.38 & 66.395 & 62.532 & Original paper \\
   & CURE~\citep{gowda2024conserve} & 86.76 & 49.69 & 68.225 & 63.190 & Original paper \\
   2024 & PART~\citep{zhang2024improving} & 83.77 & 41.41 & 62.590 & 55.423 & Original paper \\
   & VFD~\citep{cao2024vanilla} & 88.30 & 46.60 & 67.450 & 61.005 & Original paper \\  
   & ReBAT~\citep{wang2024balance} & 78.71 & 51.49 & 65.100 & 62.255 & Original paper \\
  \midrule
  \midrule
  \multirow{2}{*}{\textbf{Ours}} & \textbf{PGD-AT + DUCAT} & \underline{88.81} & \textbf{58.61} & \textbf{73.710} & \textbf{70.617} & - \\
  & \textbf{Cons-AT + DUCAT} & \textbf{89.51} & \underline{57.18} & \underline{73.345} & \underline{69.782} & - \\
  \bottomrule
  \end{tabular}
  }
\end{table*}

AWP~\citep{wu2020adversarial} proposed a double perturbation mechanism that can flatten the loss landscape by weight perturbation to improve robust generalization. MMA~\citep{ding2020mma} proposed to use adaptive $\epsilon$ for adversarial perturbations to directly estimate and maximize the margin between data and the decision boundary. GAIRAT~\citep{zhang2021geometry} proposed that a natural data point closer to (or farther from) the class boundary is less (or more) robust, and the corresponding adversarial data point should be assigned with larger (or smaller) weight. DAJAT~\citep{addepalli2022efficient} aims to handle the conflicting goals of enhancing the diversity of the training dataset and training with data that is close to the test distribution by using a combination of simple and complex augmentations with separate batch normalization layers. ACAT~\citep{addepalli2022efficient} is a two-step variant of DAJAT to improve computational efficiency. OAAT~\citep{addepalli2022scaling} aligns the predictions of the model with that of an Oracle during AT to achieve robustness within larger bounds. 

Enlightening, TE~\citep{dong2022exploring} proposed that one-hot labels can be noisy for them because they naturally lie close to the decision boundary, which makes it essentially difficult to assign high-confident one-hot labels for all perturbed samples within the $\epsilon$-ball of them~\citep{stutz2020confidence,cheng2022cat}. So the model may try to memorize these hard samples during AT, resulting in robust overfitting. This may also be another reason why AT leads to more complicated decision boundary, and explains why robust overfitting is harder to alleviate than the clean one. Based on a similar idea, KD + SWA~\citep{chen2021robust} investigates two empirical means to inject more learned smoothening during AT, namely leveraging knowledge distillation and self-training to smooth the \textit{logits}, as well as performing stochastic weight averaging~\citep{izmailov2018averaging} to smooth the weights, and DAAT~\citep{yang2022one} adaptively adjusts the perturbation ball to a proper size for each of the natural examples with the help of a naturally trained calibration network.

More recently, RiFT~\citep{zhu2023improving} introduces module robust criticality, a measure that evaluates the significance of a given module to model robustness under worst-case weight perturbations, to exploit the redundant capacity for robustness by fine-tuning the adversarially trained model on its non-robust-critical module. ADR~\citep{wu2024annealing} generates soft labels as a better guidance mechanism that accurately reflects the distribution shift under attack during AT. CURE~\citep{gowda2024conserve} finds that selectively updating specific layers while preserving others can substantially enhance the network's learning capacity, and accordingly leverages a gradient prominence criterion to perform selective conservation, updating, and revision of weights. PART~\citep{zhang2024improving} partially reduces $\epsilon$ for less influential pixels, guiding the model to focus more on key regions that affect its outputs. VFD~\citep{cao2024vanilla} conducts knowledge distillation from a pre-trained model optimized towards high accuracy to guide the AT model towards generating high-quality and well-separable features by constraining the obtained features of natural and adversarial examples. ReBAT~\citep{wang2024balance} views AT as a dynamic mini-max game between the model trainer and the attacker, and proposes to alleviate robust overfitting by rebalancing the two players by either regularizing the trainer's capacity or improving the attack strength.

\section{DUCAT Algorithm}\label{app:algo}

In this section, we provide the specific algorithm to facilitate the understanding of the proposed DUCAT. This algorithm demonstrates the basic format of DUCAT (\ie, integrated with PGD-AT), and focuses on its novel parts, thus leaving out some general training details such as optimizer settings,

\begin{algorithm}[htb]
    \caption{\textbf{DU}mmy \textbf{C}lasses-based \textbf{AT} (\textbf{DUCAT})}
    \label{alg:DUCAT}
    \textbf{Input}: Training dataset $\mathcal{D} = \{(\mathbf{x}_i, y_i)\}_{i = 1, ..., n}$ of a $C$-class classification task, target model $\bm{\theta}$ \\
    \textbf{General Parameters}: Training epoch $T$, resuming epoch $T_r$, training adversary $\mathcal{A}$, loss function $\mathcal{L}$, learning rate $r$ \\
    \textbf{DUCAT Hyper-parameters}: Start epoch $t$, preference weight $\alpha$, benign weight $\beta_1$ and adversarial weight $\beta_2$ for two-hot soft label construction \\
    \textbf{Output}: Robust model $\bm{\theta}'$
    \begin{algorithmic}[1] 
        \STATE Initialize $t_{\textit{curr}} \leftarrow T_r$ \textbf{if provided} $T_r$ \textbf{else} $0$
        \STATE $\bm{\theta}' \leftarrow \textit{doubleLastLayer}\,(\bm{\theta})$
        \WHILE{$t_{\textit{curr}} < T$}
        \STATE Initialize $\mathcal{D}_{\textit{curr}}, \,\mathcal{D}_{\textit{benign}}, \, \mathcal{D}_{\textit{adv}} \leftarrow [], \, \mathcal{A}_{\textit{curr}} \leftarrow \mathcal{A}(\bm{\theta}')$
        \WHILE{$\mathbf{x}_i, y_i \in \mathcal{D}$}
        \STATE $\mathbf{x}_i' \leftarrow \mathcal{A}_{\textit{curr}}(\mathbf{x}_i)$
        \IF{$t_{\textit{curr}} \geq t$}
            \STATE $\mathbf{y}_i \leftarrow \textit{onehot}\,(y_i)$
            \STATE $\mathcal{D}_{\textit{benign}}.\textit{append}\,(\mathbf{x}_i,\, (\beta_1 \cdot \mathbf{y}_i) \, \| \, ((1 - \beta_1) \cdot \mathbf{y}_i))$
            \STATE $\mathcal{D}_{\textit{adv}}.\textit{append}\,(\mathbf{x}_i',\, ((1 - \beta_2) \cdot \mathbf{y}_i) \, \| \, (\beta_2 \cdot \mathbf{y}_i))$
        \ELSE
            \STATE $\mathcal{D}_{\textit{curr}}.\textit{append}\,(\mathbf{x}_i',\, y_i)$
        \ENDIF
        \ENDWHILE
        \IF{$t_{\textit{curr}} \geq t$}
            \STATE $\bm{\theta}' \!\leftarrow\! \bm{\theta}' - r \cdot [ \, \alpha \cdot \nabla_{\bm{\theta}'}\mathcal{L}(\mathcal{D}_{\textit{benign}}) + (1\!-\!\alpha) \cdot \nabla_{\bm{\theta}'}\mathcal{L}(\mathcal{D}_{\textit{adv}})]$
        \ELSE
            \STATE $\bm{\theta}' \leftarrow \bm{\theta}' - r \cdot \nabla_{\bm{\theta}'}\mathcal{L}(\mathcal{D}_{\textit{curr}})$ 
        \ENDIF
        \STATE $t_{\textit{curr}} \leftarrow t_{\textit{curr}} + 1$
        \ENDWHILE
        \STATE \textbf{return} $\bm{\theta}'$
    \end{algorithmic}
\end{algorithm}

training adversary, mini-batch and learning rate decay, which have been detailed in Appendix~\ref{subapp:training}. Notice that as suggested in Appendix~\ref{app:ablation}, DUCAT can be used for both training a randomly initialized model from scratch and fine-tuning an already adversarially trained model for a better trade-off between accuracy and robustness, so we introduce the resuming epoch $T_r$ to enable customization in the latter case.

\begin{figure*}[htb]
  \centering
  \begin{subfigure}{0.42\linewidth}
    \centering
    \includegraphics[width=1\linewidth]{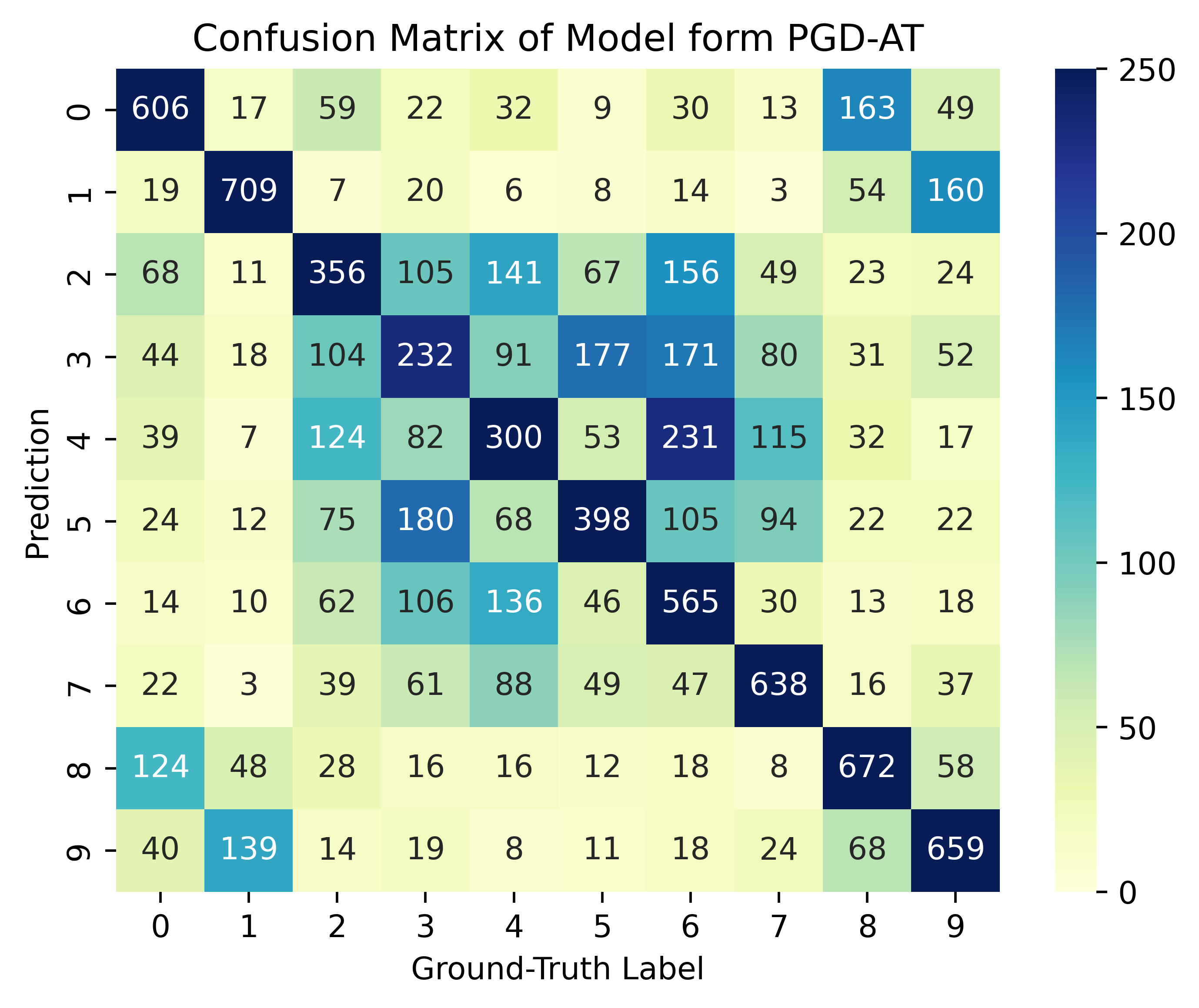}
  \end{subfigure}
  \begin{subfigure}{0.42\linewidth}
    \centering
    \includegraphics[width=1\linewidth]{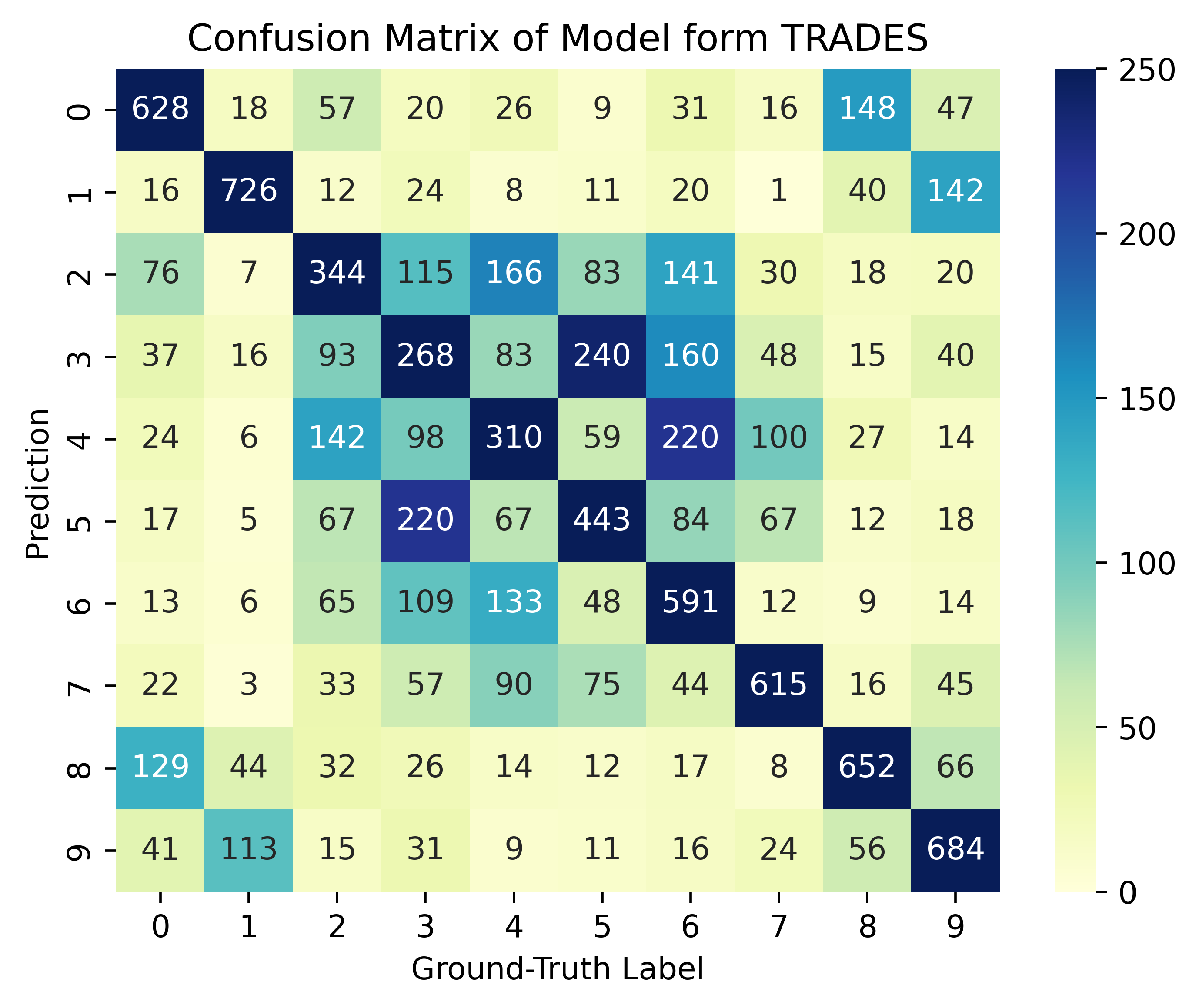}
  \end{subfigure}
  \begin{subfigure}{0.42\linewidth}
    \centering
    \includegraphics[width=1\linewidth]{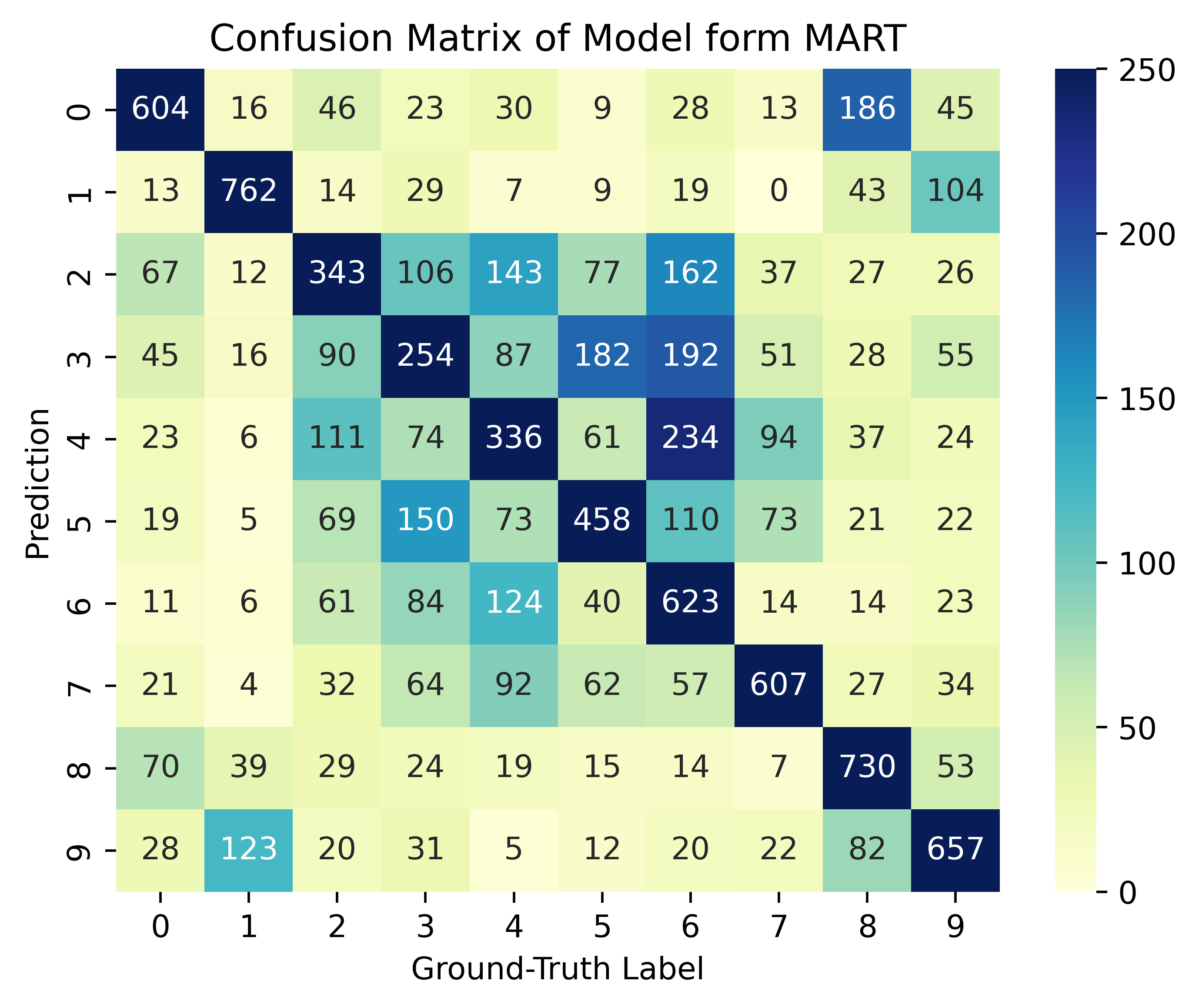}
  \end{subfigure}
  \begin{subfigure}{0.42\linewidth}
    \centering
    \includegraphics[width=1\linewidth]{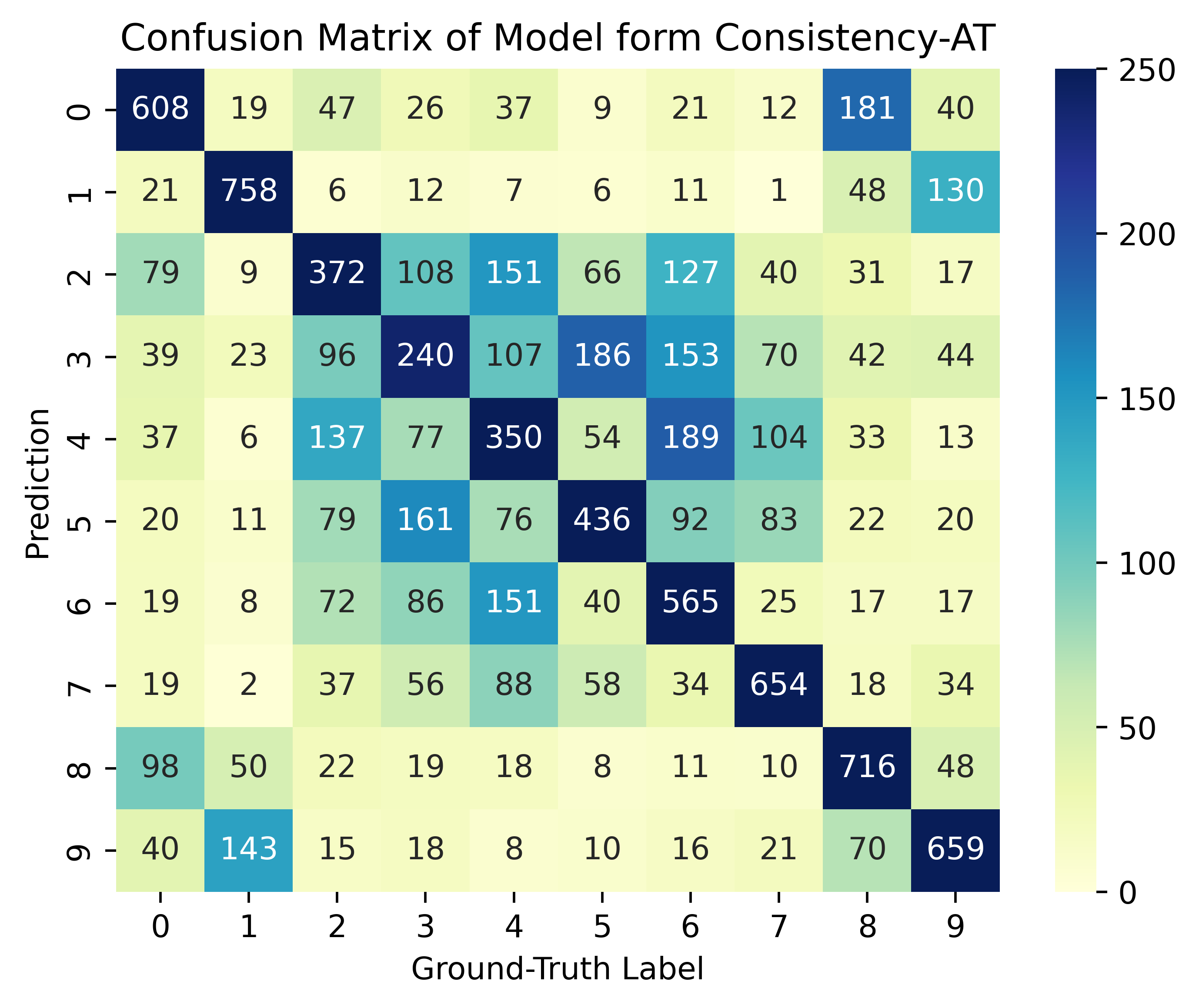}
  \end{subfigure}
  \caption{Corresponding to Figure~\ref{fig:cross_methods_1} in Section~\ref{para:always}, under the PGD-10 adversary, similar failure patterns can be observed in the confusion matrixes respectively with the protection across the four experimental AT methods.}
  \label{fig:cross_methods_2}
\end{figure*}

\section{Experimental Setup}\label{app:setup}

This section supplements the specific experimental setup of our work. To be specific, we first provide the training details in Appendix~\ref{subapp:training} to ensure fair comparison and facilitate easy reproduction, followed by a detailed description of the threat model considered in the main experiments in Appendix~\ref{subapp:threat}. Finally, all of the six adopted evaluation measures are introduced in Appendix~\ref{subapp:exp_measure}.

\subsection{Training Details}\label{subapp:training} 

We adopt SGD optimizer with momentum 0.9, batch size 128, weight decay $5 \!\times\! 10^{-4}$, initial learning rate 0.1, total 130 training epochs with learning rate decay by a factor of 0.1 at 100 and 105 epochs, respectively. Except for extending the training epoch which is originally 110 for better convergence, we follow the common settings as suggested by \citet{pang2021bag}, which studies various tricks and hyper-parameters of AT. Standard data pre-processing is also involved in the training process, with normalization of all natural images into [0, 1] and data augmentation including random crop with 4-pixel zero padding and 50\% random horizontal flip. Then, regarding the hyper-parameters specific to different AT methods, for the proposed DUCAT, we uniformly assign $\alpha = 0.5$, $\beta_1 = 0.75$ and $\beta_2 = 1$, respectively starting at epoch 105 for CIFAR-10 and 100 for the other two datasets. And for the benchmarks, we adopt the same settings as their original papers. 

The experiments are conducted on Ubuntu 22.04 with 512GB RAM and 8 $\!\times\!$ NVIDIA GeForce RTX 4090 GPUs, and are implemented with Python 3.8.19 and PyTorch 1.8.1 $\!$+$\!$ cu111.

\begin{table*}[htb]
  \caption{The additional results of integrating the proposed DUCAT to four AT benchmarks on CIFAR-10, CIFAR-100 and Tiny-ImageNet with WideResNet-28-10 under $\ell_{\infty}$ adversaries, corresponding to Table~\ref{tab:resnet18} in Section~\ref{para:effec}. All results are acquired from three runs.}
  \label{tab:wrn2810}
  \centering
  \setlength{\tabcolsep}{3pt}
  \renewcommand\arraystretch{1.4}
  \resizebox{13.9cm}{!}{
  \begin{tabular}{clllllll}
    \toprule
    \multicolumn{1}{c}{\textbf{Dataset}} & \multicolumn{1}{c}{\textbf{Method}} & \multicolumn{1}{c}{\hspace{-1em} \textbf{Clean}} & \multicolumn{1}{c}{\hspace{-1em} \textbf{PGD-10}} & \multicolumn{1}{c}{\hspace{-1em} \textbf{PGD-100}} & \multicolumn{1}{c}{\hspace{-1em} \textbf{AA}} & \multicolumn{1}{c}{\hspace{-1em} \textbf{Mean}} & \multicolumn{1}{c}{\hspace{-1em} \textbf{NRR}} \\
    \midrule\midrule

    & PGD-AT
      & 87.49 & 55.81 & 54.37 & 50.96 & 69.225 & 64.406 \\
    & \hspace{1em}\textbf{+ DUCAT}
      & 90.02 $^{\color{ForestGreen}{(+2.9\%)}}$
      & 64.53 $^{\color{ForestGreen}{(+15.6\%)}}$
      & 62.76 $^{\color{ForestGreen}{(+15.4\%)}}$
      & 58.77 $^{\color{ForestGreen}{(+15.3\%)}}$
      & 74.395 $^{\color{ForestGreen}{(+7.5\%)}}$
      & 71.113 $^{\color{ForestGreen}{(+10.4\%)}}$ \\
    \cmidrule(r){2-8}
    & TRADES
      & 85.35 & 57.15 & 56.17 & 51.88 & 68.615 & 64.533 \\
    & \hspace{1em}\textbf{+ DUCAT}
      & 83.62 $^{\color{Salmon}{(-2.0\%)}}$
      & 57.77 $^{\color{ForestGreen}{(+1.1\%)}}$
      & 57.00 $^{\color{ForestGreen}{(+1.5\%)}}$
      & 53.78 $^{\color{ForestGreen}{(+3.7\%)}}$
      & 68.700 $^{\color{ForestGreen}{(+0.1\%)}}$
      & 65.460 $^{\color{ForestGreen}{(+1.4\%)}}$ \\
    \cmidrule(r){2-8}
    \multirow{2}{*}{CIFAR-10}
    & MART
      & 82.78 & 58.47 & 57.47 & 50.89 & 66.835 & 63.031 \\
    & \hspace{1em}\textbf{+ DUCAT}
      & 84.64 $^{\color{ForestGreen}{(+2.2\%)}}$
      & 58.26 $^{\color{Salmon}{(-0.4\%)}}$
      & 57.07 $^{\color{Salmon}{(-0.7\%)}}$
      & 54.64 $^{\color{ForestGreen}{(+7.4\%)}}$
      & 69.640 $^{\color{ForestGreen}{(+4.2\%)}}$
      & 66.482 $^{\color{ForestGreen}{(+5.5\%)}}$ \\
    \cmidrule(r){2-8}
    & Cons-AT
      & 86.90 & 55.71 & 54.41 & 50.83 & 68.865 & 64.146 \\
    & \hspace{1em}\textbf{+ DUCAT}
      & 89.86 $^{\color{ForestGreen}{(+3.4\%)}}$
      & 67.89 $^{\color{ForestGreen}{(+21.9\%)}}$
      & 64.17 $^{\color{ForestGreen}{(+17.9\%)}}$
      & 59.44 $^{\color{ForestGreen}{(+16.9\%)}}$
      & 74.650 $^{\color{ForestGreen}{(+8.4\%)}}$
      & 71.551 $^{\color{ForestGreen}{(+11.6\%)}}$ \\
    \cmidrule(r){2-8}
    & Average
      & 85.63 & 56.79 & 55.61 & 51.14 & 68.385 & 64.029 \\
    & \hspace{1em}\textbf{+ DUCAT}
      & 87.04 $^{\color{ForestGreen}{(+1.6\%)}}$
      & 62.11 $^{\color{ForestGreen}{(+9.4\%)}}$
      & 60.25 $^{\color{ForestGreen}{(+8.4\%)}}$
      & 56.66 $^{\color{ForestGreen}{(+10.8\%)}}$
      & 71.846 $^{\color{ForestGreen}{(+5.1\%)}}$
      & 68.652 $^{\color{ForestGreen}{(+7.2\%)}}$ \\
    \midrule\midrule

    & PGD-AT
      & 59.95 & 32.18 & 31.21 & 27.27 & 43.610 & 37.489 \\
    & \hspace{1em}\textbf{+ DUCAT}
      & 70.02 $^{\color{ForestGreen}{(+16.8\%)}}$
      & 34.28 $^{\color{ForestGreen}{(+6.5\%)}}$
      & 30.73 $^{\color{Salmon}{(-1.5\%)}}$
      & 27.95 $^{\color{ForestGreen}{(+2.5\%)}}$
      & 48.985 $^{\color{ForestGreen}{(+12.3\%)}}$
      & 39.952 $^{\color{ForestGreen}{(+6.6\%)}}$ \\
    \cmidrule(r){2-8}
    & TRADES
      & 59.51 & 32.51 & 32.34 & 27.71 & 43.610 & 37.814 \\
    & \hspace{1em}\textbf{+ DUCAT}
      & 59.70 $^{\color{ForestGreen}{(+0.3\%)}}$
      & 33.77 $^{\color{ForestGreen}{(+3.9\%)}}$
      & 33.38 $^{\color{ForestGreen}{(+3.2\%)}}$
      & 28.55 $^{\color{ForestGreen}{(+3.0\%)}}$
      & 44.125 $^{\color{ForestGreen}{(+1.2\%)}}$
      & 38.627 $^{\color{ForestGreen}{(+2.2\%)}}$ \\
    \cmidrule(r){2-8}
    \multirow{2}{*}{CIFAR-100}
    & MART
      & 56.84 & 34.12 & 33.70 & 27.97 & 42.405 & 37.491 \\
    & \hspace{1em}\textbf{+ DUCAT}
      & 60.97 $^{\color{ForestGreen}{(+7.3\%)}}$
      & 36.80 $^{\color{ForestGreen}{(+7.9\%)}}$
      & 35.98 $^{\color{ForestGreen}{(+6.8\%)}}$
      & 32.40 $^{\color{ForestGreen}{(+15.8\%)}}$
      & 46.685 $^{\color{ForestGreen}{(+10.1\%)}}$
      & 42.314 $^{\color{ForestGreen}{(+12.9\%)}}$ \\
    \cmidrule(r){2-8}
    & Cons-AT
      & 60.84 & 32.48 & 31.64 & 27.74 & 44.290 & 38.106 \\
    & \hspace{1em}\textbf{+ DUCAT}
      & 72.97 $^{\color{ForestGreen}{(+19.9\%)}}$
      & 35.90 $^{\color{ForestGreen}{(+10.5\%)}}$
      & 32.24 $^{\color{ForestGreen}{(+1.9\%)}}$
      & 28.31 $^{\color{ForestGreen}{(+2.1\%)}}$
      & 50.640 $^{\color{ForestGreen}{(+14.4\%)}}$
      & 40.793 $^{\color{ForestGreen}{(+7.1\%)}}$ \\
    \cmidrule(r){2-8}
    & Average
      & 59.29 & 32.82 & 32.22 & 27.67 & 43.479 & 37.725 \\
    & \hspace{1em}\textbf{+ DUCAT}
      & 65.92 $^{\color{ForestGreen}{(+11.2\%)}}$
      & 35.19 $^{\color{ForestGreen}{(+7.2\%)}}$
      & 33.08 $^{\color{ForestGreen}{(+2.7\%)}}$
      & 29.30 $^{\color{ForestGreen}{(+5.9\%)}}$
      & 47.609 $^{\color{ForestGreen}{(+9.5\%)}}$
      & 40.422 $^{\color{ForestGreen}{(+7.2\%)}}$ \\
    \midrule\midrule

    & PGD-AT
      & 47.79 & 23.97 & 23.59 & 20.00 & 33.895 & 28.199 \\
    & \hspace{1em}\textbf{+ DUCAT}
      & 62.58 $^{\color{ForestGreen}{(+30.9\%)}}$
      & 25.86 $^{\color{ForestGreen}{(+7.9\%)}}$
      & 23.68 $^{\color{ForestGreen}{(+0.4\%)}}$
      & 18.65 $^{\color{Salmon}{(-6.8\%)}}$
      & 40.615 $^{\color{ForestGreen}{(+19.8\%)}}$
      & 28.736 $^{\color{ForestGreen}{(+1.9\%)}}$ \\
    \cmidrule(r){2-8}
    & TRADES
      & 51.14 & 24.84 & 24.58 & 20.02 & 35.580 & 28.776 \\
    & \hspace{1em}\textbf{+ DUCAT}
      & 51.91 $^{\color{ForestGreen}{(+1.5\%)}}$
      & 25.77 $^{\color{ForestGreen}{(+3.7\%)}}$
      & 25.39 $^{\color{ForestGreen}{(+3.3\%)}}$
      & 20.22 $^{\color{ForestGreen}{(+1.0\%)}}$
      & 36.065 $^{\color{ForestGreen}{(+1.4\%)}}$
      & 29.104 $^{\color{ForestGreen}{(+1.1\%)}}$ \\
    \cmidrule(r){2-8}
    \multirow{2}{*}{Tiny-ImageNet}
    & MART
      & 45.57 & 26.21 & 25.92 & 21.07 & 33.320 & 28.816 \\
    & \hspace{1em}\textbf{+ DUCAT}
      & 50.71 $^{\color{ForestGreen}{(+11.3\%)}}$
      & 28.48 $^{\color{ForestGreen}{(+8.7\%)}}$
      & 28.64 $^{\color{ForestGreen}{(+10.5\%)}}$
      & 26.15 $^{\color{ForestGreen}{(+24.1\%)}}$
      & 38.430 $^{\color{ForestGreen}{(+15.3\%)}}$
      & 34.506 $^{\color{ForestGreen}{(+19.8\%)}}$ \\
    \cmidrule(r){2-8}
    & Cons-AT
      & 50.12 & 25.05 & 24.42 & 20.64 & 35.380 & 29.239 \\
    & \hspace{1em}\textbf{+ DUCAT}
      & 62.77 $^{\color{ForestGreen}{(+25.2\%)}}$
      & 25.62 $^{\color{ForestGreen}{(+2.3\%)}}$
      & 25.01 $^{\color{ForestGreen}{(+2.4\%)}}$
      & 22.09 $^{\color{ForestGreen}{(+7.0\%)}}$
      & 42.430 $^{\color{ForestGreen}{(+19.9\%)}}$
      & 32.680 $^{\color{ForestGreen}{(+11.8\%)}}$ \\
    \cmidrule(r){2-8}
    & Average
      & 48.66 & 25.02 & 24.63 & 20.43 & 34.544 & 28.758 \\
    & \hspace{1em}\textbf{+ DUCAT}
      & 56.99 $^{\color{ForestGreen}{(+14.9\%)}}$
      & 26.43 $^{\color{ForestGreen}{(+5.7\%)}}$
      & 25.68 $^{\color{ForestGreen}{(+4.3\%)}}$
      & 21.78 $^{\color{ForestGreen}{(+6.6\%)}}$
      & 39.385 $^{\color{ForestGreen}{(+14.0\%)}}$
      & 31.257 $^{\color{ForestGreen}{(+8.7\%)}}$ \\
    \bottomrule
  \end{tabular}
  }
\end{table*}

\subsection{Threat Models}\label{subapp:threat}

In the main experiments, we consider untargeted white-box attacks under $\ell_{\infty}$ norm, with the maximal perturbation budget $\epsilon = 8/255$ and optimization step size $\alpha = 2/255$, which is the most commonly adopted threat model for evaluating adversarial robustness. Additionally, in Appendix~\ref{app:other}, we further consider three other categories of threat models, respectively regarding different perturbation budgets, as well as adversaries adopting $\ell_{2}$ norm or targeted attack.

\begin{table*}[htb]
\caption{The supplementary experiments evaluating four different perturbation budgets under $\ell_{\infty}$ adversaries on CIFAR-10 and ResNet-18. All results are acquired from three runs.}
  \label{tab:budget}
  \centering
  \renewcommand\arraystretch{1.4}
  \resizebox{13.9cm}{!}{
    \begin{tabular}{lccccccccccccc}
    \toprule
    \multirow{2}{*}{\hspace{0.7em} \textbf{Method}} & \multirow{2}{*}{\textbf{Clean}} & \multicolumn{3}{c}{$\epsilon$ \textbf{= 2/255}} & \multicolumn{3}{c}{$\epsilon$ \textbf{= 4/255}} & \multicolumn{3}{c}{$\epsilon$ \textbf{= 16/255}} & \multicolumn{3}{c}{$\epsilon$ \textbf{= 32/255}} \\
    \cmidrule(r){3-5} \cmidrule(r){6-8} \cmidrule(r){9-11} \cmidrule(r){12-14}
     &  & \textbf{AA} & \textbf{Mean} & \textbf{NRR} & \textbf{AA} & \textbf{Mean} & \textbf{NRR} & \textbf{AA} & \textbf{Mean} & \textbf{NRR} & \textbf{AA} & \textbf{Mean} & \textbf{NRR} \\
    \midrule
    \midrule
    PGD-AT & 82.92 & 71.43 & 77.175 & 76.746 & 63.90 & 73.410 & 72.177 & 13.70 & 48.310 & 23.508 & 0.39 & 41.655 & 0.776 \\
    \hspace{1em} \textbf{+ DUCAT} & \textbf{88.81} & \textbf{74.70} & \textbf{81.755} & \textbf{81.149} & \textbf{68.93} & \textbf{78.870} & \textbf{77.616} & \textbf{28.00} & \textbf{58.405} & \textbf{42.578} & \textbf{6.74} & \textbf{47.775} & \textbf{12.527} \\
    \midrule
    TRADES & \textbf{79.67} & \textbf{68.59} & \textbf{74.130} & \textbf{73.718} & \textbf{62.09} & \textbf{70.880} & \textbf{69.787} & 19.51 & 49.590 & 31.338 & 1.19 & \textbf{40.430} & 2.351 \\
    \hspace{1em} \textbf{+ DUCAT} & 77.74 & 66.85 & 72.295 & 71.886 & 60.75 & 69.245 & 68.200 & \textbf{22.50} & \textbf{50.120} & \textbf{34.900} & \textbf{2.03} & 39.885 & \textbf{3.956} \\
    \midrule
    MART & 77.93 & 64.03 & 70.980 & 70.297 & 58.96 & 68.445 & 67.131 & 20.56 & 49.245 & 32.537 & 0.95 & 39.440 & 1.886 \\
    \hspace{1em} \textbf{+ DUCAT} & \textbf{80.65} & \textbf{65.10} & \textbf{72.875} & \textbf{72.046} & \textbf{60.89} & \textbf{70.770} & \textbf{69.392} & \textbf{24.63} & \textbf{52.640} & \textbf{37.741} & \textbf{7.52} & \textbf{44.085} & \textbf{13.752} \\
    \midrule
    Cons-AT & 83.42 & \textbf{72.59} & 78.005 & 77.627 & 65.55 & 74.485 & 73.413 & 14.80 & 49.110 & 25.142 & 0.43 & 41.925 & 0.863 \\
    \hspace{1em} \textbf{+ DUCAT} & \textbf{89.51} & 72.53 & \textbf{81.020} & \textbf{80.132} & \textbf{67.04} & \textbf{78.275} & \textbf{76.662} & \textbf{28.09} & \textbf{58.800} & \textbf{42.759} & \textbf{7.80} & \textbf{48.655} & \textbf{14.345} \\
    \midrule
    \midrule
    Average & 80.99 & 69.16 & 75.075 & 74.606 & 62.62 & 71.805 & 70.631 & 17.14 & 49.065 & 28.293 & 0.74 & 40.865 & 1.472 \\
    \hspace{1em} \textbf{+ DUCAT} & \textbf{84.18} & \textbf{69.80} & \textbf{76.990} & \textbf{76.316} & \textbf{64.40} & \textbf{74.290} & \textbf{72.973} & \textbf{25.81} & \textbf{54.995} & \textbf{39.502} & \textbf{6.02} & \textbf{45.100} & \textbf{11.237} \\
     & \color{ForestGreen}{+3.9\%} & \color{ForestGreen}{+0.9\%} & \color{ForestGreen}{+2.6\%} & \color{ForestGreen}{+2.3\%} & \color{ForestGreen}{+2.8\%} & \color{ForestGreen}{+3.5\%} & \color{ForestGreen}{+3.3\%} & \color{ForestGreen}{+50.6\%} & \color{ForestGreen}{+12.1\%} & \color{ForestGreen}{+39.6\%} & \color{ForestGreen}{+710.3\%} & \color{ForestGreen}{+10.4\%} & \color{ForestGreen}{+663.2\%} \\
    \bottomrule
    \end{tabular}
    }
\end{table*}

\begin{figure*}[htb]
  \centering
  \begin{subfigure}{0.325\linewidth}
    \centering
    \includegraphics[width=1\linewidth]{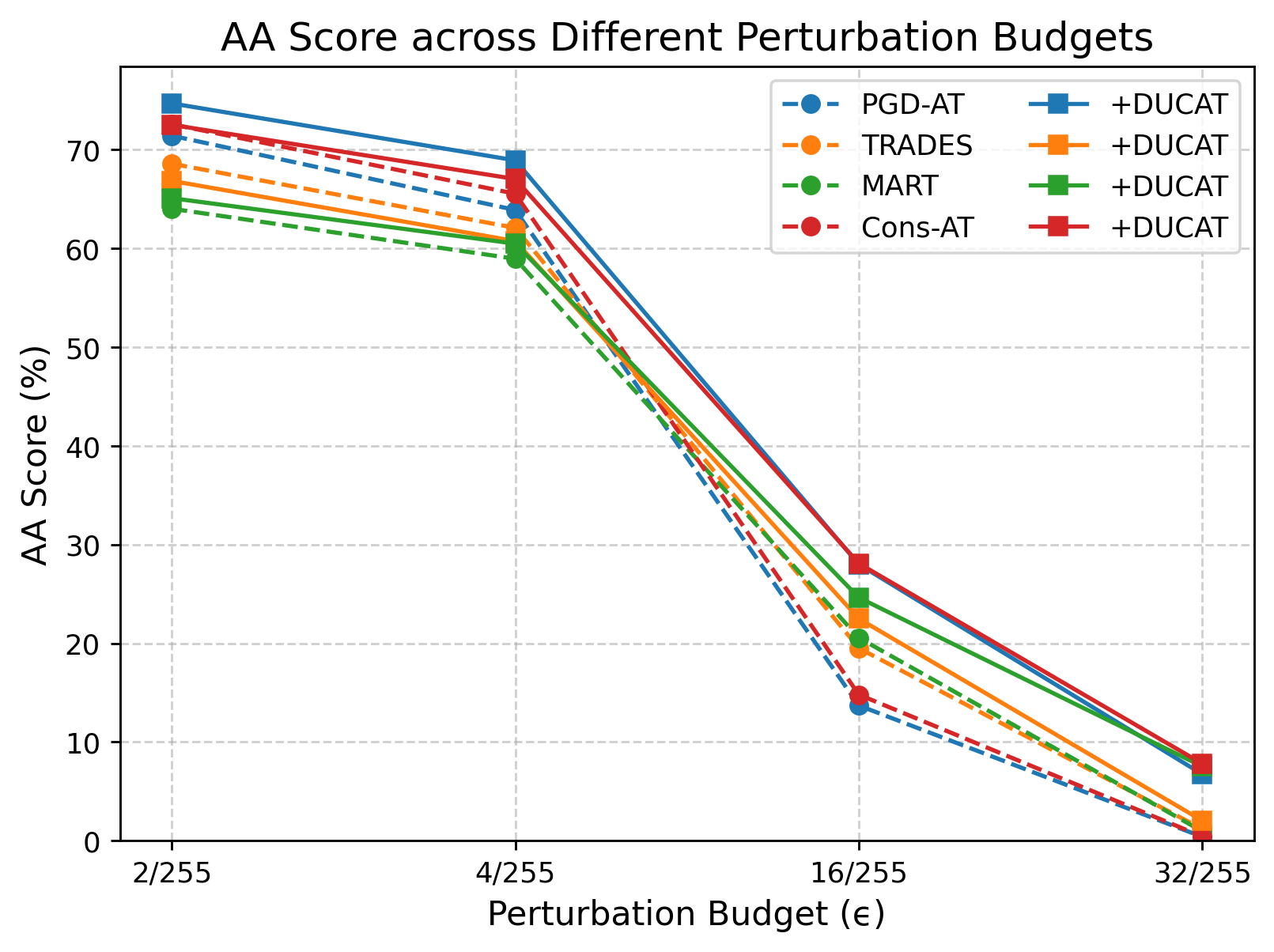}
  \end{subfigure}
  \begin{subfigure}{0.325\linewidth}
    \centering
    \includegraphics[width=1\linewidth]{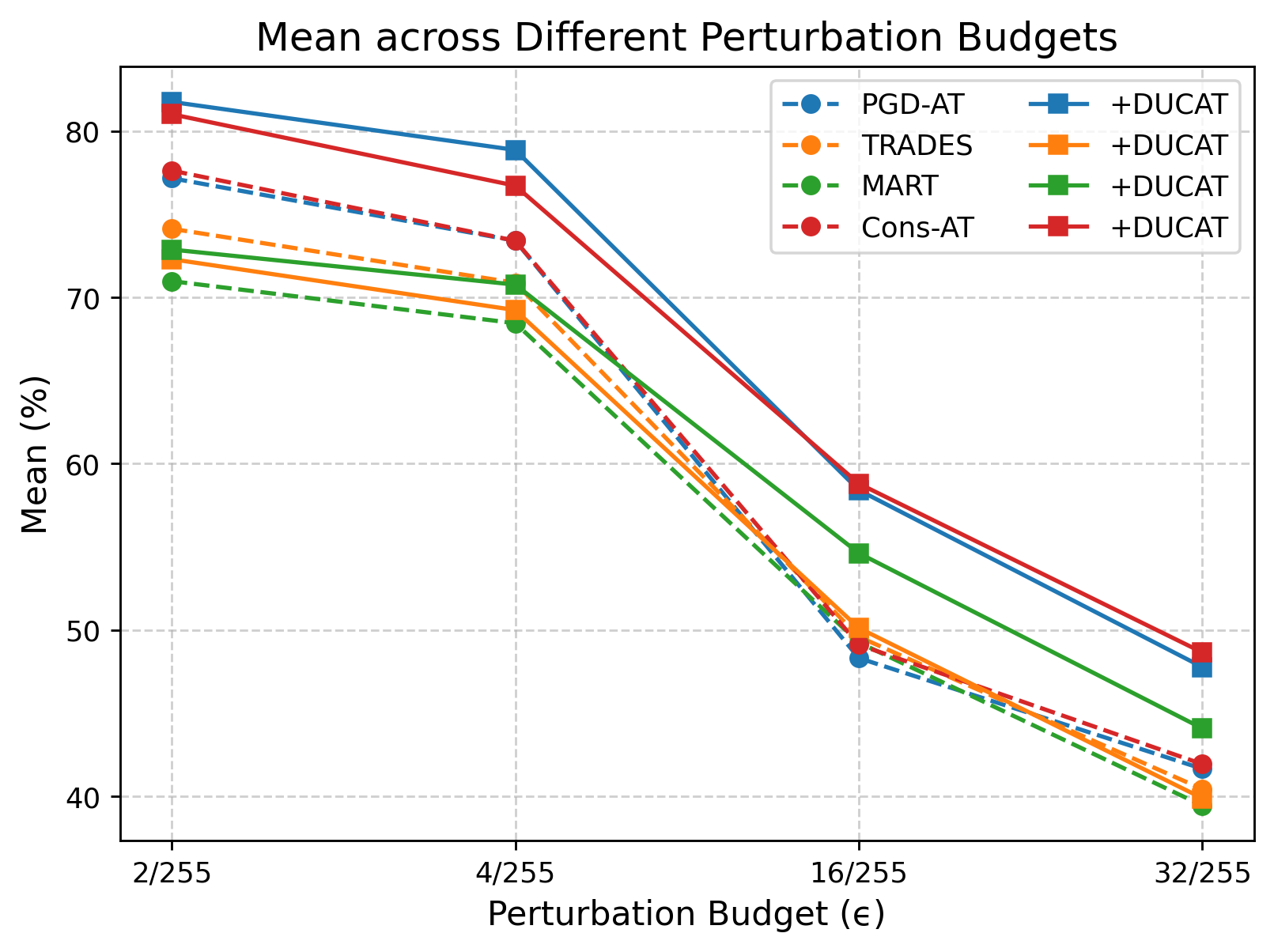}
  \end{subfigure}
  \begin{subfigure}{0.325\linewidth}
    \centering
    \includegraphics[width=1\linewidth]{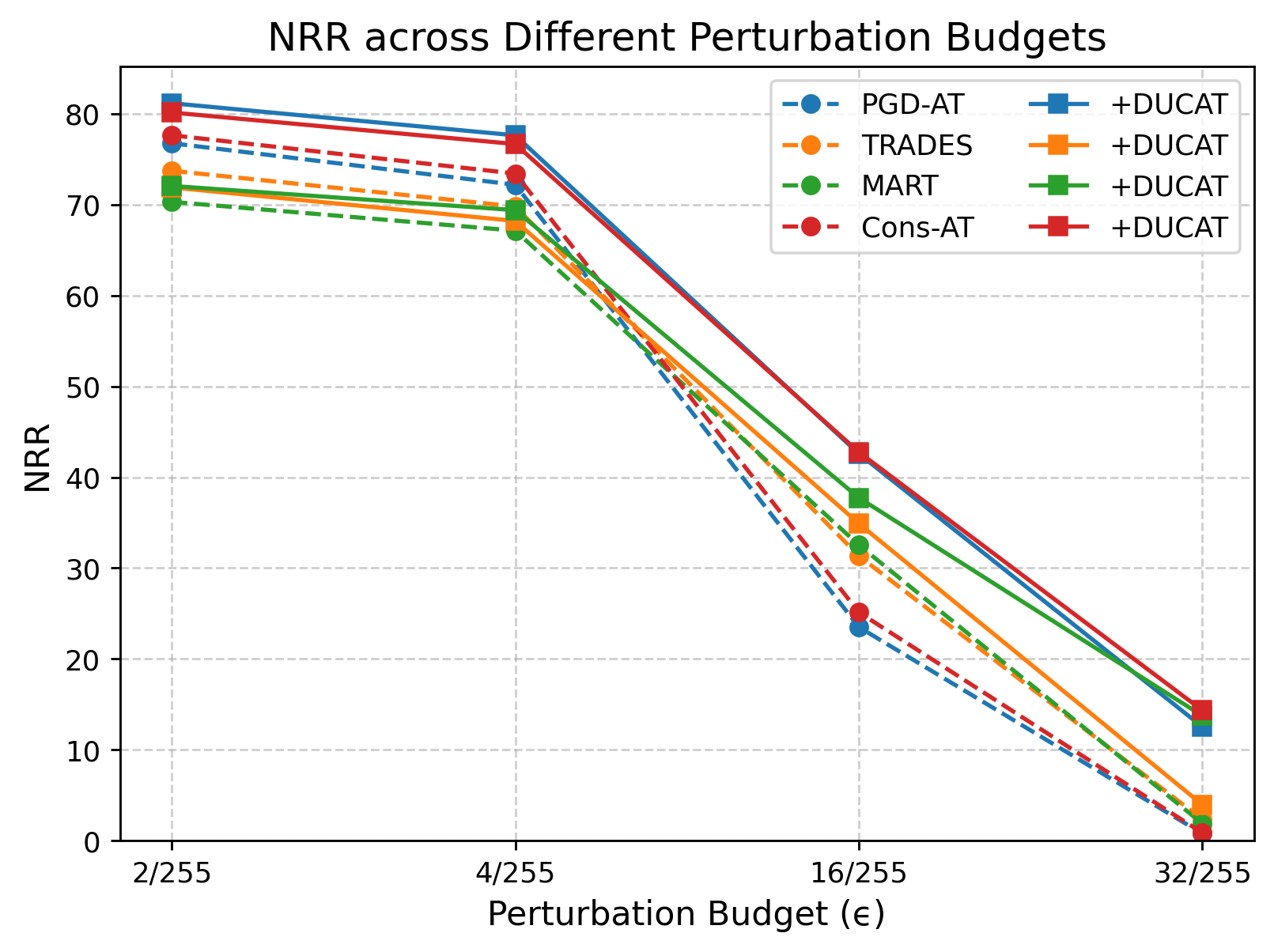}
  \end{subfigure}
  \caption{Corresponding to Table~\ref{tab:budget}, the figures show the change of the evaluation measures with the four different perturbation budgets under $\ell_{\infty}$ adversaries on CIFAR-10 and ResNet-18.}
  \label{fig:budget}
\end{figure*}

For adversary capacity, we assume that full model information is available, so it would also be visible to the adversary that the DNN under DUCAT defense outputs \textit{logits} with length $2C$. However, as standardized by several pioneers in this field, such as \citet{szegedy2014intriguing,goodfellow2015explaining,papernot2017practical}, the strict white-box knowledge includes ``model architecture, parameters, gradients, and training dataset'', while \textbf{excludes} the runtime projection. This is quite reasonable since the runtime projection is not a part of the computational graph of the model, and the specific correspondences between benign and dummy classes are also diverse and underlying. Specifically, it is not necessary to implement them in order such that class $k + C$ is exactly the dummy class of class $k$. In that case, the adversary is not able to know the specific one-to-one correspondences unless inferring them by multiple querying as in black-box situations.

\subsection{Evaluation Measures}\label{subapp:exp_measure}

In the main experiments, five measures were adopted to evaluate the AT methods from three different perspectives. Firstly, we report clean accuracy on benign test data, which reflects the ability of the robust models to maintain performance in benign cases. Secondly, we evaluate adversarial robustness with various adversarial attack methods, including PGD-10 and PGD-100~\citep{madry2018towards}, as well as Auto-Attack (AA)~\citep{croce2020reliable} which consists of four advanced adaptive attacks, namely white-box APGD-CE~\citep{croce2020reliable}, APGD-DLR~\citep{croce2020reliable}, FAB~\citep{croce2020minimally}, and black-box Square Attack~\citep{andriushchenko2020square}. We adopt default attack settings aligning with Appendix~\ref{subapp:threat} and random starts. Finally, to test the performance \textit{w.r.t.} the current accuracy-robustness trade-off problem in AT, we report the mean of the clean accuracy and the robust score (mainly under Auto-Attack, with exceptional cases specifically stated), and also introduce the Natural-Robustness Ratio (NRR)~\citep{gowda2024conserve}:
\begin{equation}
\textit{NRR} = \frac{2 \times \textit{Clean Accuracy} \times \textit{Adversarial Robustness}}{\textit{Clean Accuracy} + \textit{Adversarial Robustness}},
\end{equation}
which emphasizes how well the robust model strikes a balance between clean accuracy and robustness. A lower NRR value suggests that the model might prioritize one side over the other. All the results reported are averages of three runs, with the ``best'' performance of each run recorded on the best checkpoint achieving the highest PGD-10 accuracy (\ie, the selection is only based on the training adversary), which ensures that the whole training process is isolated from the test adversaries for evaluating the real robustness.

\begin{table*}[htb]
\caption{The supplementary results under $\ell_{2}$ adversary rather than the main $\ell_{\infty}$ one on CIFAR-10 and ResNet-18. All results are acquired from three runs.}
  \label{tab:l2}
  \centering
  \renewcommand\arraystretch{1.1}
    \begin{tabular}{lcccc}
    \toprule
    \hspace{0.7em} \textbf{Method} & \textbf{Clean} & \textbf{AA} & \textbf{Mean} & \textbf{NRR} \\
    \midrule
    \midrule
    PGD-AT & 82.92 & 58.18 & 70.550 & 68.383 \\
    \hspace{1em} \textbf{+ DUCAT} & \textbf{88.81} & \textbf{66.58} & \textbf{77.695} & \textbf{76.108} \\
    \midrule
    TRADES & \textbf{79.67} & \textbf{56.43} & \textbf{68.050} & \textbf{66.069} \\
    \hspace{1em} \textbf{+ DUCAT} & 77.74 & 55.44 & 66.590 & 64.720 \\
    \midrule
    MART & 77.93 & 57.44 & 67.685 & 66.136 \\
    \hspace{1em} \textbf{+ DUCAT} & \textbf{80.65} & \textbf{68.13} & \textbf{74.390} & \textbf{73.865} \\
    \midrule
    Cons-AT & 83.42 & 59.36 & 71.390 & 69.361 \\
    \hspace{1em} \textbf{+ DUCAT} & \textbf{89.51} & \textbf{67.06} & \textbf{78.285} & \textbf{76.676} \\
    \midrule
    \midrule
    Average & 80.99 & 57.85 & 69.420 & 67.493 \\
    \hspace{1em} \textbf{+ DUCAT} & \textbf{84.18} & \textbf{64.30} & \textbf{74.240} & \textbf{72.910} \\
    & \color{ForestGreen}{+3.9\%} & \color{ForestGreen}{+11.2\%} & \color{ForestGreen}{+6.9\%} & \color{ForestGreen}{+8.0\%} \\
    \bottomrule
    \end{tabular}
\end{table*}

\begin{figure*}[htb]
  \centering
  \begin{subfigure}{0.325\linewidth}
    \centering
    \includegraphics[width=1\linewidth]{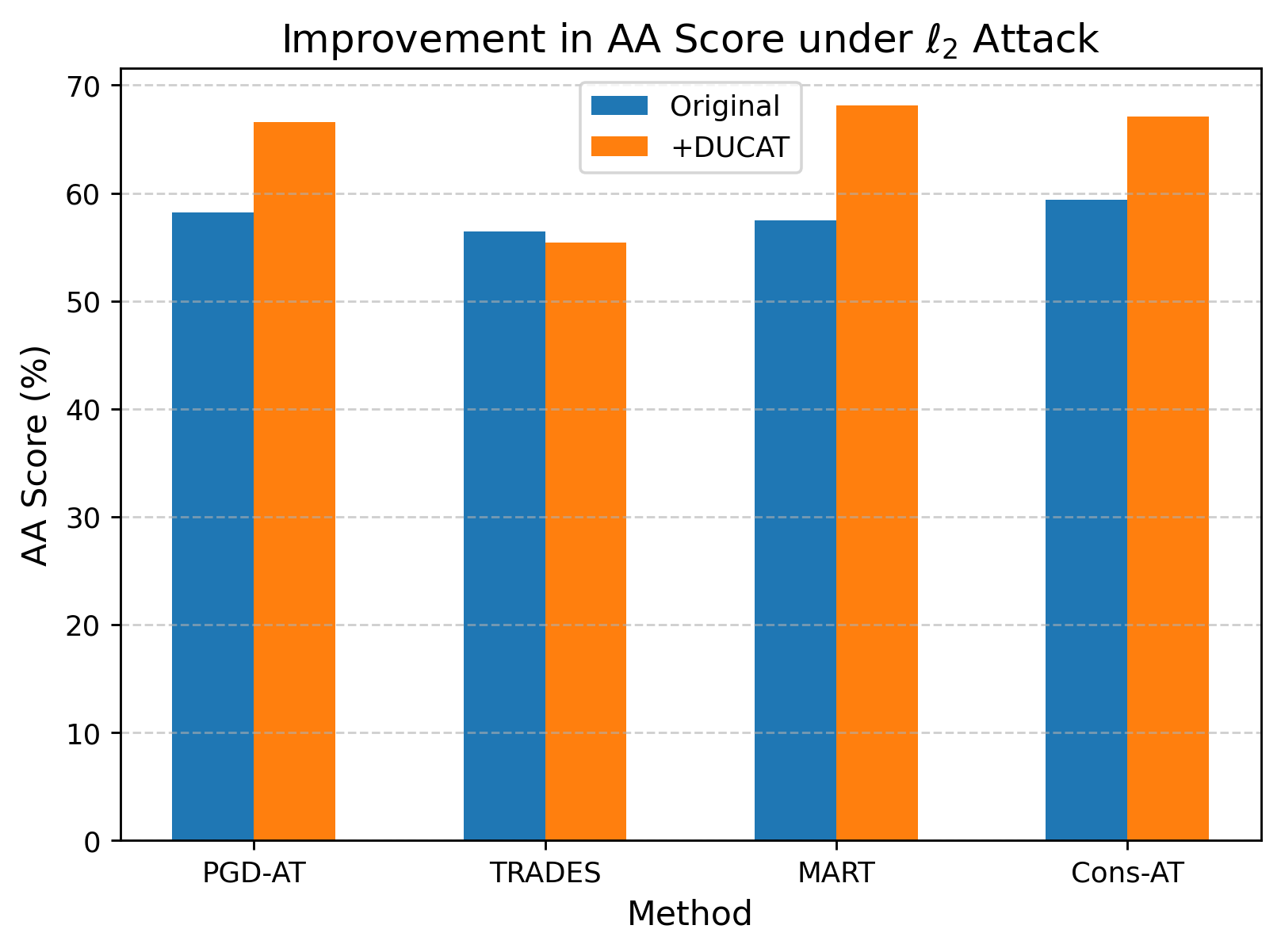}
  \end{subfigure}
  \begin{subfigure}{0.325\linewidth}
    \centering
    \includegraphics[width=1\linewidth]{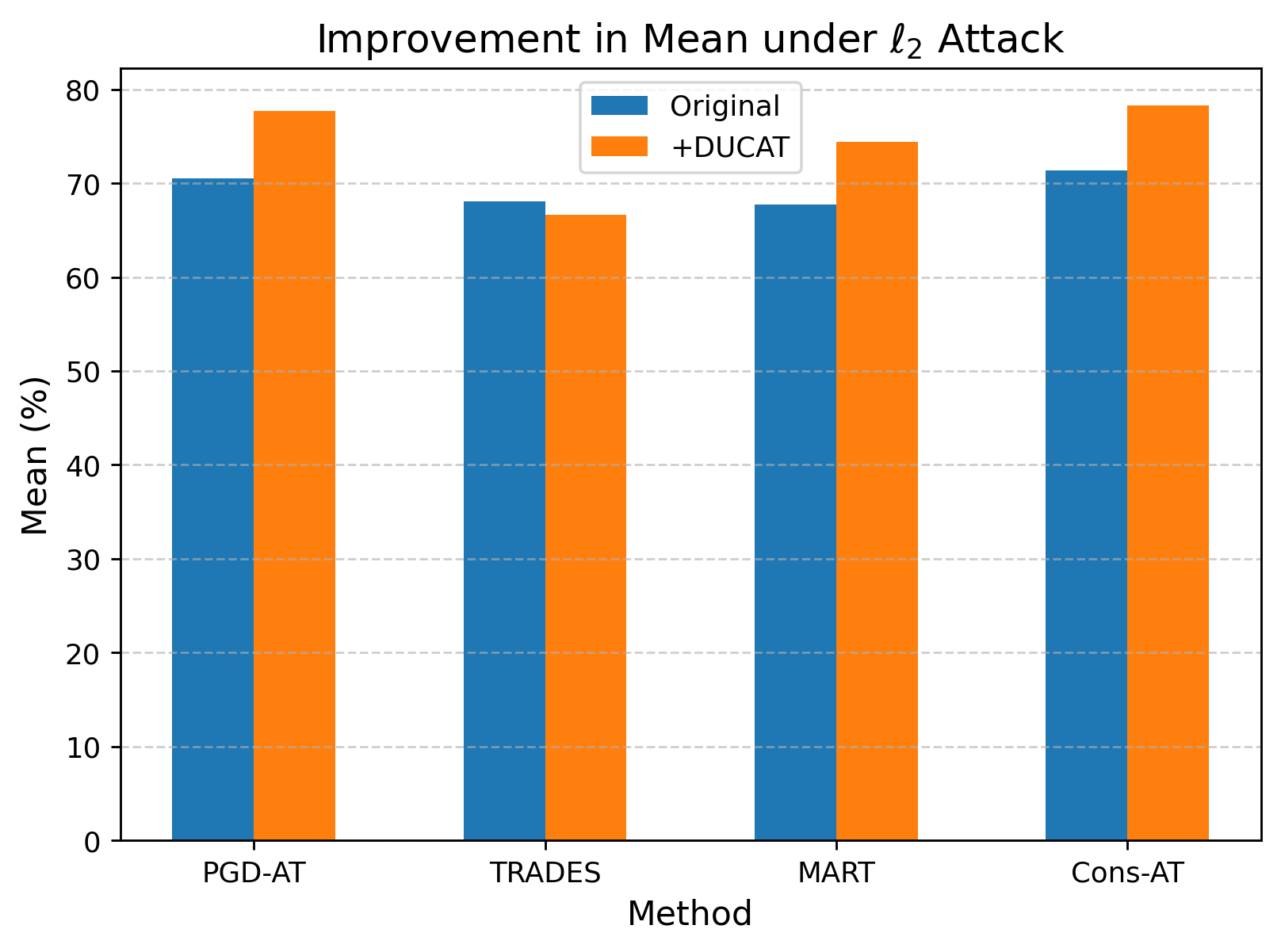}
  \end{subfigure}
  \begin{subfigure}{0.325\linewidth}
    \centering
    \includegraphics[width=1\linewidth]{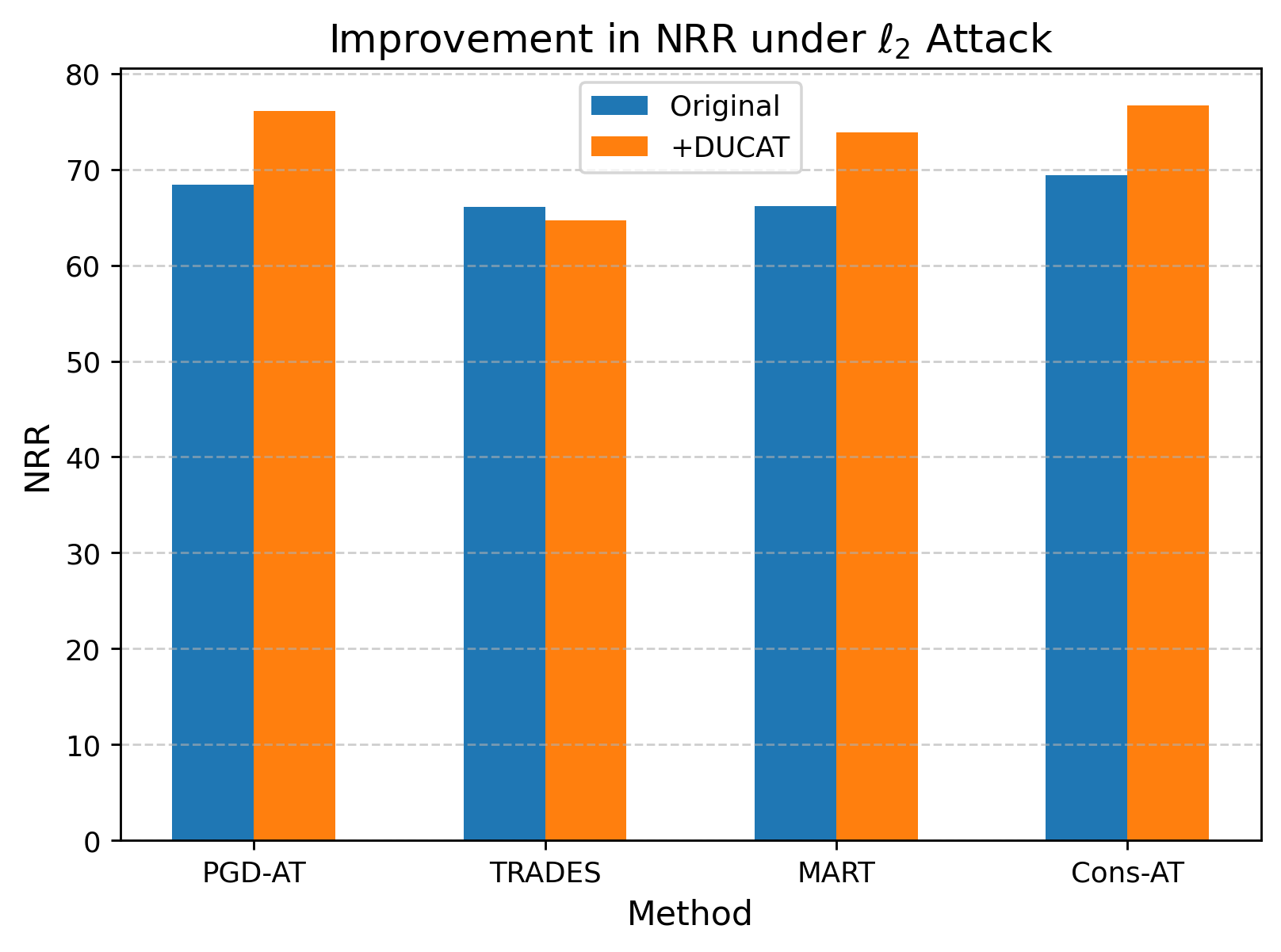}
  \end{subfigure}
  \caption{Corresponding to Table~\ref{tab:l2}, the figures show the improvement in the evaluation measures under $\ell_{2}$ adversary rather than the main $\ell_{\infty}$ one on CIFAR-10 and ResNet-18.}
  \label{fig:l2}
\end{figure*}

\section{Generalization to Other Threat Models}\label{app:other}

Beyond the threat model for the main experiments given in Appendix~\ref{subapp:threat}, this section additionally considers three different categories of threat models, aiming at demonstrating the generalization ability of the proposed DUCAT method. Specifically, four perturbation budgets larger or smaller than the ones in the main threat model are further tested in Appendix~\ref{subapp:budget}. Also, adversaries that produce $\ell_{2}$ perturbations or adopt targeted attacks rather than $\ell_{\infty}$ perturbations and untargeted attacks as in the main experiments are considered in Appendix~\ref{subapp:l2} and Appendix~\ref{subapp:target}, respectively. The supplementary experimental results confirm the \textbf{wide effectiveness of our DUCAT method under all these additional threat models}, which strongly supports its \textbf{real-world generalization}.

\subsection{Generalization to Various Perturbation Budgets}\label{subapp:budget}

\begin{table*}[htb]
  \caption{Additional results on targeted PGD-10 adversary compared with the default untargeted one. The ``(ori)'' / ``(du)'' respectively suggest that the target classes are randomly selected from the original / dummy classes. All results are acquired from three runs.}
  \label{tab:targeted}
  \centering
  \renewcommand\arraystretch{1.3}
  \resizebox{13.9cm}{!}{
  \begin{tabular}{lcccccccccc}
    \toprule
    \multirow{2.5}{*}{\hspace{0.7em} \textbf{Method}} & \multirow{2.5}{*}{\textbf{Clean}} & \multicolumn{3}{c}{\textbf{Untargeted PGD-10}} & \multicolumn{3}{c}{\textbf{Targeted PGD-10 (ori)}} & \multicolumn{3}{c}{\textbf{Targeted PGD-10 (du)}} \\
    \cmidrule(r){3-5} \cmidrule(r){6-8} \cmidrule(r){9-11}
     &  & \textbf{Robust} & \textbf{Mean} & \textbf{NRR} & \textbf{Robust} & \textbf{Mean} & \textbf{NRR} & \textbf{Robust} & \textbf{Mean} & \textbf{NRR} \\
    \midrule
    \midrule
    PGD-AT & 82.92 & 51.81 & 67.365 & 63.773 & \textbf{73.02} & 77.970 & 77.656 & - & - & - \\
    \hspace{1em} \textbf{+ DUCAT}  & \textbf{88.81} & \textbf{65.10} & \textbf{76.955} & \textbf{75.129} & 70.23 & \textbf{79.520} & \textbf{78.435} & 70.77 & \textbf{79.790} & \textbf{78.770} \\
    \midrule
    TRADES & \textbf{79.67} & 52.14 & \textbf{65.905} & \textbf{63.030} & 72.68 & \textbf{76.175} & \textbf{76.015} & - & - & - \\
    \hspace{1em} \textbf{+ DUCAT}  & 77.74 & \textbf{52.66} & 65.200 & 62.788 & \textbf{73.63} & 75.685 & 75.629 & \textbf{73.32} & 75.530 & 75.465 \\
    \midrule
    MART & 77.93 & 53.61 & 65.770 & 63.522 & 70.48 & 74.205 & 74.018 & - & - & - \\
    \hspace{1em} \textbf{+ DUCAT}  & \textbf{80.65} & \textbf{58.42} & \textbf{69.535} & \textbf{67.758} & \textbf{73.87} & \textbf{77.260} & \textbf{77.111} & \textbf{72.89} & \textbf{76.770} & \textbf{76.574} \\
    \midrule
    Cons-AT & 83.42 & 53.20 & 68.310 & 64.968 & \textbf{74.26} & 78.840 & 78.574 & - & - & - \\
    \hspace{1em} \textbf{+ DUCAT}  & \textbf{89.51} & \textbf{66.83} & \textbf{78.170} & \textbf{76.525} & 72.41 & \textbf{80.960} & \textbf{80.057} & 72.80 & \textbf{81.155} & \textbf{80.295} \\
    \midrule
    \midrule
    Average & 80.99 & 52.69 & 66.838 & 63.843 & \textbf{72.61} & 76.798 & 76.569 & - & - & - \\
     \hspace{1em} \textbf{+ DUCAT} & \textbf{84.18} & \textbf{60.75} & \textbf{72.465} & \textbf{70.572} & 72.54 & \textbf{78.356} & \textbf{77.924} & 72.45 & \textbf{78.311} & \textbf{77.872} \\
     & \color{ForestGreen}{+3.9\%} & \color{ForestGreen}{+15.3\%} & \color{ForestGreen}{+8.4\%} & \color{ForestGreen}{+10.5\%} & \color{Red}{-0.1\%} & \color{ForestGreen}{+2.0\%} & \color{ForestGreen}{+1.8\%} & \color{Red}{-0.2\%} & \color{ForestGreen}{+2.0\%} & \color{ForestGreen}{+1.7\%} \\
    \bottomrule
  \end{tabular}
  }
\end{table*}

This section investigates the generalization of the proposed DUCAT method under four different perturbation budgets, respectively $\epsilon=$ 2/255, 4/255, 16/255 and 32/255, of the $\ell_{\infty}$ threat model on CIFAR-10 and ResNet-18. As reported in Table~\ref{tab:budget} and illustrated in Figure~\ref{fig:budget}, the experimental results demonstrate a consistent and notable improvement by DUCAT across all perturbation levels. Specifically, under smaller perturbation budgets (\ie, $\epsilon=$ 2/255 and 4/255), DUCAT exhibits relatively moderate yet consistent gains in robustness and the trade-off, showing the same trend as in the main experiments. More notably, as the perturbation budget expands to more challenging values (\ie, $\epsilon=$ 16/255 and 32/255), the advantages of our DUCAT become more significant. In particular, at $\epsilon=32/255$, the robustness of the benchmarks has been degraded close to 0, while DUCAT still retains a certain defensive capability. As a result, on average, DUCAT significantly boosts AA robustness from $0.74\%$ to $6.02\%$, an approximate 7-fold increase, and enhances NRR from $1.47\%$ to $11.24\%$, marking a substantial improvement of $663.2\%$. Despite the huge improvement, such a result is within our expectation because the existence of dummy classes enables the model to better learn distinguished features of benign and adversarial samples respectively, instead of just remembering their ``standardized'' differences in perturbations. Thus, the defense mechanism of DUCAT under unseen perturbation budgets is more like a generalization of model perception, instead of a strict tolerance that is, to some extent, bounded by the specific training perturbation budget. Such findings clearly validate the effectiveness and superior generalization capability of DUCAT, strongly supporting its potential applicability in real-world diverse adversarial environments.

\subsection{Generalization to $\ell_{2}$ Adversary}\label{subapp:l2}

In this section, we further evaluate the generalization of DUCAT under adversarial attacks constrained by $\ell_2$ norm rather than $\ell_\infty$ norm adopted by default. 
As illustrated in Table~\ref{tab:l2} and Figure~\ref{fig:l2}, the experimental results also indicate significant performance enhancements brought by DUCAT. Specifically, under AA evaluation, DUCAT achieves an average robustness increase of $11.2\%$ from $57.85\%$ to $64.30\%$, showcasing remarkable robustness improvements across diverse benchmark methods. Additionally, the Mean and NRR respectively improve by around $6.9\%$ and $8.0\%$ with DUCAT integrated, indicating a better trade-off between accuracy and robustness. These outcomes demonstrate that DUCAT could also achieve substantial generalization against different norms chosen by the adversary to produce attacks. Besides, while acknowledging that the result in TRADES is not as desirable as in other benchmarks, this is predictable and not a serious problem as discussed in Appendix~\ref{subapp:trades+ducat}.

\subsection{Generalization to Targeted Attacks}\label{subapp:target}

Someone may wonder, as the DUCAT defense is visible to white-box adversaries through the double-size last layer, whether they can do something to bypass this defense. One idea that might be representative is, given that DUCAT benefits from ``inducing'' adversaries to perturb benign samples to the one-to-one corresponding dummy classes (\ie, in other words, each dummy class serves as the easiest attack target for the benign samples from the corresponding original class) to run-time detect and recover them, if it is possible to change from untargeted adversaries to targeted ones, so that to release such one-to-one correspondences by compulsively appointing other attack targets. However, in this part, we demonstrate that no matter randomly appointing different original classes or dummy ones as the attack targets, the targeted PGD-10 adversary is still not sufficiently dangerous under the DUCAT defense. As the results shown in Table~\ref{tab:targeted}, probably due to the original difficulty of targeted attacks compared to untargeted ones~\citep{rathore2020untargeted,cai2023ensemble}, the effectiveness of targeted PGD-10 is even worse than untargeted PGD-10. 

\begin{figure*}[htb]
  \centering
  \begin{subfigure}{0.468\linewidth}
    \centering
    \includegraphics[width=1\linewidth]{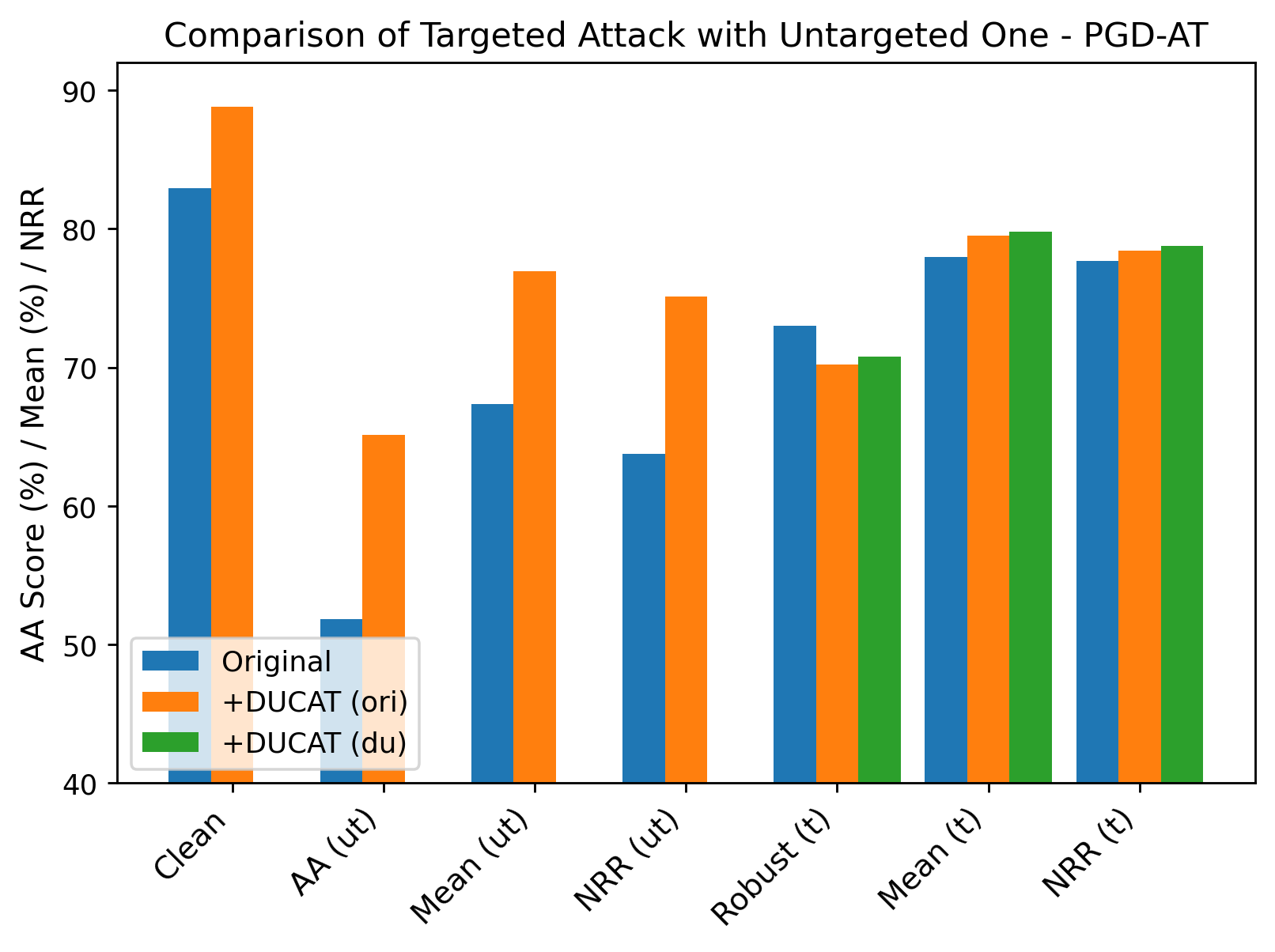}
  \end{subfigure}
  \begin{subfigure}{0.468\linewidth}
    \centering
    \includegraphics[width=1\linewidth]{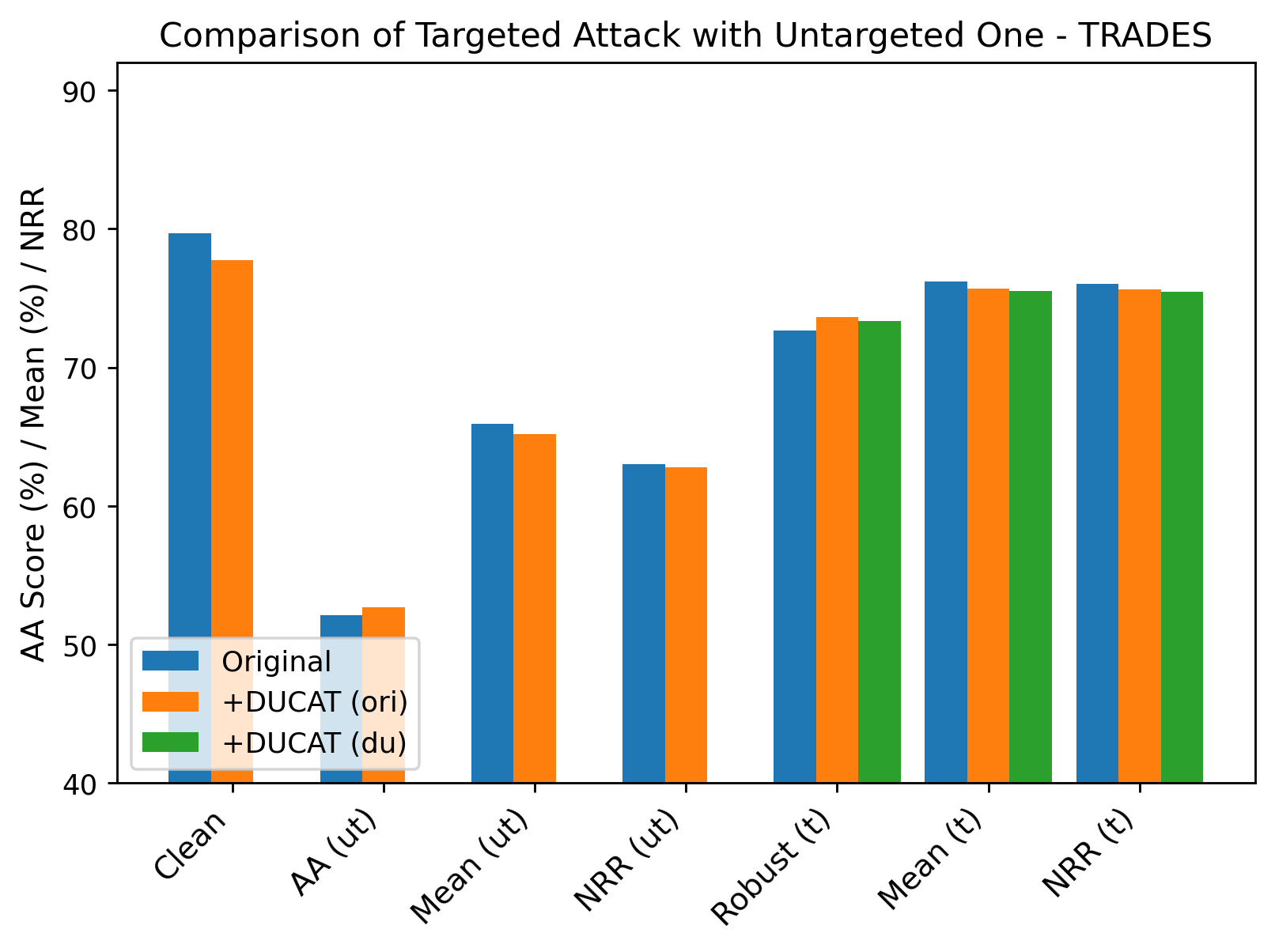}
  \end{subfigure}
  \begin{subfigure}{0.468\linewidth}
    \centering
    \includegraphics[width=1\linewidth]{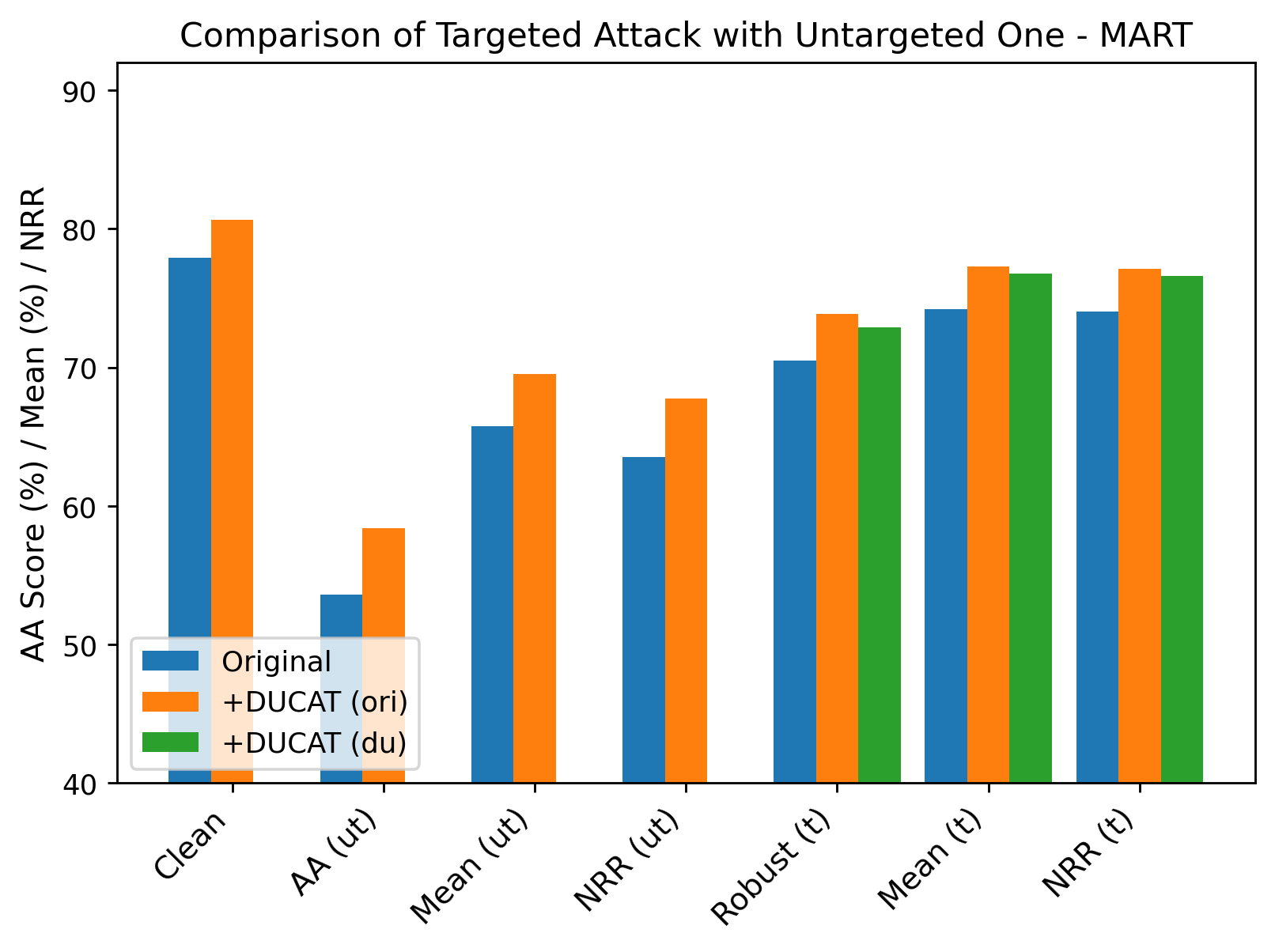}
  \end{subfigure}
  \begin{subfigure}{0.468\linewidth}
    \centering
    \includegraphics[width=1\linewidth]{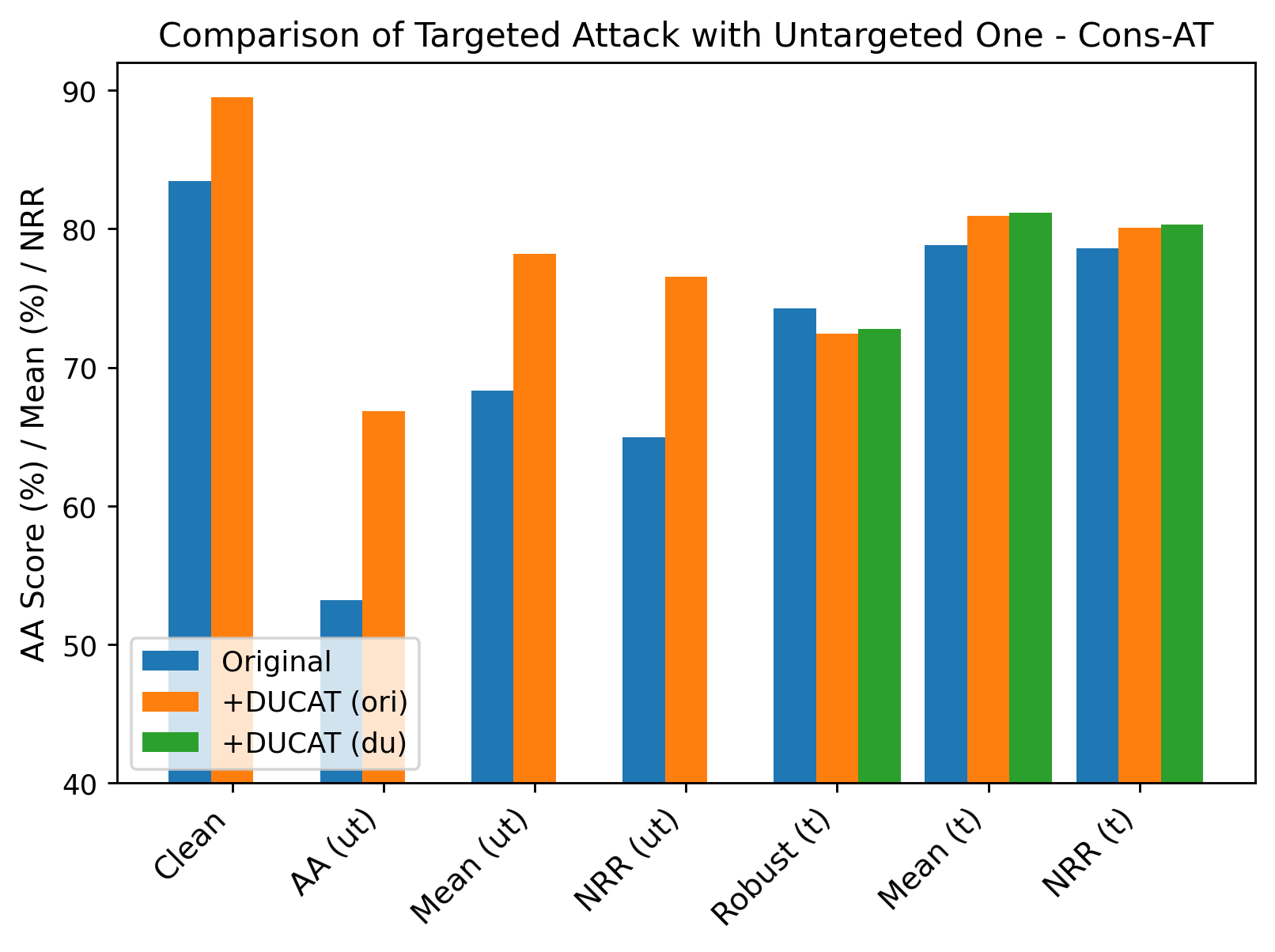}
  \end{subfigure}
  \caption{Corresponding to Table~\ref{tab:targeted}, the figures illustrate the additional results on the targeted PGD-10 adversary compared with the default untargeted one in the four benchmarks. The ``(ut)'' / ``(t)'' respectively indicate ``untargeted'' and ``targeted'' adversaries, and similarly, the ``(ori)'' / ``(du)'' respectively suggest random target selection from the original / dummy classes.}
  \label{fig:targeted}
\end{figure*}

Our results confirm two notable points. First, even in the worst case under the most powerful adversary among the possible ones, our DUCAT can still achieve an improvement in clean accuracy of about 4\% while maintaining a similar level in robustness on average of the four benchmarks, which ends up with about 2\% advancement in the trade-off problem as indicated by Mean and NRR. While acknowledging that it is not as impressive as in the untargeted threat model, such improvements are still non-trivial. Second, although the targeted threat model is theoretically expected to degrade our DUCAT compared with the untargeted one, due to the reduced target range and increased perturbation difficulty~\citep{rathore2020untargeted,cai2023ensemble}, specifically adopting targeted attacks against DUCAT could not further degrade the worst practical robustness achieved by untargeted attacks. In other words, for any real-world adversaries aiming at attacking DUCAT-defended models, simply adopting untargeted attacks would still be more dangerous, under which situation our advantages are more impressive.

\section{Comparison with Secondary Related Works}\label{app:comp}

In this section, we supplement additional comparisons with two kinds of secondary related works, namely the synthetic data-based AT methods in Appendix~\ref{subapp:synthetic}, and an inference time method that also acquires robustness beyond the conventional AT paradigm in Appendix~\ref{subapp:MI}. Although these methods \textbf{do not directly compete with the proposed DUCAT as they focus on completely different perspectives} from ours (\ie, it is possible to employ DUCAT \textbf{together} with them), we respect their contributions to the same accuracy-robustness trade-off problem, and conduct some basic experiments to facilitate a more comprehensive understanding for readers.

\begin{table*}[ht]
\caption{DUCAT with the open-source 1M CIFAR-10 synthetic data from \citet{wang2023better} can achieve better clean accuracy and competitive robustness compared to it under significantly lighter training settings. The superscript $*$ is to indicate the involvement of the additional synthetic data.}\label{tab:synthetic}
  \centering
  \renewcommand\arraystretch{1.1}
  \begin{tabular}{ccccc}
  \toprule
  \textbf{Method} & \textbf{Clean} & \textbf{AA} & \textbf{Mean} & \textbf{NRR} \\
  \midrule
  \citet{wang2023better} & 91.12 & \textbf{63.35} & 77.235 & 74.739 \\
  \textbf{Ours (PGD-AT + DUCAT)$^*$}  & \textbf{92.60} & 62.74 & \textbf{77.670} & \textbf{74.800} \\
  \bottomrule
  \end{tabular}
\end{table*}

\begin{table*}[ht]
  \caption{Comparison with another work aiming at acquiring adversarial robustness beyond the conventional AT further showcases the advantage of our DUCAT. 
  }\label{tab:MI}
  \centering
  \renewcommand\arraystretch{1.1}
  \begin{tabular}{ccccc}
  \toprule
  \textbf{Method} & \textbf{Clean} & \textbf{PGD-10} & \textbf{Mean} & \textbf{NRR} \\
  \midrule
  MI~\citep{pang2020mixup} & 84.20 & 64.50 & 74.350 & 73.045 \\
  \textbf{Ours (PGD-AT + DUCAT)}  & \textbf{88.81} & \textbf{65.10} & \textbf{76.955} & \textbf{75.129} \\
  \bottomrule
  \end{tabular}
\end{table*}

\subsection{Comparison with Synthetic Data-Based AT}\label{subapp:synthetic}

In the main experiments, we exclude related works with synthetic data, such as \citet{rebuffi2021fixing}, \citet{sehwag2022robust} and \citet{wang2023better}, because we think 1) there are certain concerns about their fairness compared with conventional AT, and 2) they focus more on data rather than algorithms, thus are not in direct competition with AT algorithms but can be easily integrated into them.

For fairness, although different from those extra data-based methods, synthetic data-based ones seem to fairly use the same training dataset as conventional AT, the problem is the dramatic additional computational cost they rely on. For instance, they need to train a diffusion model first~\citep{rebuffi2021fixing,sehwag2022robust,wang2023better}, which is even much more time-consuming than training the robust model itself (e.g., training a ResNet-18 with PGD-AT on CIFAR-10 typically needs only about 4 hours with RTX4090, while training a DDPM (a common types of diffusion) model on CIFAR-10 usually needs up to 30 hours with A100). Although it is argued that both training generative models and sampling from them is a one-time cost~\citep{sehwag2022robust}, that is only true for experimental scenarios with standard datasets, yet certainly not for specific real-world practices. Also, synthetic data-based methods rely on extra standard-trained models for pseudo-labeling of the generated unlabeled data~\citep{rebuffi2021fixing,sehwag2022robust,wang2023better}. Besides, their training epochs and batch size are significantly larger than typical AT~\citep{rebuffi2021fixing,sehwag2022robust,wang2023better}, which respectively means the need for more training time and the VRAM resource. We should always take these costs into consideration when adopting synthetic data-based methods.

Sometimes, a more important philosophy behind a fairness assumption is that, methods following the assumption would not be directly in competition with the ones not, and here is such a case. While conventional AT methods focus more on the advancement in algorithms, synthetic data-based methods contribute more to the data aspect. As a result, these two kinds of works would not degrade the contributions of each other but could be integrated easily. Actually, \citet{rebuffi2021fixing}, \citet{sehwag2022robust} and \citet{wang2023better} all involve conventional AT. More specifically, all of them are essentially TRADES plus different data augmentation approaches with diffusion models. In Table~\ref{tab:synthetic}, we preliminarily showcase that by simply using the open-source 1M CIFAR-10 synthetic data from \citet{wang2023better} without other tricks it adopts (e.g., weight averaging and cyclic learning rate schedule) and the more training epochs it needs (i.e., at least 400 epochs), we can easily train a WideResNet-28-10 achieving better clean accuracy and competitive robust performance compared with \citet{wang2023better}, by DUCAT with just the same training details of our main experiments. Considering the simplicity and less computational cost of DUCAT, this preliminary result is itself considerable, while there are more promising directions like directly integrating DUCAT into synthetic data-based methods to be further explored.

\subsection{Comparison with Another Novel Idea beyond Conventional AT}\label{subapp:MI}

As DUCAT suggests a novel paradigm to acquire adversarial robustness beyond conventional AT, there are also a few relevant previous works from this perspective. Representatively, \citet{pang2020mixup} develops Mixup Inference (MI) to mixup input with random clean samples at inference time, thus shrinking and transferring the equivalent perturbation if the input is adversarial. Although as an inference time method, MI does not directly compete with DUCAT, it is also a notable attempt at new possible directions for adversarial robustness. Thus, we additionally provide a performance comparison between DUCAT and MI on the CIFAR-10 dataset. Specifically, we follow the same settings of our main experiments for DUCAT, while adopt the performance originally reported in \citet{pang2020mixup} for MI. As the robust performance of MI is not evaluated by AA there, for fairness, we compare robust accuracy under the PGD-10 adversary below, as both the two works involve it. The comparison results are shown in Table~\ref{tab:MI}, where DUCAT outperforms MI though given that the results of DUCAT are acquired from ResNet-18 while those of MI are from more powerful ResNet-50.

\section{Ablation Study}\label{app:ablation}

\begin{figure*}[htb]
  \centering
  \begin{subfigure}{0.42\linewidth}
    \centering
    \includegraphics[width=0.923\linewidth]{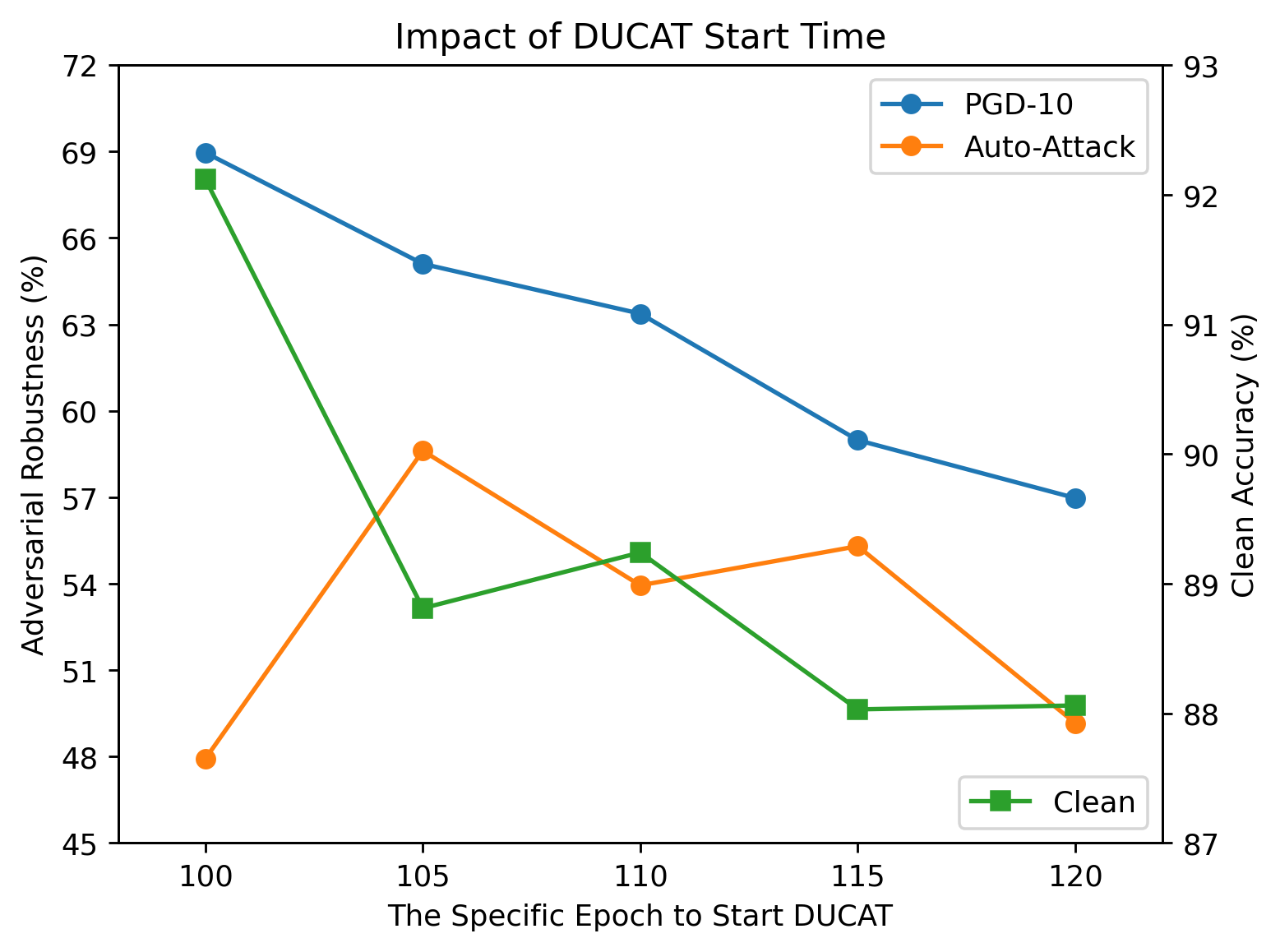}
    \caption{DUCAT start epoch $t$.}\label{fig:ablation1}
  \end{subfigure}
  \begin{subfigure}{0.42\linewidth}
    \centering
    \includegraphics[width=0.923\linewidth]{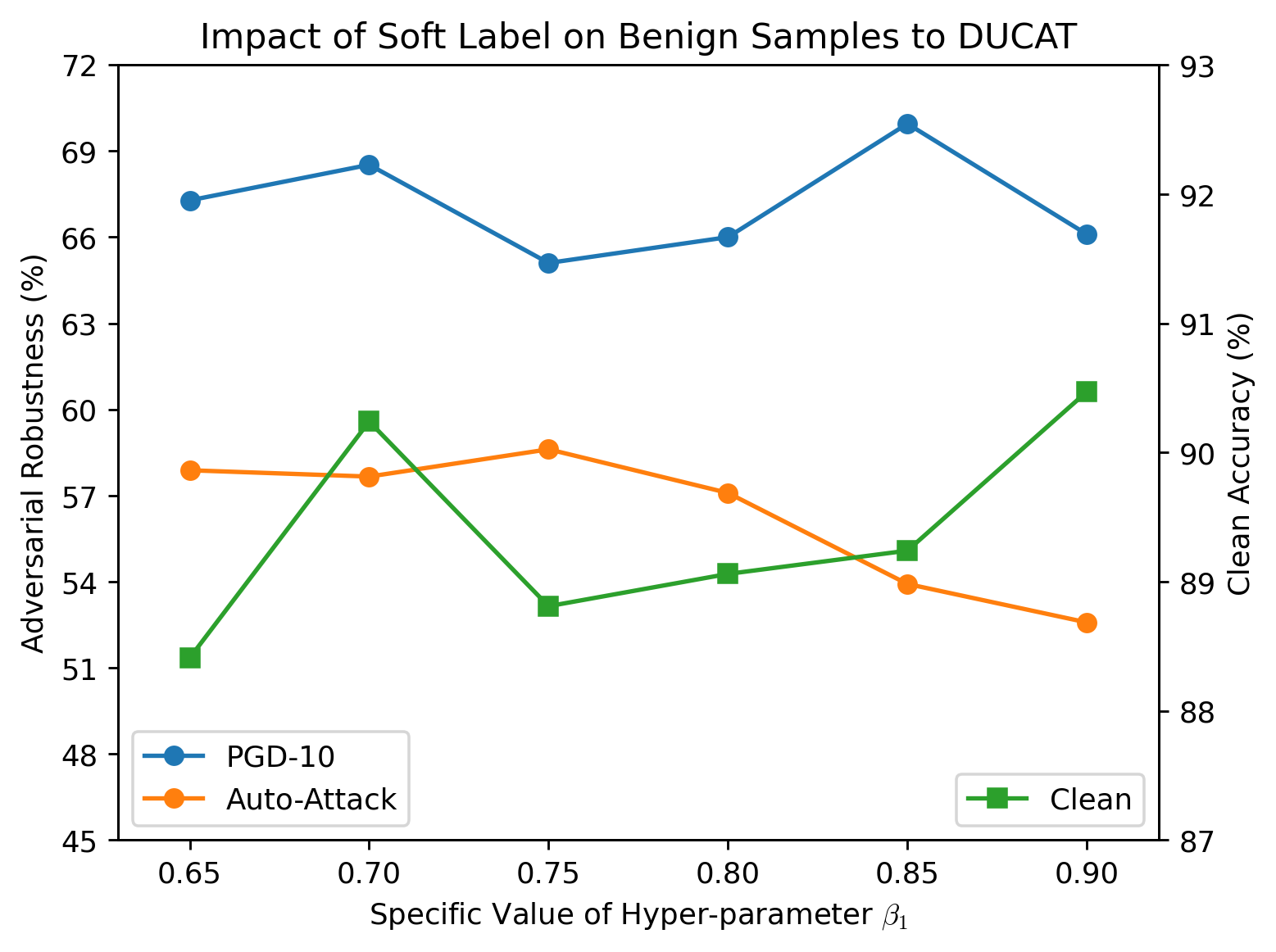}
    \caption{Benign soft label weight $\beta_1$.}\label{fig:ablation2}
  \end{subfigure}
  \begin{subfigure}{0.42\linewidth}
    \centering
    \vspace{0.5em}
    \includegraphics[width=0.923\linewidth]{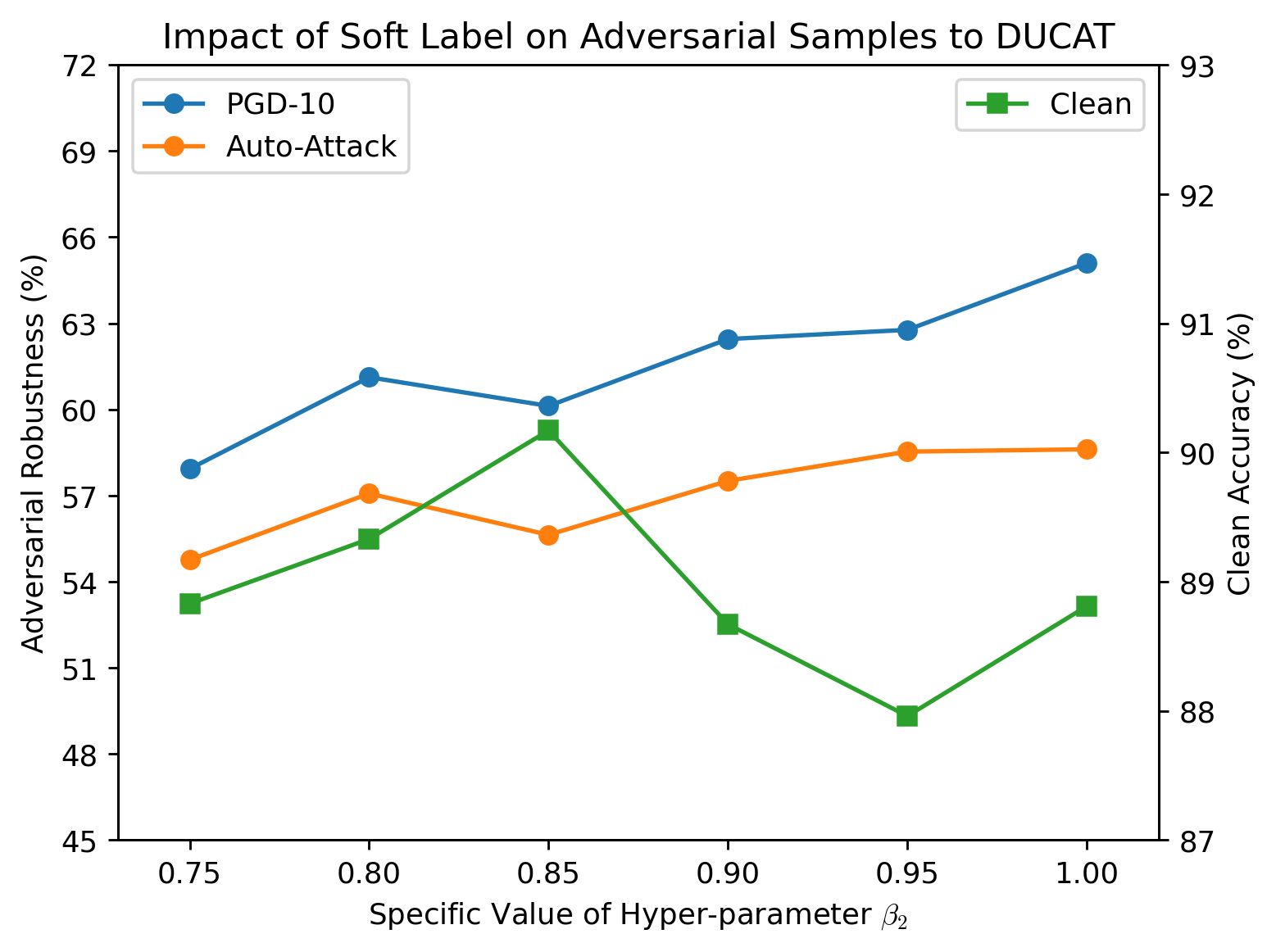}
    \caption{Adversarial soft label weight $\beta_2$.}\label{fig:ablation3}
  \end{subfigure}
  \begin{subfigure}{0.42\linewidth}
    \centering
    \vspace{0.5em}
    \includegraphics[width=0.923\linewidth]{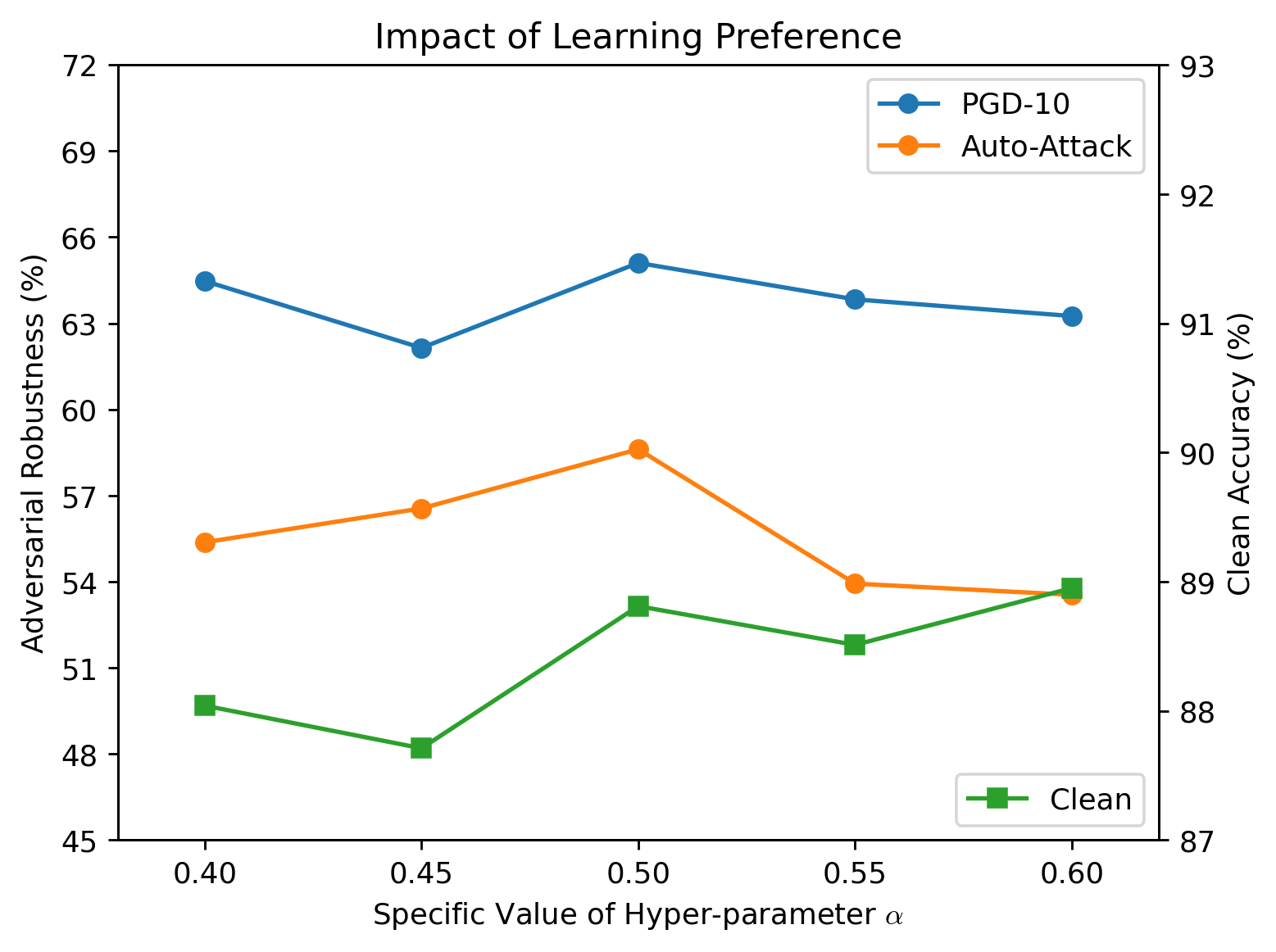}
    \caption{Explicit trade-off preference $\alpha$.}\label{fig:ablation4}
  \end{subfigure}
  \caption{Comprehensive ablation studies of the proposed DUCAT on CIFAR-10 and ResNet-18 \textit{w.r.t.} four hyper-parameters, namely $t$, $\beta_1$, $\beta_2$ and $\alpha$. The y-axises are aligned across the subfigures.}
  \label{fig:ablation}
\end{figure*}

Provided the practical simplicity of the proposed DUCAT, there are mainly four hyper-parameters that impact the trade-off between accuracy and robustness for different reasons, namely the specific epoch $t$ to start DUCAT, and the $\beta_1$, $\beta_2$, $\alpha$ in Equation~(\ref{eq:risk}). As illustrated in Figure~\ref{fig:ablation}, we conduct comprehensive ablation studies for each of them on CIFAR-10 and ResNet-18, regarding the trade-off between clean accuracy (\ie, right y-axis and green line) and adversarial robustness (\ie, left y-axis, along with blue and orange lines respectively for PGD-10 and AA).

\textbf{Ablation Study on $t$} (Figure~\ref{fig:ablation1})\textbf{.} The first important phenomenon observed is that an appropriate epoch $t$ to start integrating DUCAT matters. Specifically, when starting too early (\eg, $t=100$), the overfitting to the training adversary PGD-10 can be serious, resulting in up to 20\% robust generalization gap to test-time AA. Yet, if starting too late (\eg, $t=120$), the learning can be insufficient in the first place, with both accuracy and robustness unsatisfying. This also reveals an important potential of DUCAT to serve as an adversarial \textbf{fine-turning} technology that can swiftly enhance already implemented robust DNN without retraining them from scratch. More specifically, provided that the existing target model is originally built through conventional AT by an appropriate epoch (\eg, epoch $100$), we can resume the training from that epoch with DUCAT integrated. Then after a few epochs (\eg, before epoch $130$), the model can achieve a better trade-off between clean accuracy and robustness beating existing SOTAs. This makes DUCAT also valuable in updating real-world robust applications.

\textbf{Ablation Study on $\beta_1$} (Figure~\ref{fig:ablation2}) \textbf{and $\beta_2$} (Figure~\ref{fig:ablation3}) \textbf{.} For the two-hot soft label construction weight $\beta_1$ and $\beta_2$ respectively for benign and adversarial samples, though as expected, the accuracy (or robustness) increases while robustness (or accuracy) drops as $\beta_1$ (or $\beta_2$) approaches 1, as long as within a reasonable range such that the original and dummy labels respectively serve as the primary learning target of benign and adversarial samples, the specific weight $\beta_1$ and $\beta_2$ mildly impact DUCAT performance, demonstrating its stability. Interestingly, the optimal range of $\beta_2$ (\eg, [0.9, 1]) is larger in value than $\beta_1$ (\eg, [0.7, 0.8]), which is probably because learning adversarial samples is still harder than benign ones after being separated as different objectives, thus need to ensure more training samples. 

\textbf{Ablation Study on $\alpha$} (Figure~\ref{fig:ablation4})\textbf{.} The hyper-parameter $\alpha$ is designed to explicitly adjust the weights of benign and adversarial losses for the model update procedure, directly injecting the learning preference to attach more importance to either benign or adversarial samples, which is a general strategy not specific to our method and has a predictable impact (\ie, either better accuracy or robustness). In this work, we would not suggest fine-tuning $\alpha$ as an approach to impact the trade-off, because our idea implies equal importance of benign and adversarial samples respectively regarding the learning of accuracy and robustness, and our unique two-hot soft label construction has already considered an appropriate balance between them. However, as some previous works on the trade-off problem adopt a similar hyper-parameter (\eg, the regularization parameter $\lambda$ in TRADES), we still provide such an option in case our user does have any specific requirements to customize it in their practices. Predictably, as illustrated in Figure~\ref{fig:ablation4}, as $\alpha$ increases, which means more importance is attached to the benign samples, there is overall an increasing (or decreasing) trend for the clean accuracy (or robustness) of the model. Nevertheless, it is also worth noting that $\alpha = 0.5$ as our default one, which means no specific preference on accuracy or robustness, is basically the optimal choice, just as we expected.

\section{Further Discussion}\label{app:discussion}

In this section, we provide additional discussion on three aspects where further clarifications might be found useful, namely 1) the difference with DuRM~\citep{wang2023frustratingly}, which involves dummy classes in a different way, 2) the comparison with the conventional soft-label strategy, as we propose a novel two-hot soft label, and 3) the specific analysis about TRADES + DUCAT, since the improvements there is not as impressive as others.

\subsection{Difference with DuRM}\label{subapp:albeit}

Albeit the previous work DuRM~\citep{wang2023frustratingly} also involves additional dummy classes, there are three essential differences between it and our work:
\begin{itemize}[leftmargin=2em] 
    \item 
    \textbf{Motivation}. DuRM expects dummy classes to provide implicit supervision for original ones to facilitate standard generalization of ERM. Differently, we involve dummy classes to tolerate different distributions between benign and adversarial samples, so that the two AT objectives namely clean accuracy and robustness can no longer be at odds with each other.
    \item 
    \textbf{Significance}. DuRM tends to adopt a smaller $C_d$ (\eg, typically 1 or 2) than $C$, and expects there should be no samples actually classified to the dummy classes. In contrast, our paradigm always adopts another $C$ dummy classes, clearly encouraging adversarial samples to be classified into them, and utilizing the one-to-one correspondences between original and dummy classes to detect and recover adversarial samples at inference time, thus acquiring certain robustness.
    \item 
    \textbf{Achievement}. Despite AT has been mentioned as one of several application scenarios of DuRM, as it actually does not involve any special designs for AT, the reported improvement is trivial (\ie, no significant improvement compared with PGD-AT). On the contrary, our approach achieves SOTA performance as shown in Section~\ref{sec:exp}.
\end{itemize}

\subsection{Difference between Our Two-Hot Soft Label and Conventional Ones}\label{subapp:two-hot}

Careful readers may notice that the suggested two-hot soft label is distinguished from the conventional soft label, such as the ones proposed for label smoothing~\citep{muller2019does,shafahi2019label,wu2024annealing}, in both motivation and specific implementation. In short, previous soft label technologies convert one-hot label vectors into one-warm vectors that represent a low-confidence classification~\citep{shafahi2019label}. In contrast, the two-hot soft label for DUCAT is neither built in a one-warm format nor aiming at a low-confidence classification. More specifically, from the perspective of motivation, our two-hot soft label is to explicitly bridge corresponding original and dummy classes as the suboptimal alternative target of each other, so that their separation also becomes learnable, while the conventional ones aim to promote learning from the soft target as better guidance that reflects the underlying distribution of data~\citep{muller2019does,shafahi2019label,wu2024annealing}. On the other hand, from the perspective of the specific implementation, different from conventional soft label combining one-hot target with uniform or crafted distribution~\citep{muller2019does,shafahi2019label,wu2024annealing}, our two-hot soft label just combines two one-hot targets namely as the primary and alternative targets. All in all, the unique two-hot soft label is proposed for the first time, strongly supporting the outstanding effectiveness of DUCAT.

\subsection{Analysis about Less Improvement of TRADES + DUCAT than Others}\label{subapp:trades+ducat}

Although DUCAT as an independent AT method (referred to as PGD-AT + DUCAT) achieves SOTA performance, it helps relatively less in enhancing TRADES than other AT benchmarks when serving as a plug (\ie, the improvement of TRADES + DUCAT over TRADES is not impressive as others). In our opinion, this is due to the particular training adversarial samples used by TRADES, which is less appropriate for building dummy clusters than typical ones. 

Specifically, different from most AT methods including the other three benchmarks, PGD-AT, MART and Cons-AT, that by default generate training adversarial samples by PGD-10, TRADES particularly crafts training adversarial samples through maximizing its own KL-divergence regularization term~\citep{zhang2019theoretically}. Because TRADES defines a boundary error measured through this KL term, which occurs when specific data points are sufficiently close to the decision boundary that can easily cross it under slight perturbation~\citep{zhang2019theoretically}, thus it makes sense to accordingly craft training adversarial samples to reduce this error. But as a consequence, these training adversarial samples naturally focus more on boundary establishment instead of reasonable data distribution patterns, which predictably degrades our DUCAT, as DUCAT assumes most of the adversarial samples from one class should belong to the corresponding dummy class.

\section{Limitations}\label{app:limit}

Despite the significant advancement achieved by this work, here we identify two limitations respectively for our new AT paradigm and the proposed DUCAT method. We hope these discussions could facilitate future works on the trade-off problem between clean accuracy and adversarial robustness among the AT community. 

Firstly, although the conventional trade-off between accuracy and robustness in the current AT paradigm has been released by our new AT paradigm, with both of them achieving new SOTA performance, we can still observe certain tension between these two objectives, though much slighter than before. It can be found in the ablation studies illustrated by Figures~\ref{fig:ablation} that the trend of clean and robust curves has still not fully aligned with each other in certain ranges. This implies that our new AT paradigm still does not harmonize them perfectly, thus just serving as a stepping stone instead of an end-all solution. Also, while the clean accuracy under DUCAT is significantly improved compared with the current AT methods, there is still a gap compared with standard training (\eg, the SOTA accuracy of ResNet-18 on CIFAR-10 in the clean context is about 95\%). 

Secondly, integrating DUCAT with different benchmarks shows different degrees of advancement. As discussed in Appendix~\ref{subapp:trades+ducat}, compared with the remarkable results with PGD-AT and Cons-AT, as well as the considerable one with MART, the outcomes on TRADES seem less impressive. This implies that, despite the effectiveness and simplicity of the proposed DUCAT, there should be room for better specific methods under our new AT paradigm to outperform it, especially regarding the different mechanisms of specific benchmarks like TRADES.

\end{document}

%% file: math_commands.tex
\usepackage{amsmath,amsfonts,bm}

\def\eqref#1{equation~\ref{#1}}

\def\1{\bm{1}}

\DeclareMathAlphabet{\mathsfit}{\encodingdefault}{\sfdefault}{m}{sl}
\SetMathAlphabet{\mathsfit}{bold}{\encodingdefault}{\sfdefault}{bx}{n}

\DeclareMathOperator*{\argmax}{arg\,max}
\DeclareMathOperator*{\argmin}{arg\,min}